\documentclass[Afour,sageh,times]{sagej}

\usepackage{moreverb,url}
\usepackage[bookmarksopen,bookmarksnumbered,citecolor=red,urlcolor=red]{hyperref}
\usepackage{todonotes}
\usepackage{tcolorbox}
\usepackage{algorithmic}
\usepackage{algorithm}
\newtcolorbox[auto counter]{algorithmbox}[2][]{colback=red!5!white,colframe=red!75!black,fonttitle=\bfseries, title=\alg\thetcbcounter: #2,#1}

\newcommand\BibTeX{{\rmfamily B\kern-.05em \textsc{i\kern-.025em b}\kern-.08em
T\kern-.1667em\lower.7ex\hbox{E}\kern-.125emX}}

\newcommand{\alg}{Algorithm~}

\newcommand{\eq}{Eq.~}
\newcommand{\fig}{Fig.~}

\newcommand{\rf}{F}
\newcommand{\rl}{L}
\newcommand{\om}{V}

\newcommand{\cm}{M}

\newcommand{\hc}{C}
\newcommand{\rg}{G}
\newcommand{\qd}{Q}
\newcommand{\gpm}{T}
\newcommand{\coll}{W}
\newcommand{\pdf}{\mathbf{pdf}}

\newcommand{\argmax}[1]{\underset{#1}{\operatorname{argmax}}\medspace}

\graphicspath{{resources/}}

\def\volumeyear{2018}

\begin{document}

\runninghead{Kopicki et al.}

\title{Learning better generative models for dexterous, single-view grasping of novel objects}

\author{Marek S. Kopicki\affilnum{1}, Dominik Belter\affilnum{2} and Jeremy L. Wyatt\affilnum{1}}

\affiliation{\affilnum{1}School of Computer Science, University of Birmingham, Edgbaston, Birmingham, B15 2TT.\\
\affilnum{2}Institute of Control and Information Engineering, Poznan University of Technology, Poland.}

\corrauth{Jeremy L. Wyatt, School of Computer Science, University of Birmingham, Edgbaston, Birmingham, B15 2TT.
}

\email{jeremy.l.wyatt@gmail.com}

\begin{abstract}
This paper concerns the problem of how to learn to grasp dexterously, so as to be able to then grasp novel objects seen only from a single view-point. Recently, progress has been made in data-efficient learning of generative grasp models which transfer well to novel objects. These generative grasp models are learned from demonstration (LfD). One weakness is that, as this paper shall show, grasp transfer under challenging single view conditions is unreliable. Second, the number of generative model elements rises linearly in the number of training examples. This, in turn, limits the potential of these generative models for generalisation and continual improvement. In this paper, it is shown how to address these problems. Several technical contributions are made: (i) a view-based model of a grasp; (ii) a method for combining and compressing multiple grasp models; (iii) a new way of evaluating contacts that is used both to generate and to score grasps. These, together, improve both grasp performance and reduce the number of models learned for grasp transfer. These advances, in turn, also allow the introduction of autonomous training, in which the robot learns from self-generated grasps. Evaluation on a challenging test set shows that, with innovations (i)-(iii) deployed, grasp transfer success rises from 55.1\% to 81.6\%. By adding autonomous training this rises to 87.8\%. These differences are statistically significant. In total, across all experiments, 539 test grasps were executed on real objects.
\end{abstract}

\keywords{learning, generative models, dexterous grasping}

\maketitle

\label{sec:introduction}
\section{Introduction}

Dexterous grasping of novel objects is an active research area. The scenario considered here is that in which a novel object must be dexterously grasped, after being seen from just a single viewpoint. We refer to this scenario as {\em dexterous, single-view grasping of novel objects}. This is essentially the grasping problem solved by humans and establishes a high bar for robots.

%
%

The combination of scenario features means that it is hard to apply planning methods based on analytic mechanics. This is because they require knowledge of frictional coefficients, object mass, and complete object shape to evaluate a proposed grasp. None of these are either known a priori nor easily recovered from a single view. 

Alternatively, there are learning methods. Broadly speaking, these divide into those that learn {\em generative models} and those that learn {\em evaluative models}. Generative models take sensor data as input and produce one or more candidate grasps. Evaluative models take sensor data and a grasp candidate as input and produce an estimate of the quality of that grasp on the target object. In this paper, we consider how to improve generative model learning.

\begin{figure}[t]
\centering
\begin{tabular}{cc}
\hspace{-2mm}\includegraphics[width=\columnwidth]{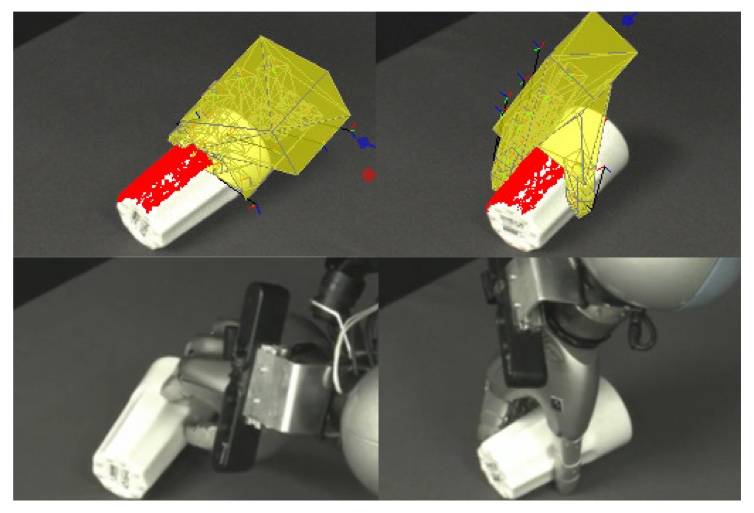}
\end{tabular}
\caption{{Left: a memory based generative model learner fails to grasp a beaker. Right: the new approach described here succeeds on the same case.}}
\label{fig:fail-success}
\end{figure}

This paper builds on recent work on learning grasps from demonstration (LfD). There are now LfD methods for learning probabilistic generative models of grasps. Specifically, the baseline method that this paper builds on learns such a model from a small number of examples, and can then use it to generate dexterous grasps for novel objects.  One drawback is that, while this baseline method can work for many single-view grasps, it is not reliable on challenging cases. Such cases include objects placed so that the surface recovery is limited, and where grasping requires contact to be made on a hidden back surface, as shown in Figure~\ref{fig:fail-success}. In addition, there are limits to its ability to take advantage of an increasing quantity of training data, since the approach is a purely memory-based learner and has no ability to combine models from different training examples.

This paper shows how to make single-view grasping more reliable. This involves several innovations. First, {\em (Innovation 1)} we show how to learn multiple, view-specific grasp models from a single example grasp. These view-specific models enable grasps to be generated that compensate for missing back surfaces of deep objects, a typical occurrence in single-view grasping (Figure~\ref{fig:fail-success}). Second, {\em (Innovation 2)} we move beyond memory-based models by showing how to combine information from multiple training grasps into a smaller number of generative models. This compression leads to an improvement in model generalisation and inferential efficiency on test objects.  Third, {\em (Innovation 3)} we present a novel way to calculate the likelihood of finger-object contacts in a candidate grasp. This new likelihood function is used both to generate and evaluate candidate grasps. Together, these three innovations improve the performance of dexterous, single-view grasping from 55.1\% to 81.6\% on a test set of novel objects placed in difficult poses. Finally, we show how learning can be scaled by using self-generated grasps on test objects as further training data. This raises the grasp success rate to 87.8\%.

Given these innovations, the paper tests the following hypotheses. 

\begin{itemize}
\item[{\bf H1}] Even without an enlarged set of training grasps, the combined innovations 1-3  improve the grasp success rate.

\item[{\bf H2}] View-based grasp modelling enables better generation of grasps for thick objects.

\item[{\bf H3}] The grasp success rate improves progressively as innovations are added.

\item[{\bf H4}] With all innovations the grasp success rate improves as the training data increases.

\item[{\bf H5}] With all innovations, learning is better able than the baseline algorithm to exploit an increased amount of training data.

\item[{\bf H6}] With all innovations the algorithm dominates the baseline algorithm without any innovations.
\end{itemize}


The paper is structured as follows. First, related work is described. Second, the approach is described in detail. This begins with a description of the basic framework for probabilistic generative grasp learning. It then proceeds to a description of the new learning algorithm, notably the view-based model representation and the technique for generative model compression. Following this, the new contact likelihood function and its uses are described. Finally, the results of an empirical study are presented.

\label{sec:relatedWork}
\section{Related work}

\subsection{Overview}

We identify four broad approaches to grasp planning. First, there are those that use analytic mechanics to evaluate grasp quality. Second, there are methods that engineer a mapping from sensor data to a candidate grasp or grasps. Third, there are methods that learn this mapping. Finally, there are methods that instead learn a mapping from sensor data and a candidate grasp to a prediction of grasp quality or grasp success probability. To place our work in context, we review the properties of each of these, plus relevant methods in grasp execution. We cannot do justice to the entire literature, but sketch the main developments.
Recent surveys of grasping include those by \citet{bohg2014data} on data-driven grasping and \citet{sahbani2012overview} on analytic methods.


\subsection{Analytic approaches}
Analytic approaches to grasping use the laws of mechanics to predict the outcome of a grasp~\citep{bicchi2000a,Liu2000,Pollard2004,miller2004}. These analyses require a model of the target object's shape, mass, mass distribution, and surface friction. They also need a model of the gripper kinematics and the exertable contact forces in different configurations. Obtaining these permits computation of the resistable external wrenches for a grasp. Based on this, a number of so-called grasp quality metrics can be defined~\citep{Ferrari1992,Roa2015,Shimoga1996}.

\begin{figure}[t]
\centering
\begin{tabular}{cc}
\hspace{-2mm}\includegraphics[width=1.0\linewidth]{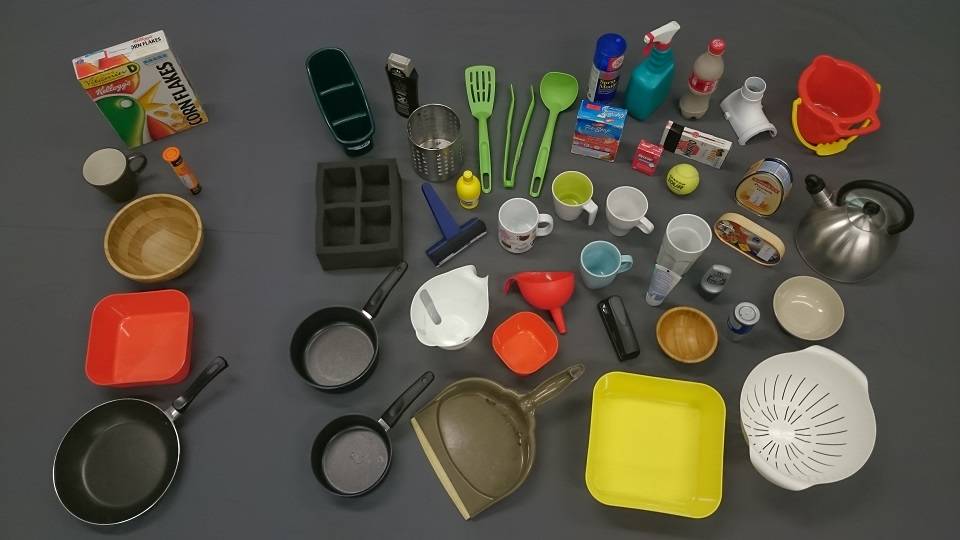}
\end{tabular}
\caption{{The objects used for training (left) and testing (right). A critical aspect of the actual training and testing sets is the object pose relative to a fixed initial camera location.}}
\label{fig:objects}
\end{figure}

Analytic approaches have been successfully applied to find grasps for multi-fingered hands~\citep{Boutselis2014,Gori2014,Hang2014,Rosales2012,Saut2012,ciocarlie2009hand}. All of these essentially pose grasp generation as optimisation against a grasp-quality metric. The appeal is that analytic methods are interpretable and scrupulous, but there are drawbacks. Estimation of the object's properties is challenging. One solution is to build a library of objects that might be grasped, matching partially reconstructed novel objects to similar ones in the library~\citep{goldfeder2011data}. Alternatively, parameters of the target object such as mass \citep{zheng2005a,shapiro2004a} or friction \citep{Rosales2014Active} must be recovered on the fly using vision and touch. One approach is for a complete object model to be recovered from a partial view (e.g. a single depth image) by assuming shape symmetries ~\citep{bohg2011a}, by using a 3D CNN~\citep{Varley2017}, or by a hierarchical shape approximation~\citep{huebner2008minimum}. The complete object can then be fed as input to an engine such as GraspIt. Search for a grasp can be improved by employing low-dimensional hand pose representations\citep{ciocarlie2009hand}. There are, however, several assumptions underpinning many analytic methods, such as hard contacts with a fixed contact area \citep{bicchi2000a} and static friction \citep{Shimoga1996}. There are methods that extend modelling to, for example, soft contacts \citep{ciocarlie2005grasp}. A more fundamental problem is that even a small error in the estimated shape, mass or friction can render an apparently good grasp unstable \citep{zheng2005a}. This can be mitigated to an extent by using independent contact regions (ICRs) \citep{ponce1995a,rusu2009c}. Despite this, there is some evidence grasp quality metrics based on mechanics are not strongly indicative of real-world grasp success \citep{bekiroglu2011b,kim2013a,goins2014a}. In addition, such metrics can be costly to compute many times during a grasp search procedure. Analytic models are quite general, however, and so can be used for tasks such as planning grasps in clutter \citep{dogar2012physics}.

\subsection{Engineered mappings from sensor to grasp}
Difficulties with analytic methods led to investigation of vision-based grasp planning. These methods use RGB or depth images, or representations like point clouds or meshes. Grasp generation becomes a search for object shapes that fit the robot's gripper \citep{popovic2010,trobina1995a,klingbeil2011a,richtsfeld2008a,kootstra2012a,pas2014a}.
This includes finding parallel object edges in intensity images \citep{popovic2010} or planar sections in range images \citep{klingbeil2011a}. Potential grasps can be also found by matching curved patches in a point cloud that can support contacts \citep{Kanoulas2017}. These candidate grasps are then refined using their shape and pose properties.  The rule based approach works well for pinch gripping, but does not scale well to dexterous grasping because of the increased size of the search space. Visual servoing can be used to improve grasp reliability~\citep{kragic03ijrr}. Partially known shapes can be grasped using heuristics both for grasp generation and for the reactive finger closing strategy used to execute the grasp under tactile sensing~\citep{Hsiao2010ContactreactiveGO}. Such reactive strategies can also be derived for pose and or shape uncertainty automatically in a decision theoretic framework~\citep{hsiao2011robust,arruda2016active}. These reactive strategies can include push manipulation to make a good grasp more likely~\citep{Dogar_2010}. Finally, the grasp itself may also be formulated by taking uncertainty into account as a constraint in the planning process~\citep{li2016dexterous}.

\begin{figure*}[t!]
\centering
\includegraphics[width=\textwidth]{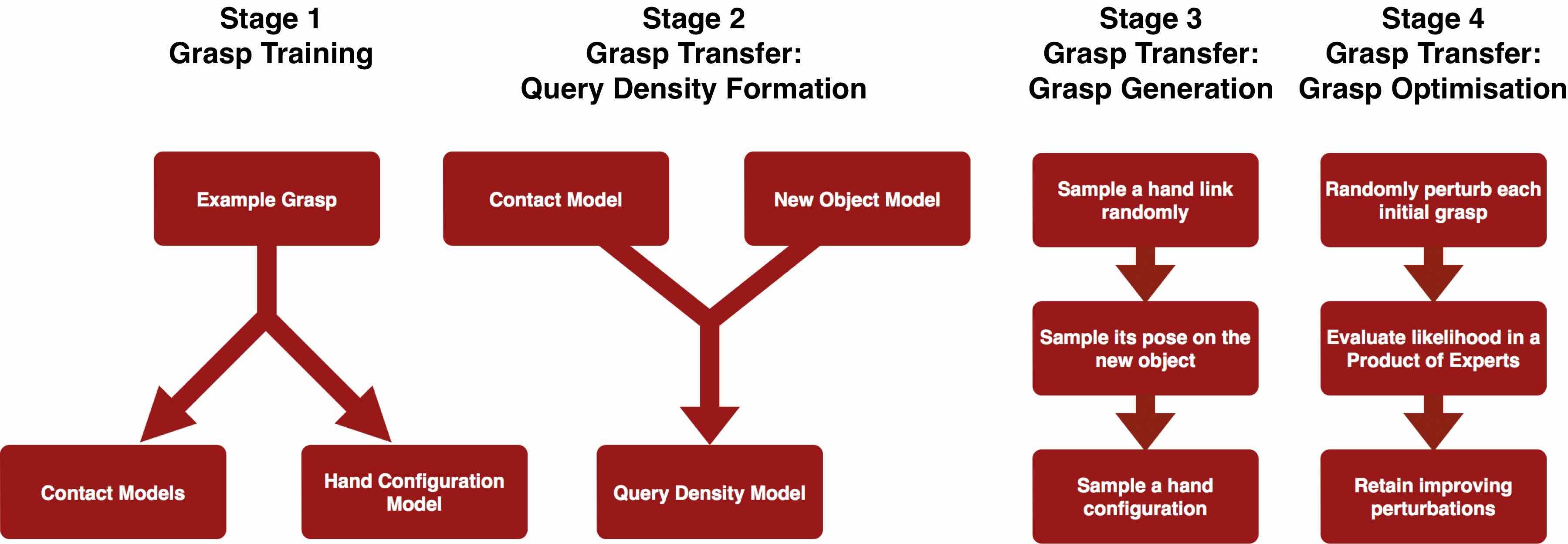}
\caption{The structure of grasp training and testing in four stages. Stage 1: an example grasp is shown kinesthetically. Multiple contact models (one for each hand-link) and a hand configuration model are learned. Stage 2: when a new object is presented a partial point cloud model is constructed and combined with each contact model to form a set of query densities. Stage 3: many grasps are generated, each by selecting a hand-link, sampling a link pose on the new object from the query density and sampling a hand configuration. Stage 4: grasp optimisation maximises grasp likelihood. This stage is repeated until convergence. \label{fig:flow} }
\end{figure*}

\subsection{Learning a mapping from sensor to grasp}
The next wave of grasp generation methods learned this mapping from data to grasp instead. Most of these methods learn relations between features extracted from the object representation, such as SIFT or other features~\citep{Saxena2008,fischinger2012}, shape primitives~\citep{Platt2006}, box-decompositions~\citep{huebner2008selection}  or object parts~\citep{kroemer2012a,detry2013a}. The grasp itself can be parametrised as a grasp position~\citep{Saxena2008}, gripper pose~\citep{Herzog2014} or a set of contact points~\citep{ben2012generalization, bohg2010a}. Some methods learn grasps from demonstration~\citep{ekvall2004a,hillenbrand2012a,kopicki2014a,kopicki2015ijrr,hsiao2006a}, and in the case of \citet{kopicki2015ijrr} create a generative model able to generate many grasp candidates for the target object. Others learn a distribution of possible grasps indexed by features from semi-autonomous grasping experiments \citep{detry2011a}. Recently, deep learning has been applied to learn such mappings \citep{Redmon2015,Kumra2017iros}.

\subsection{Learning a grasp evaluation function}
Learning approaches have also been applied to the problem of acquiring a grasp evaluation function from data. For example grasp stability for an executed grasp can be learned~\citep{bekiroglu2011assessing}. An evaluation of a proposed grasp can also be learned. This problem has been tackled recently using data intensive learning methods. Most of these methods predict the grasp quality for a parallel-jaw gripper~\citep{Pinto2016,Lenz2015,Johns2016,Mahler2016,Mahler2017,Redmon2015,Seita2016,Wang2016,Gualtieri2016,Levine2017}.
\citet{Pinto2016}, for example, learn a function that predicts the probability of grasp success for an image patch and gripper angle. 
To reduce the quantity of real grasps a rigid-body simulation~\citep{Johns2016,bousmalis2017using} or synthetic dataset~\citep{Mahler2016,Mahler2017} may be used. A synthetic data set requires that analytic grasp metrics be computed offline. 
\citet{Mahler2016} use a synthetic data set to predict the probability of force closure under uncertainty in object pose, gripper pose, and friction coefficient. \citet{Seita2016} performed supervised learning of grasp quality using deep learning and random forests. \citet{Gualtieri2016} predict whether a grasp will be good or bad using a CNN trained on depth images, using instance or category knowledge of the object to help. \citet{Hyttinen2015} used tactile signatures fed to a trained classifier to predict object grasp stability.

\subsection{Deep learning of dexterous grasping}
A small number of papers have explored deep learning as a method for dexterous grasping. \citep{lu2017planning,varley2015,veres2017modeling,zhou20176dof,kappler2015leveraging}. All of these methods use simulation to generate the set of training examples for learning. \citet{kappler2015leveraging} showed the ability of a CNN to predict grasp quality for multi-fingered grasps, but uses complete point clouds as object models and only varies the wrist pose for the pre-grasp position, leaving the finger configurations the same. \citet{varley2015} and later \citet{zhou20176dof} went beyond this, each being able to vary the hand pre-shape, and predicting from a single image of the scene. Each of these posed search for the grasp as a pure optimisation problem (using simulated annealing or quasi-Newton methods) on the output of the CNN. They all, also, take the approach of learning an evaluative model, and generate candidates for evaluation uninfluenced by prior knowledge. \citet{veres2017modeling}, in contrast, learn a deep generative model. Finally, \citet{lu2017planning} learn an evaluative model, and then, given an input image, optimise the inputs that describe the wrist pose and hand pre-shape to this model via gradient descent, but do not learn an evaluative model. In addition, the grasps start with a heuristic grasp that is then varied within a limited envelope. Of the papers on dexterous grasp learning with deep networks only those by \citet{varley2015} and \citet{lu2017planning} have been tested on real objects, with eight and five test objects each, producing success rates of 75\% and 84\% respectively. 

\subsection{Relation of this work to the literature}
The main similarities and differences between this work and previous methods reported in the literature may be summarised as follows. First, our method falls within the category of learning a mapping from sensory input to grasp. Thus, it differs from methods that learn an evaluation function for a proposed grasp. Second, like~\citep{kopicki2015ijrr} it learns a generative model, and is thus able to generate many candidate grasps for a new object, rather than just one.  One particular property of the method built upon~\citep{kopicki2015ijrr} is that it can learn from a very small number of demonstrations. There are two main drawbacks of that previous work. First, it is not sufficiently robust when grasping a novel object from a single view. Second, it is purely memory based, so it cannot merge learned models, and so doesn't extract the best models from an increasing amount data. In this paper, these drawbacks are addressed.

\section{Generative Grasp Modelling Basics}

The general approach is one of Learning from Demonstration (LfD). We first sketch the general structure of learning a generative grasp model from demonstration \citep{kopicki2015ijrr}. This structure is the same across both the algorithm described in \cite{kopicki2015ijrr} and the algorithm described here. For simplicity, we refer to the algorithm from \cite{kopicki2015ijrr} as the {\em vanilla} algorithm. Throughout we assume that the robot's hand comprises $N_L$ rigid \emph{links}: a palm, and several phalanges per finger. We denote the set of hand-links $L =\{L_i\}$. \footnote{For clarity we refer throughout to hand-links.} 

First, an example grasp is presented, and then a model of that grasp is learned. This grasp model comprises two different types of sub-model (Figure~\ref{fig:flow}: Stage 1). Both of these types of model are probability densities. The first type is a  {\em contact-model} of the spatial relation between a hand-link and the local object shape near its point of contact. A contact-model is learned for every hand-link in contact with the object. The second type is a {\em hand-configuration model}, that captures the overall hand-shape.

Given this learned model of a grasp, new grasps can be generated and evaluated for novel objects. This grasp transfer occurs in three stages (Figure~\ref{fig:flow}: Stages 2-4). First, each contact model is combined with the available point cloud model of the new object, to construct a third type of density called a {\em query density} (Stage 2). We build one query density for every contact model created during learning. Each query density is a probability density over where the hand-link will be on the new object.

Next, we generate initial candidate grasps on the new object (Stage 3) from the generative model of a grasp. Each candidate grasp is produced in three steps. First a hand-link is chosen at random. Then a pose for this hand-link on the new object is sampled. Finally, a configuration of the hand is sampled. The sampling of the hand-link pose uses the query density for that hand-link. The sampling of the hand-configuration uses the hand-configuration model. By forward kinematics these three samples (hand-link, hand-link pose and hand configuration) together determine a grasp. Many such initial grasps are generated in this way. 

In the final stage each initial grasp is refined by simulated annealing (Stage 4). The goal is to improve each initially generated grasp so as to maximise its likelihood according to the generative model. The optimisation criterion is a product of experts. There is one expert for each hand-link and one expert for the hand-configuration. These experts are the query-densities and the hand-configuration model. 

Generative grasp learning of this type has several desirable properties. First, it is known to be somewhat robust to partial point-cloud data for both training and test objects. Second, it displays generalisation to globally different test shapes. Third, the  speed of inference is quite good if the learner is only given a small number of training grasps (2000 candidate grasps are generated and refined in $<1$ sec on a modern 16-core PC). Fourth, there is evidence of robustness to variation in the orientation of the novel object. 

However, as mentioned previously, there are also weaknesses in this type of generative grasp model. These are: (i) a need to further improve robustness when grasping from a single-view of the test object; (ii) a need to extract the best possible generative models as the number of training grasps grows.

Having sketched the basic structure of generative grasp learning we now detail model learning and grasp inference for our new algorithm. We start by describing the basic representations that underpin the work. As we proceed we will highlight the innovations made.

\section{Representations}\label{sec:representations}

The method requires that we define several models: an object model (partial and acquired from sensing); a model of the contact between a hand-link and the object; and a model of the whole hand configuration. 

Since all of these models are probability densities, underpinning all of them is a density representation.
We first describe the kernel density representation we employ. Then we describe the various local surface descriptors we may use as the basis for the contact and query density models. We follow this with a description of each model type. 

\label{sec:representations.kde}
\subsection{Kernel Density Estimation}

$SO(3)$ denotes the group of rotations in three dimensions. A feature belongs to the space $SE(3) \times \mathbb R^{N_r}$, where $SE(3) = \mathbb R^3 \times SO(3)$ is the group of 3D \emph{poses}, and surface descriptors $r$ are composed of $N_r$ real numbers. We extensively use probability density functions (PDFs) defined on $SE(3) \times \mathbb R^{N_r}$.  We represent these PDFs non-parametrically with a set of $N$ features (or particles) $x_j$
\begin{equation}
S = \left\lbrace x_j : x_j \in \mathbb R^3 \times SO(3) \times \mathbb R^{N_r} \right\rbrace_{j \in [1,N]}.
\end{equation}
The probability density in a region is determined by the local density of the particles in that region. The underlying PDF is created through \emph{kernel density estimation} \citep{silverman1986a}, by assigning a kernel function $\mathcal{K}$ to each particle supporting the density, as
\begin{equation}\label{eq:d}
\pdf(x) \simeq \sum_{j=1}^N w_j \mathcal{K}(x| x_{j}, \sigma),
\end{equation}
where  $\sigma$ is the kernel bandwidth and $w_j \in \mathbb R^{+}$ is a weight associated with $x_j$ such that $\sum_j w_j = 1$. We use a kernel that factorises into three functions defined by the separation of $x$ into $p \in \mathbb R^3$ for position, a quaternion $q \in SO(3)$ for orientation, and $r \in \mathbb{R}^{N_r}$ for the surface descriptor. Furthermore, let us define $\mu$ and $\sigma$:
\begin{subequations}
\begin{align}
x &= (p, q, r),\label{eq:feature_a}\\
\mu &= (\mu_p, \mu_q, \mu_r),\label{eq:feature_b}\\
\sigma &= (\sigma_p, \sigma_q, \sigma_r).\label{eq:feature_c}
\end{align}
\label{eq:feature}
\end{subequations}
We define our kernel as
\begin{equation}\label{eq:kernel_in_se3}
\mathcal{K}(x | \mu, \sigma) = \mathcal{N}_3(p| \mu_p, \sigma_p) \Theta(q| \mu_q, \sigma_q) \mathcal{N}_{N_r}(r| \mu_r, \sigma_r)
\end{equation}
where $\mu$ is the kernel mean point, $\sigma$ is the kernel bandwidth, $\mathcal{N}_n$ is an $n$-variate isotropic Gaussian kernel, and ${\Theta}$ corresponds to a pair of antipodal von Mises-Fisher distributions which form a Gaussian-like distribution on $SO(3)$ \cite{fisher1953a,sudderth2006a}. It is the natural equivalent of the Gaussian for circular variables such as orientation. The value of ${\Theta}$ is given by
\begin{equation}
\Theta(q|\mu_q, \sigma_q) = C_4(\sigma_q) \frac {e^{\sigma_q \; \mu_q^T q} + e^{-\sigma_q \; \mu_q^T q}}2
\end{equation}
where $C_4(\sigma_q)$ is a normalising constant, and $\mu_q^T q$ denotes the quaternion dot product. 

Using this representation, conditional and marginal probabilities can easily be computed from \eq\eqref{eq:d}.
The marginal density $\pdf(r)$ is computed as

\begin{subequations}
\begin{align}
\label{eq:marginal}
\pdf(r) & \! \! = \! \! \! \iint \! \! \sum_{j=1}^N \! \! w_j \mathcal{N}_3(p| p_j, \sigma_p) \Theta(q| q_j, \sigma_q) \mathcal{N}_{N_r}(r| r_j, \sigma_r) \textnormal{d}p\textnormal{d}q \\
& =  \sum_{j=1}^N w_j \mathcal{N}_{N_r}(r| r_j, \sigma_r),
\end{align}
\end{subequations}
where $x_j = (p_j, q_j, r_j)$.
The  conditional density $\pdf(p,q|r)$ is given by
\begin{equation}\label{eq:conditional}
\pdf(p,q|{r}) = \frac{\pdf(p, q, {r})}{\pdf({r})}\\
\end{equation}

\subsection{Surface Descriptors}\label{sec:representations.features}

To condition the contact models it is necessary to have some descriptor of the local surface properties in the region of the contact. In principle these descriptors could capture any property, including local curvatures, surface smoothness, etcetera, that may influence the finger pose. In this paper we consider two different surface descriptors based solely on point-cloud data: local curvatures and fast point feature histograms (FPFH). We briefly describe the former. The latter is described in \citet{rusu2009a}.

\label{sec:representations.features.curv}
The principal curvatures are the surface descriptor used in the vanilla algorithm. To create the descriptor, all the points in the point cloud are augmented with a frame of reference and a local curvature descriptor. For compactness, we also denote the pose of a feature as $v$. As a result,
\begin{equation}
x = (v, r), \qquad v = (p, q).
\label{eq:surface.feature}
\end{equation}

The surface normal at $p$ is computed from the nearest neighbours of $p$ using a PCA-based method (e.g. \cite{kanatani2005statistical}). 
The surface descriptors are the local \emph{principal curvatures}, as  described in \citet{spivak1979comprehensive}. Their directions are denoted $k_1, k_2 \in \mathbb R^3$, and the curvatures along $k_1$ and $k_2$ form a $2$-dimensional feature vector $r = (r_1, r_2) \in \mathbb R^2$. 
The surface normal and the principal directions define the orientation $q$ of a frame that is associated with the point $p$.




\begin{figure}[!t]
\centerline{\includegraphics[width=\columnwidth]{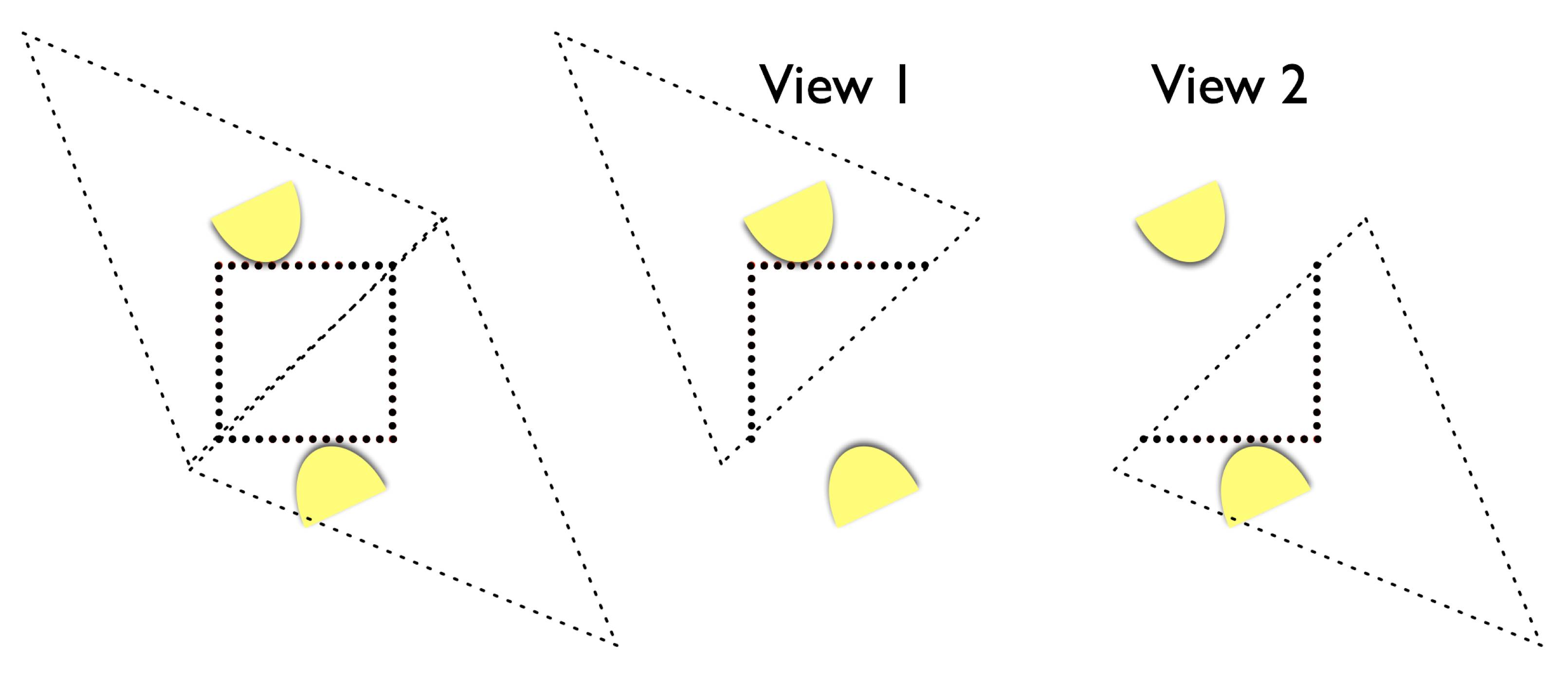}}
\caption[Views and Contacts]{A two fingered grasp of an object, shown in cross-section. Left: the vanilla algorithm incorporates point-cloud data from all views. Centre and right: the new algorithm learns a separate model for each view. \label{fig:views-diagram}}
\end{figure}

\subsection{Object View Model}\label{sec:representations.object}

The new method proposed in this article requires a set of view-based models of each training object. Let there be several object-grasp training examples $g=1 \ldots N_{\rg}$. Let each of these  examples be observed from $m$ views $m=1 \ldots N_{\om_{g}}$. A model of the grasped object from this view, denoted
$\om_{mg}(v, r)$, is computed from a single point cloud $m$ captured by a depth sensor as a set of $N_{\om_{mg}}$ features $\lbrace (v_{jmg}, r_{jmg}) \rbrace$. This set of features defines, in turn, a joint probability distribution, which we call the \emph{object-view model}:
\begin{equation}
\om_{mg}(v, r) \equiv \pdf_{mg}^\om(v, r) \simeq \sum_{j=1}^{N_{\om_{mg}}} w_{jmg} \mathcal{K}(v, r|{x_{jmg}}, \sigma_{x})
\label{eq:om}
\end{equation}
where $\om$ is short for $\pdf^\om$, $x_{jmg} = (v_{jmg}, r_{jmg})$,  $\mathcal{K}$ is defined in \eq\eqref{eq:kernel_in_se3} with bandwidth $\sigma_{x} = (\sigma_{v}, \sigma_{r})$, and where all weights are equal, $w_{jmg} = 1/{N_{\om_{mg}}}$. In summary, this object-view model $\om_{mg}$ represents the object surface as a pdf over surface points and descriptors.


\section{Contact Learning}\label{sec:learning}

Having set up the basic representations, we now describe the learning procedure. This includes the new view-based grasp model, and the procedure to merge contact-models learned from different grasp examples. This section corresponds to the left branch of Stage 1 of Figure~\ref{fig:flow}.

\label{sec:learning.views}
\subsection{Views}
\begin{figure}[t!]
\centerline{\includegraphics[width=0.8\columnwidth]{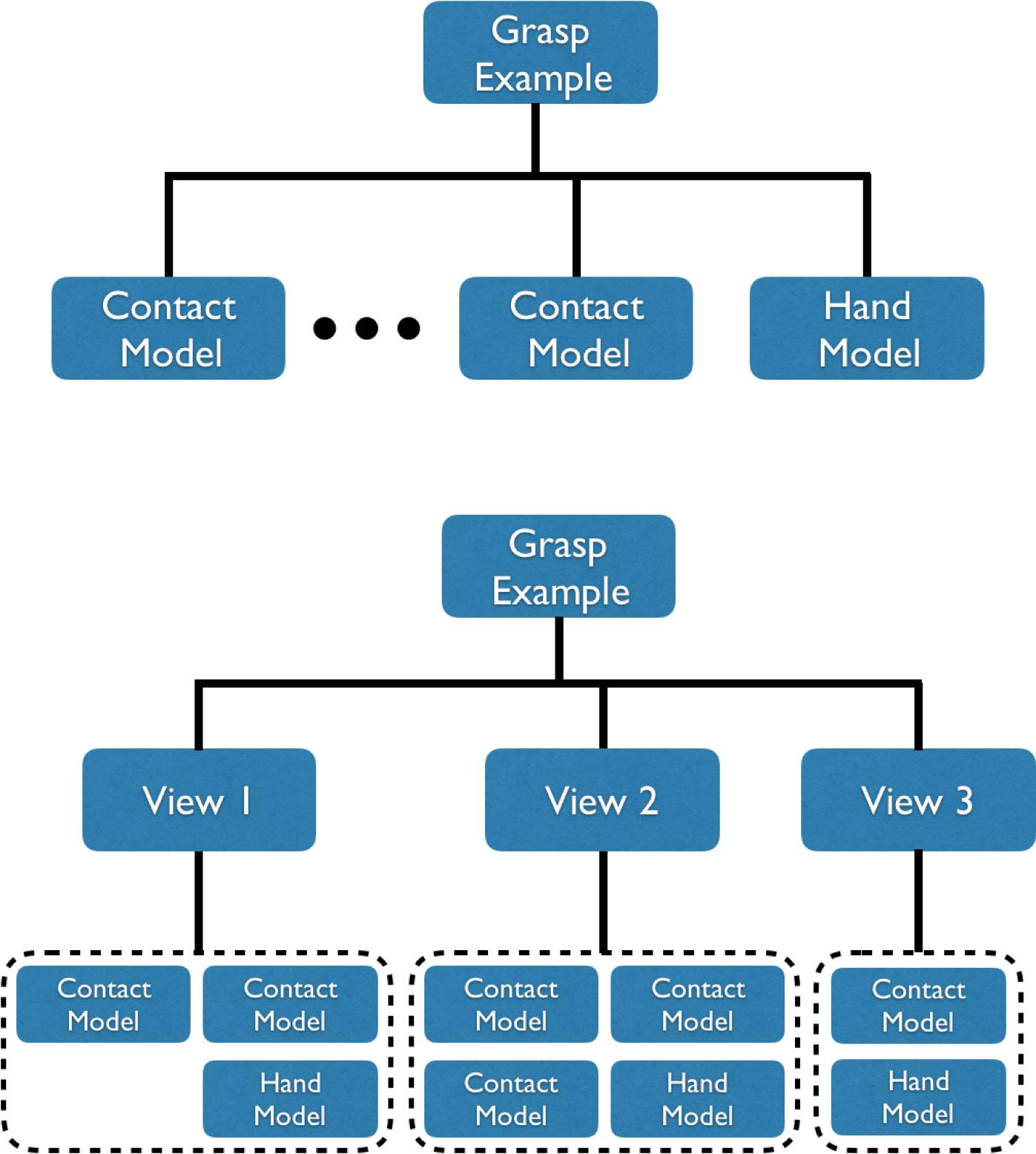}}
\caption[Grasp Model Organisation]{Top: in the vanilla model there is one model per grasp example. Bottom: in this paper there is one grasp model per view. Each view-based model contains contact models for the contacts that fall within the view and a copy of the hand-shape model. \label{fig:views-schema}}
\end{figure}
In \citet{kopicki2015ijrr} the representation required that all views of the training object were registered into a single point cloud (Figure~\ref{fig:views-diagram} left). A model of the grasp was learned from this registered point cloud. Instead, in this paper, a separate grasp-model is learned for each view (Figure~\ref{fig:views-diagram} centre and right). Each view-based grasp model contains both a model of the hand shape---thereby modelling all the hand-link positions relative to one another---and a model of each of the hand-object contacts that can be seen in that view. Thus each view-based grasp model excludes contact information for contacts it cannot see. 

This means that the grasp models are organised by view. (Figure~\ref{fig:views-schema}). The purpose is that the learned models more closely reflect the partial information available to the robot when grasping an object from a single available view. At inference time it means that the grasp optimisation procedure will not try to force all hand-links which were in contact in the training grasp into contacts with visible surfaces, instead relying on the hand shape model to implicitly guide hand-links to hidden back surfaces.

\label{sec:learning.recf}
\subsection{Contact Receptive Field}

Having defined the structure of the view-based grasp model, we now define the contact models that form part of it. This involves defining a receptive field around a contact, which determines how important different points on the object surface are in the contact model.

\begin{figure}[t]
\centerline{\includegraphics[width=6cm]{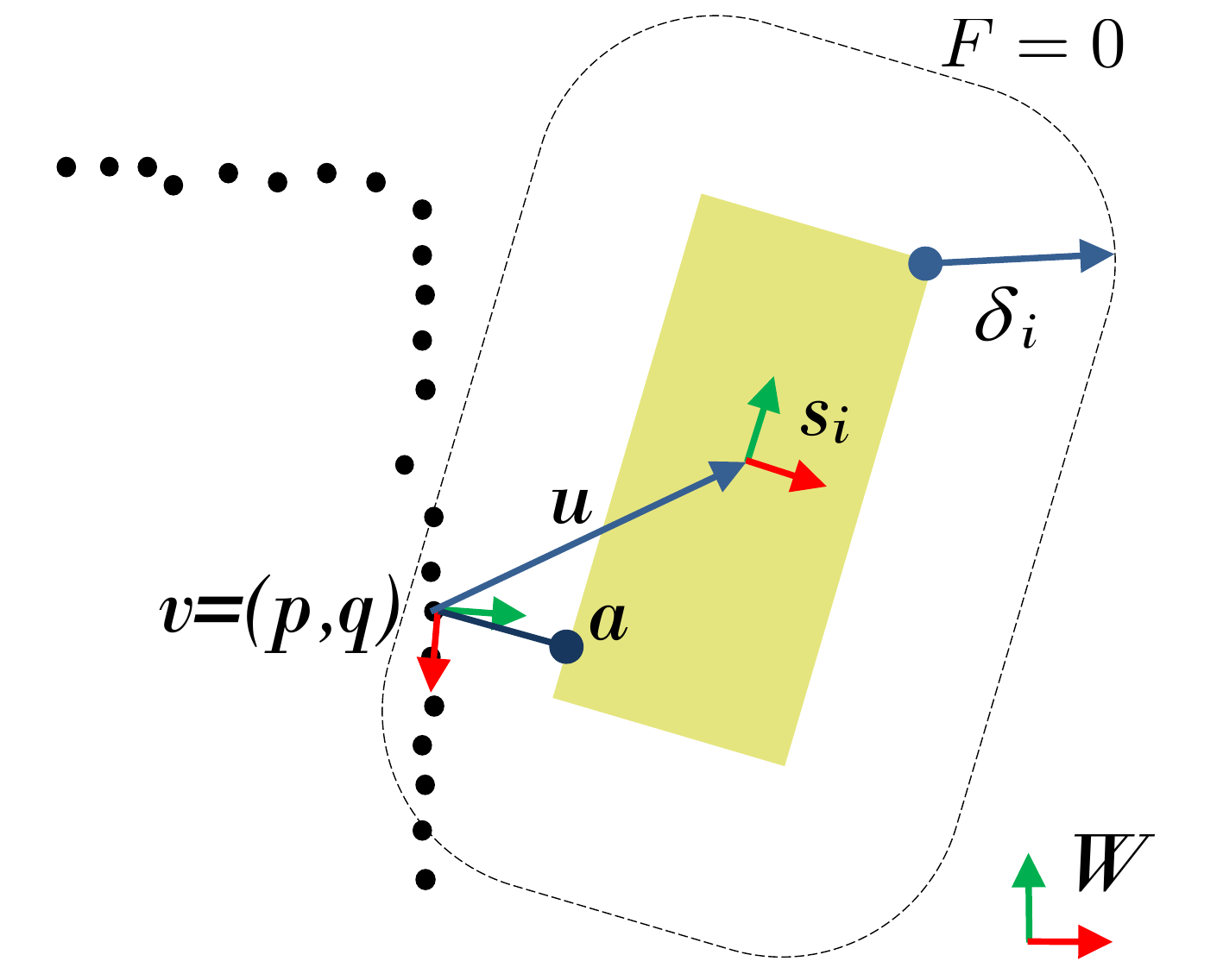}}
\caption[Contact receptive field]{The contact receptive field $\rf$ associated with the $i^{th}$ hand-link $\rl_i$ (solid yellow block) with link pose $s_i$. The black dots are samples from the surface of an object. The distance $a$ between feature $v$ and the closest point $a$ on the link's surface is shown. The rounded rectangle illustrates the cut-off distance $\delta_i$. The poses $v$ and $s_i$ are expressed in the world frame $W$. The arrow $u$ is the pose of $\rl_i$ relative to the frame for the surface feature $v$.}
\label{fig:contact_recfield}
\end{figure}

The \textit{contact receptive field} $\rf_i$ is a region of space relative to the associated hand-link $\rl_i$ (see \fig\ref{fig:contact_recfield}) which specifies the neighbourhood of that link. The contact receptive field $\rf_i$ is realised as a function of surface feature pose $v$:
\begin{equation}
\rf_{i} : SE(3) \rightarrow [0, 1]
\label{eq:contact_recfield_model}
\end{equation}
the value of which determines the relevance of a particular surface feature $x = (v, r)$ to a given hand-link $\rl_i$ in terms of the likelihood of the physical contact. We use contact receptive fields which are family of parameterised functions for which the value falls off quickly with the distance to the link:
\begin{equation}
\rf_{i}(v|\lambda_i,\delta_i) = \begin{cases}\exp(-\lambda_i ||p-a||^2) \quad &\textnormal{ if } ||p-a|| < \delta_i\\
0 \quad &\textnormal{ otherwise},\end{cases}
\label{eq:contact_recfield_func}
\end{equation}
where $\lambda_i > 0$ and $a$ is the point on the surface of $\rl_i$ that is closest to $v = (p, q)$. This means that the contact receptive field will only take account of the local shape, while falling off smoothly. A variety of monotonic, fast declining functions could be used instead.

\label{sec:learning.model}
\subsection{Contact Model Density}
Now we have defined the receptive field we can define the contact model itself. We denote by $u_{ij} = (p_{ij}, q_{ij})$ the pose of link $\rl_i$ relative to the pose $v_j$ of the $j$-th object feature. In other words, $u_{ij}$ is defined as
\begin{equation}
u_{ij} = v_j^{-1} \circ s_i,
\label{eq:local.pose}
\end{equation}
where $s_i$ denotes the pose of $\rl_i$, $\circ$ denotes the pose composition operator, and $v_j^{-1}$ is the inverse of $v_j$, with $v_j^{-1} = (-q_j^{-1}p_j, q_j^{-1})$ (see \fig\ref{fig:contact_recfield}). 

Contact model $\cm_{img}$ encodes the joint probability distribution of object surface features for the $i$-th hand-link, $m$-th view and $g$-th object-grasp example: 
\begin{equation}\label{eq:M}
\cm_{img}(U, R) \equiv \pdf^\cm_{img}(U, R)
\end{equation}
where $\cm_{img}$ is short for $\pdf^\cm_{img}$, $R$ is the random variable modelling object surface features of grasp-view example $g$, and $U$ models the pose of $\rl_i$ \emph{relative} to the frame of reference defined for the feature. In other words, denoting realisations of $R$ and $U$ by $r$ and $u$, $\cm_{img}(u, r)$ gives us the probability of finding $\rl_i$ at pose $u$ relative to the frame of a nearby object surface patch exhibiting surface descriptor $r$.

The contact model for link $i$, view $m$ and object $g$ is estimated as:
\begin{align}\label{eq:cm_n}
\cm_{img}(u, r) \simeq \frac 1Z \sum_{j=1}^{N_{mg}} \rf_{i}(v_{jmg}) \mathcal{K}(u, r | u_{ijmg}, r_{jmg}, \sigma ) 
\end{align}
where $Z > 0$ is a normalising constant, $u_{ijmg} = v_{jmg}^{-1} \circ s_i$, $\sigma = (\sigma_v, \sigma_r)$, $N_{mg}$ is the number of features in the point cloud, and $\mathcal{K}$ is kernel function \eqref{eq:kernel_in_se3} defined at poses from \eq\eqref{eq:local.pose}. 

We now also introduce, for the first time, the idea of a \textit{contact model norm}, which estimates the extent of the likely area of a physical contact of hand-link $i$ with surface features visible from view $m$ of grasp-object pair $g$:
\begin{equation}\label{eq:cm_norm}
\|\cm_{img}\| =  \sum_{j=1}^{N_{mg}} \rf_{i}(v_{jmg})
\end{equation}
We use this norm to help estimate which links are reliably involved in an grasp.

\subsection{Contact model selection}\label{sec:learning.selection}

A view-based learning framework has the consequence that not all learned models are useful. In any grasp-view pair some hand-links may make poor contacts with the observed parts of the object. This can also simply be caused by the relevant contact not being visible from a particular viewpoint. In both cases we must determine which contact-models should be created and which ignored. This is the purpose of lines 1-13 of Algorithm~\ref{alg:cm_proto}.

The contact model selection procedure determines, for a given grasp example $g$, view $m$ and hand-link $i$, whether the contact model $\cm_{img}$ should be created. It proceeds in two phases. The first phase uses the contact model norm \eqref{eq:cm_norm} to prune out unreliable grasps (Algorithm~\ref{alg:cm_proto}, lines 1-3). The decisions are recorded in a set of binary variables termed \textit{contact hypotheses}, $b_{img}$:

\begin{equation}
b_{img} = \begin{cases} 1 \quad &\textnormal{ if } \frac{N_\rl N_{\om_g} N_G \|\cm_{img}\|}{\sum_{jkl}\|\cm_{jkl}\|} > \eta_i \\
0 \quad &\textnormal{ otherwise},\end{cases}
\label{eq:cm_n_binary}
\end{equation}
where the contact-model is retained if $b_{img}=1$, $\eta_i \in \mathbb{R}^+$ is the threshold, $N_\rl$ is the number of hand links, $N_{\om_g}$ is the number of views of grasp example $g$, and $N_G$ is the number of grasp examples.

Having pruned out unreliable contact models from each grasp-view pair the second phase prunes out unreliable grasp-view examples  (Algorithm~\ref{alg:cm_proto}, lines 4-6). A grasp-view example is retained if the total number of non-empty contact models for a particular view $m$ and grasp $g$, determined by $b_{img}$, is higher than some minimum number $\zeta$. This is encoded in a set of binary variables termed \textit{view hypotheses}, $c_{mg}$:
\begin{equation}
c_{mg} = \begin{cases} 1 \quad &\textnormal{ if } \sum_i b_{img} > \zeta\\
0 \quad &\textnormal{ otherwise},\end{cases}
\label{eq:cm_binary}
\end{equation}
After a number of example grasps we will thus obtain a set of non-empty contact models (Algorithm~\ref{alg:cm_proto}, lines 7-13). We may index these using a triplet of indices $(i, m, g)$ corresponding to the hand-link, view and grasp example. Because of contact-model pruning not all $(i,m,g)$ will have a contact-model, i.e. for some views, links or grasps the contact model will be empty. 
We denote the set of indices for the non-empty contact models $\mathcal{M}$. The size of this set is $N_\mathcal{M}$.
\begin{equation}\label{eq:cm_selection}
(i,m,g) \in \mathcal{M}
\end{equation}

The parameters $\lambda$, $\eta$, $\zeta$ and $\sigma_{p}$, $\sigma_{q}$, $\sigma_{r}$ were chosen empirically. The time complexity for learning each contact model $\cm_{img}(u, r)$ is $O(T_i N_{\om_{mg}})$ where $T_i$ is the number of triangles in the tri-mesh describing hand link $i$, and $N_{\om_{mg}}$ is the number of features of view $m$ and example grasp $g$.

\subsection{Clustering Contact Models}\label{sec:learning.clustering}

So far, we have defined how a contact model is learned. Using this memory based scheme, the number of contact models $N_{\mathcal{M}}$ will grow linearly with the number of training examples. In a memory based learner, every contact model must be transferred to the target object. This may also limit the generalisation power of the contact models. This paper presents an alternative to memory based learning. We may exploit a growing number of training examples by merging contact models. This is the purpose of lines 14-22 of Algorithm~\ref{alg:cm_proto}. We hypothesize that this merging process will result in higher grasp success rates at transfer time. To merge models we first cluster the contact models according to similarity. Since each contact model is a kernel density estimator, the key step is to define an appropriate similarity measure between any pair of such estimators. 
Our principal aim is to produce a distance that is fast to compute and which is robust to the natural variations in the underlying data in the grasping domain. We define, first, an asymmetric divergence  and then, on top of it, a symmetric distance. Since we are using kernel density estimators we can most simply define a distance between the sets of kernel centres. 

First we define a distance between two kernels lying in $SE(3)$, $x=(p_x, q_x, r_x)$ and $y=(p_y, q_y, r_y)$ (see \eqref{eq:feature_a}), as a weighted linear combination of sub-distance measures:
\begin{equation}
d_\text{k}(x, y) = w_\text{lin}d_\text{lin}(p_x, p_y) + w_\text{ang}d_\text{ang}(q_x, q_y)\label{eq:distance_kernel}
\end{equation}
where $w_{lin}, w_{ang} \in \mathbb R^{+}$ are weights, and
\begin{subequations}
\begin{align}
d_\text{lin}(p_x, p_y) &= |p_x - p_y|^2,\label{eq:distance_kernels_lin}
\\d_\text{ang}(q_x, q_y) &= 1 - \|q_x\| \cdot \|q_y\|\label{eq:distance_kernels_ang}
\end{align}
\label{eq:distance_kernels}
\end{subequations}
The $\|.\|$ operator extracts the surface normal, and $\cdot$ denotes a dot product. 
Using this, we define the distance of kernel $x$ from kernel density $M$ as
\begin{equation}
d_\text{kd}(x, \cm) = \underset{y \in \cm}{\min} \{d_\text{k}(x, y) \}
\label{eq:distance_kernel_density}
\end{equation}
This distance considers only one kernel $y \in M$ that is the nearest to $x$. This has two major advantages. First, it allows the use of fast nearest neighbour search techniques with time complexity $\Omega(\log(N))$ rather than $\Omega(N)$, where $N$ is the number of kernels in $M$. Second, the distance given by~\eqref{eq:distance_kernel_density} is independent of the remaining kernels in the density $M$.~\footnote{This is useful in our domain since $M$ is constructed from a depth image taken from an RGBD camera. The density of points underpinning $M$ varies with the object-camera distance. Using a distance between $x$ and $M$ that depends only on the closest kernel in $M$ renders the distance much less sensitive to variations in the density of $M$. This improves generalisation.}
Additionally, for further efficiency, the distance measure ~\eqref{eq:distance_kernel_density} ignores kernel weights. This approach is valid since all weights are computed using~\eqref{eq:contact_recfield_func}. Thus, each weight depends only on the kernel position relative to the local frame of the relevant hand-link, which is already accounted for in the linear distance \eqref{eq:distance_kernels_lin}.

Next, we use this distance to define the {\em divergence} of kernel density $\cm_j$ from kernel density $\cm_i$
\begin{equation}
d_\text{dd}(\cm_i, \cm_j) = \frac{1}{N_{i}} \sum_{x \in \cm_i} d_\text{kd}(x, \cm_j)
\label{eq:distance_density_asym}
\end{equation}
where $N_i$ is the number of kernels of $\cm_i$. Note that this divergence  \eqref{eq:distance_density_asym} is asymmetric. For example $d_\text{dd}(\cm_i, \cm_j)$ may be large, while $d_\text{dd}(\cm_j, \cm_i) = 0$, if $\cm_j$ is constructed from $\cm_i$ by removing some large ``surface patch''. 



\begin{algorithm}[t]
\caption{Contact model selection and clustering \label{alg:cm_proto}}
    \begin{algorithmic}[1]
    \FORALL{grasps $g$, views $m$, links $i$} 
   		\STATE compute $b_{img}$ using Eq. \eqref{eq:cm_n_binary}
    \ENDFOR
    \FORALL{grasps $g$, views $m$}
    		\STATE compute $c_{mg}$ using Eq. \eqref{eq:cm_binary}
    \ENDFOR
    \STATE Set $\mathcal{M} = \{\}$
    \FORALL{grasps $g$, views $m$, links $i$}
    		\IF{$b_{img} > 0$ and $c_{mg} > 0$}
    			\STATE add index triplet $(i, m, g) \rightarrow \mathcal{M}$
    			\STATE $N_{\mathcal{M}} = |\mathcal{M}|$
    		\ENDIF
    	\ENDFOR
    	\STATE Set $D$ to be an $N_{\mathcal{M}} \times N_{\mathcal{M}}$ matrix 
    \FORALL{pairs of triplets $(k, j) \in \mathcal{M}$ such that $k \neq j$}
		\STATE compute distance $D_{kj} = d_{dd}(\cm_k, \cm_j)$ using Eq. \eqref{eq:distance_density_asym}
    \ENDFOR
    \STATE $\mathcal{C} \leftarrow$ affinity-propagation($\mathcal{M}, D$) 
    \FORALL{clusters $\mathcal{C}_l$}
		\STATE compute $M^\mathcal{C}_l$ using Eq. \eqref{eq:cm_prototype}
    \ENDFOR
	\RETURN $\{M^\mathcal{C}_l\}_{\forall_{l=1..N_\mathcal{C}}}$ 
    \end{algorithmic}
\end{algorithm}
To cluster the contact models, however, we require a symmetric distance, which we define as:
\begin{equation}
d(\cm_i, \cm_j) = \max \{d_\text{dd}(\cm_i, \cm_j), d_\text{dd}(\cm_j, \cm_i)\}
\label{eq:distance_density}
\end{equation}
It is worth noting some other benefits of this distance definition within our domain. First, \eqref{eq:distance_kernel_density} ignores surface descriptors. This is both because they can be high dimensional and because the shape properties they encode are already encoded in the remaining kernels of the contact models, and so are captured in \eqref{eq:distance_density_asym}. Thus, measuring the distance with respect to the surface descriptor adds no benefit. For the same reason, we do not compute the distances between pairs of local frames. Instead, we simply compare pairs of surface normals. 

This distance is calculated for every pair of contact models (Algorithm~\ref{alg:cm_proto}, lines 14-17).
The next stage is to cluster contact models using the metric defined by \eqref{eq:distance_density}. Many clustering procedures could be used.  We used affinity propagation \citep{frey2007clustering}, which requires computation of all pair-wise distances (Algorithm~\ref{alg:cm_proto}, line 18). Affinity propagation finds clusters together with cluster exemplars---the most representative cluster members. Clustering creates a partition $\mathcal{C} = \{ \mathcal{C}_{1} \ldots \mathcal{C}_{l} \ldots \mathcal{C}_{N_\mathcal{C}} \}$ of the set of contact models $\mathcal{M}$. 
Where $N_{\mathcal{C}}$ is the number of clusters and $\mathcal{C}_l$ denotes the $l$-th cluster, which is a set with $N_{\mathcal{C}_l}$ members. 
There is a one-to-one map from each contact model index $(i,m,g)$ onto its corresponding cluster and index within that cluster, $(l,k)$. Thus, we can write out cluster $\mathcal{C}_l$ as $\mathcal{C}_l = \{ M^{\mathcal{C}}_{l1}, \dots M^{\mathcal{C}}_{lk} \ldots M^{\mathcal{C}}_{l,N_{\mathcal{C}_l}} \}$. Note that, when referring to a contact model as a member of a cluster, we use the superscript $\mathcal{C}$ for clarity as to the meaning of the index in the subscript.

\begin{figure}[t]
\centerline{\includegraphics[width=\columnwidth]{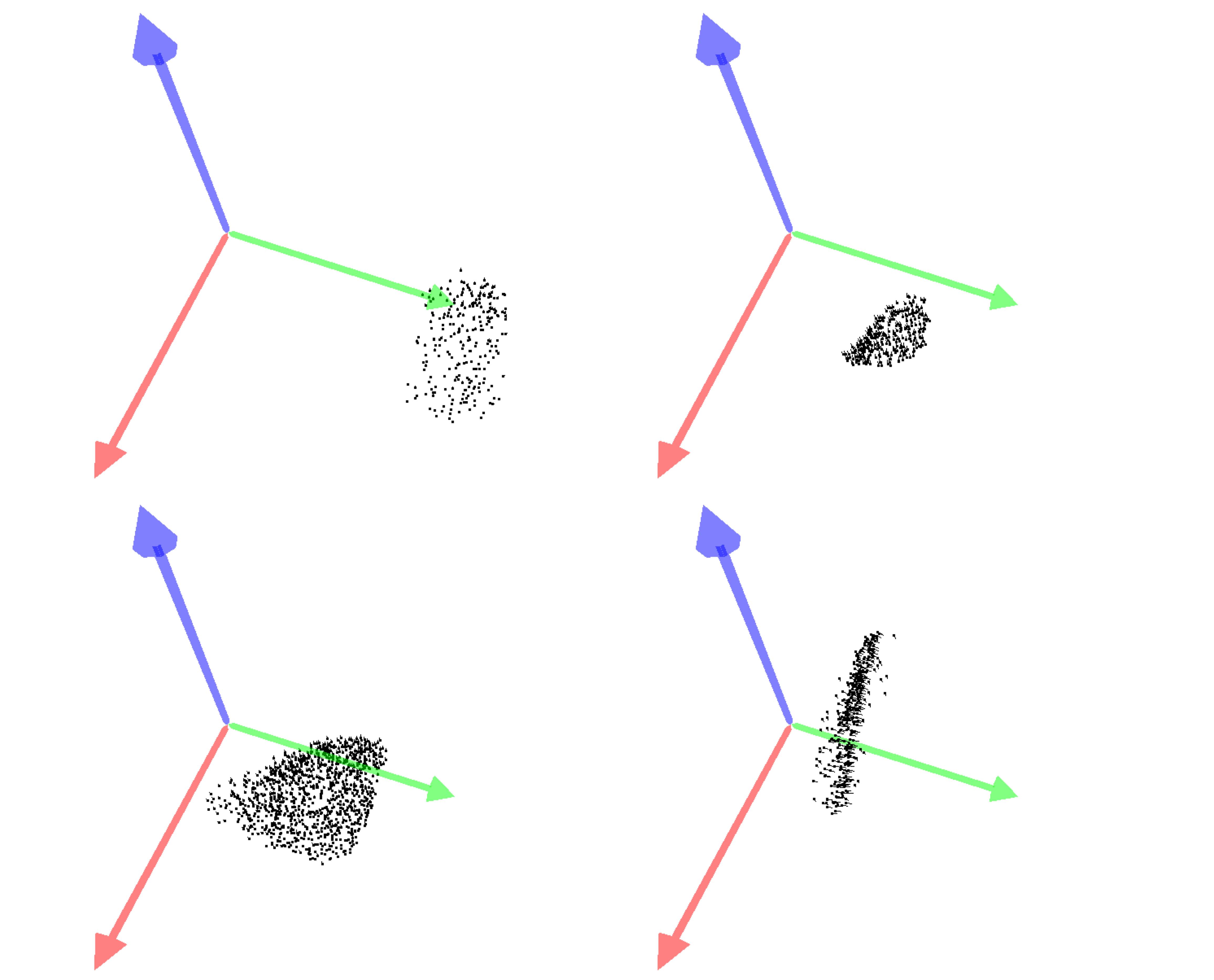}}
\caption[Contact model cluster centres]{The prototypes produced for four of the clusters after affinity propagation. Each picture shows the density over rigid body transformations for the surface relative to the frame (shown) attached to the finger link. Only the positions of the kernel centres (black dots) are shown. Surface descriptors and orientations have been marginalised out. \label{fig:cluster-centres}}
\end{figure}

In order to boost the generalisation capability of the clustered contact models (in particular for small clusters), we create \textit{cluster prototypes}, denoted $\cm^{\mathcal{C}}_l$, to replace cluster exemplars. We first define a multinomial distribution, for each cluster $l$, over the members $k$ of that cluster, $P(k|\mathcal{C}_l)$. This lets us define, in turn, a cluster prototype contact model $\cm^{\mathcal{C}}_l$ as a mixture model:
\begin{equation}
\cm^{\mathcal{C}}_l(u,r) = \sum_{M^{\mathcal{C}}_{lk} \in \mathcal{C}_l} P(k|\mathcal{C}_l) \cm^{\mathcal{C}}_{lk}(u, r)
\label{eq:cm_prototype}
\end{equation}
We evaluate this mixture with a simple Monte Carlo sampler. The probabilities $P(k|\mathcal{C}_l)$ are obtained using the density distance \eqref{eq:distance_density} between a cluster member with index $k$ and the cluster exemplar:
\begin{equation}
w^{\mathcal{C}}_{lk} = \exp \left(- \xi d(\cm^{\mathcal{C}}_{l1}, \cm^{\mathcal{C}}_{lk}) \right)
\label{eq:cm_prototype_weights}
\end{equation}
where $\xi > 0$ controls the spread of the probability density around the cluster exemplar. The $P(k|\mathcal{C}_l)$ are the normalised versions of the weights  $w^{\mathcal{C}}_{lk}$.
A cluster prototype is calculated for each cluster (Algorithm~\ref{alg:cm_proto}, lines 19-21).

\begin{figure}
\centerline{\includegraphics[width=1.1\columnwidth]{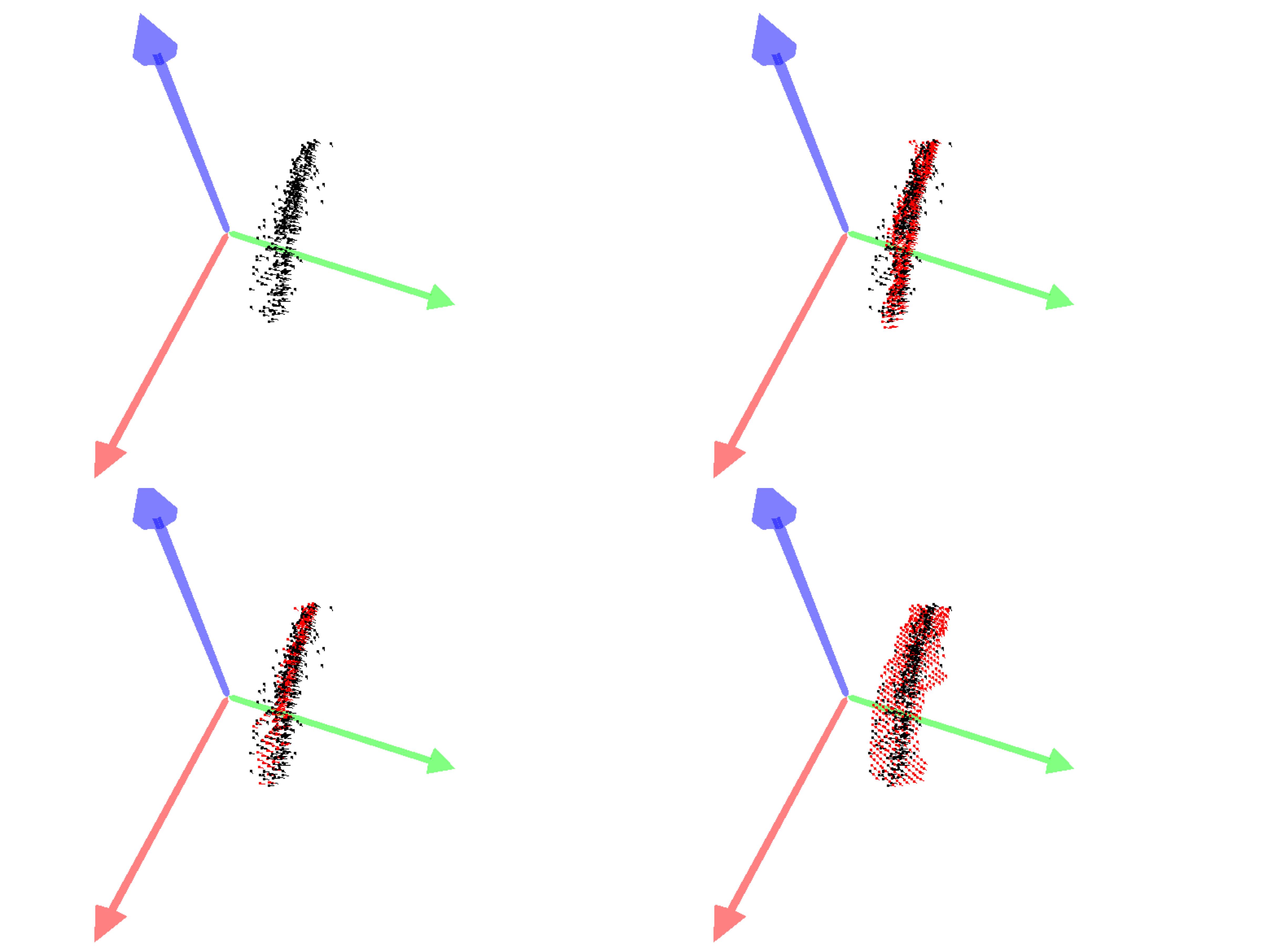}}
\caption[Contact model cluster members]{The cluster prototype (black dots) and three cluster members (red dots) for one of the clusters. This shows how the densities fall into relatively easily clustered types, showing that contact model merging via clustering does not lead to significant information loss.\label{fig:cluster-members}}
\end{figure}

We can visualise the resulting prototypes (Figure~\ref{fig:cluster-centres}) and the cluster members  (Figure~\ref{fig:cluster-members}). It can be seen that the clusters are coherent and well separated. This corresponds to the fact that, in terms of the link-to-surface relations, there are a distinct number of contact types.

\begin{figure}
\centerline{\includegraphics[width=0.9\columnwidth]{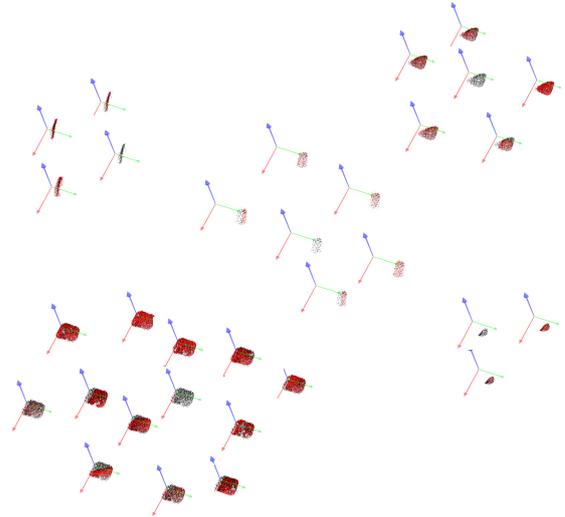}}
\caption{Visualisation of contact model clusters formed.\label{fig:clustering-results}}
\end{figure}

The parameters $w_\text{lin}$, $w_\text{ang}$ and $\xi$ were chosen empirically. The time complexity for computing all contact model pairwise distances is $O(N_\mathcal{M} (N_\mathcal{M} - 1) N \log(N))$ where $N_\mathcal{M}$ is the number of contact models after the selection procedure has been applied, and where $N$ is the average number of kernels in a contact model. The time complexity of the clustering algorithm is $O((N_\mathcal{M})^2 N_\text{steps})$ where $N_\text{steps}$ is the number of iterations of the affinity propagation algorithm~\cite{frey2007clustering}.

\subsection{Hand Configuration Model}

The hand configuration model $\hc^g$, for a grasp $g$, was originally introduced in \cite{kopicki2015ijrr} and remains the same here. It is thus described for completeness. It encodes a set of configurations of the hand joints $h_c \in \mathbb R^D$ (i.e., joint angles), that are particular to a grasp example $g$. The purpose of this model is to allow us to restrict the grasp search space, during grasp transfer, to be close to hand configurations in the training grasp. Learning this model is the right hand branch of Stage 1 in Figure~\ref{fig:flow}.

The hand configuration model encodes the hand configuration that was observed when grasping the training object, but also a set of configurations recorded during the approach towards the object. We denote by $h^t_c$ the joint angles at some small distance \emph{before} the hand reached the training object, and by $h^g_c$ the hand joint angles at the time when the hand made contact with the training object. We consider a set of configurations interpolated between $h^t_c$ and $h^g_c$, and extrapolated beyond $h^g_c$, as
\begin{equation}
h_c(\gamma) = (1 - \gamma)h^g_c + \gamma h^t_c
\label{eq:learning.configmodel.config}
\end{equation}
where $\gamma \in \Gamma$ and $\Gamma$ is a regularly spaced set of values from the real-line. For all $\gamma < 0$, configurations $h_c(\gamma)$ are beyond $h^g_c$. A hand configuration model $\hc$ is constructed by applying kernel density estimation 
\begin{equation}
\hc(h_c) \equiv \sum_{\gamma \in [-\beta, \beta]} w({h_c(\gamma)}) \mathcal{N}_D(h_c|h_c(\gamma), \sigma_{h_c}) 
\label{eq:hc}
\end{equation}
where  $w({h_c(\gamma)}) = \exp(-\alpha \|h_c(\gamma) - h^g_c \|^2)$ and $\alpha \in \mathbb R^{+}$. $\alpha$ and $\beta$ were hand tuned and kept fixed in all the experiments. One hand-configuration model $\hc^g$ is learned for each example grasp. The complexity of learning a hand-configuration model is $O(N_l N_g |\Gamma|)$, where $N_g$ is the number of example grasps.

Having completed the description of the learning procedure (Stage 1 in Figure~\ref{fig:flow}) we turn to describing our novel grasp inference procedures (Stages 2-4).

\section{Grasp Inference}\label{sec:inference}
The inference of grasps on a new object relies on three procedures: (i) a procedure of transferring contact models to the new object; (ii) a grasp generation procedure and (iii) a grasp optimisation procedure. These correspond to Stages 2, 3 and 4 of Figure~\ref{fig:flow} respectively.

In this paper the novel contribution to grasp inference is that we modify the procedure for transferring contact models so as to improve the quality of the proposed grasps. This is achieved by incorporating the density-divergence measure introduced earlier in the paper.

\subsection{Contact Query Density Computation}\label{sec:inference.query}

A query density is, for a particular hand-link and an object model, a density over the pose of that hand-link relative to the object. Query densities are used both to generate and to evaluate the likelihood of candidate grasps. Intuitively, the query density encourages a finger link to make contact with the object at locations with similar local surface properties to those in the training example. A query-density is simply the result of convolving two densities: a contact model density and the object model density. This section describes the formation of a query-density (Figure~\ref{fig:flow} Stage 2). The main innovation is that we present a new likelihood function for generating and evaluating finger contacts with the object. If we were to directly adopt the previous approach of \cite{kopicki2015ijrr} the query density $\qd^{C}_l$ would be defined as:
\begin{equation}
\qd^{\mathcal{C}}_l(s) = \iiint \gpm(s|u, v) \om(v,r) \cm^{\mathcal{C}}_l(u|r) \cm^{\mathcal{C}}_l(r) \textnormal dv \textnormal du \textnormal dr
\label{eq:qd.contactint}
\end{equation}
where $\om(v,r)$ is the \textit{test} object-view model \eqref{eq:om}. As described previously, this is a joint density over frames in \textit{global workspace coordinates} $v \in SE(3)$ and over surface descriptors $r \in \mathbb{R}^{N_r}$. The term $\gpm(s|u, v)$ is the Dirac delta function, since $s$ is determined by $u,v$. The relevant contact model is factored into the product  $\cm^{\mathcal{C}}_l(u, r) \equiv \cm^{\mathcal{C}}_l(u|r) \cm^{\mathcal{C}}_l(r)$. Algorithmically, this density is approximated using importance sampling.

In this paper, the query density is re-defined. Specifically, the term $\cm^{\mathcal{C}}_l(r)$ is replaced. 
This term defines a density over the test-object's surface shape $r$ around $v$, according to the contact model. But $r$ is only a low-dimensional summary of the ideal surface shape. To avoid the resulting loss of information we may, instead of $\cm^{\mathcal{C}}_l(r)$, use a conditional density over the {\em precise} surface shape in the neighbourhood of $v$. 
\begin{equation}
P(N^{(v)} | s, \cm^{\mathcal{C}}_l) 
\end{equation}
where $N^{(v)}$ is the surface patch, on the test object, in the neighbourhood of $v$. 

%
%
%
Substituting this for $\cm^{\mathcal{C}}_l(r)$ gives us a new query density definition:
\begin{equation}
\qd^{\mathcal{C}}_l(s) \! \! = \! \! \! \iiint  \! \! \gpm(s|u, v) \om(v,r) \cm^{\mathcal{C}}_l(u|r) P(N^{(v)}|s,\cm^{\mathcal{C}}_l) \textnormal dv \textnormal du \textnormal dr
\end{equation}
We desire that the more alike the test surface is to the training surface the higher $P(N^{(v)} | s, \cm^{\mathcal{C}}_l)$ should be.  
The density divergence defined earlier is ideal for this purpose:
\begin{equation}P(N^{(v)} | s, \cm^{\mathcal{C}}_l) \propto \exp\left( -\phi d_\text{dd}(s \diamond
\cm^{\mathcal{C}}_l, \om)\right)
\label{eq:contact-model-object-model-dist}
\end{equation}
where $\phi$ is a constant and we define $s \diamond
\cm^{\mathcal{C}}_l$ as a composition of a transform and a set. First, recall that $\cm^{\mathcal{C}}_l = \{(u_{lj}, r_{lj})\}_{j=1:N_{\mathcal{C}_l}}$ and that 
$s = v \circ u$.
Then, for every pair $(u_{lj}, r_{lj})$ in $\cm^{\mathcal{C}}_l$, we simply compose $s$ and $u_{lj}^{-1}$ 
\begin{equation}
s \diamond (u_{lj}, r_{lj}) = (s \circ u_{lj}^{-1}, r_{lj})
\end{equation}
So that, when we extend this to $\cm^{\mathcal{C}}_l$, we obtain
\begin{equation}
s \diamond \cm^{\mathcal{C}}_l = \{(s \circ u_{lj}^{-1}, r_{lj})\}_{j=1:N_{\mathcal{C}_l}}
\end{equation}
which performs a rigid body transform on the density over surface shape, defined by the contact model, so as to map it onto the test object's actual surface around $v$. To obtain $P(N^{(v)} | s, \cm^{\mathcal{C}}_l)$ the divergence of the relevant surface patch density $\om$ from the transformed contact model density $\hat{s}_{lk} \diamond \cm^{\mathcal{C}}_l$ is defined by Eq.\eqref{eq:contact-model-object-model-dist} and thus by Eq. \eqref{eq:distance_density_asym}. 
%
%
%
%

\begin{algorithm}[t]
\caption{Query density formation \label{alg:qd}}
    \begin{algorithmic}[1]
      \FORALL{samples $k=1$ to $N_{\qd^{\mathcal{C}}_l}$}
      \STATE sample $(\hat{v}_{k}, \hat{r}_{k}) \sim \om(v, r)$
      \STATE sample from conditional density $\hat{u}_{lk} \sim \cm^{\mathcal{C}}_l(u|\hat{r}_k)$
      	\STATE set $\hat{s}_{lk} = \hat{v}_{k} \circ \hat{u}_{lk}$
	\STATE set $w_{lk} = \exp\left( -\phi \, d_\text{dd}(\hat{s}_{lk} \diamond \cm^{\mathcal{C}}_l, \om)\right)$
	\STATE separate $\hat{s}_{lk}$ into position $\hat{p}_{lk}$ and quaternion $\hat{q}_{lk}$
	\ENDFOR
\STATE normalise weights $w_{lk}$ such that $\sum_k w_{lk} = 1$
\RETURN $\{(\hat{p}_{lk}, \hat{q}_{lk}, w_{lk})\}_{\forall{k=1..N_{\qd^\mathcal{C}_l}}}$
\end{algorithmic}
\end{algorithm}


As mentioned above, the query density is approximated using importance sampling. When a test object-view model, $\om(v,r)$, is presented a set of query densities $\qd^{\mathcal{C}}_l$ is calculated, one for each contact model prototype $\cm^{\mathcal{C}}_l$, $l=1\ldots N_\mathcal{C}$, according to \eqref{eq:qd.approx}. The algorithm proceeds as follows. Each $\qd^{\mathcal{C}}_l$ consists of $N_{\qd^{\mathcal{C}}_l}$ kernels centred on weighted hand-link poses:
\begin{equation}
\qd^{\mathcal{C}}_l(s) \simeq \sum^{N_{\qd^{\mathcal{C}}_l}}_{k=1} w_{lk} \mathcal{N}_3(p|{\hat{p}_{lk}}, \sigma_{p}) \Theta(q|{\hat{q}_{lk}}, \sigma_{q})
\label{eq:qd.approx}
\end{equation}
with $k$-th kernel centre $({\hat{p}_{lk}}, {\hat{q}_{lk}}) = \hat{s}_{lk}$, and $P(N^{(v)} | u, \cm^{\mathcal{C}}_l)$ gives the weight
\begin{equation}
w_{lk} \propto \exp\left( -\phi d_\text{dd}(\hat{s}_{lk} \diamond
\cm^{\mathcal{C}}_l, \om)\right),  \quad s.t. \sum_k w_{lk} = 1
\end{equation}
The sampling procedure is detailed in Algorithm~\ref{alg:qd}. First, a joint sample $(\hat{v}_k,\hat{r}_k)$ is taken from $\om$ (line 2), then $u_{lk}$ is sampled from $\cm^{\mathcal{C}}_l(u|\hat{r}_k)$ (line 3). This completely specifies a possible hand-link pose and curvature combination (line 4). Then the importance weight is calculated (line 5). The weights are normalised before the set of kernels is returned (lines 8-9).
%
%
%

The parameter $\phi$ was chosen empirically. The time complexity for computing contact query density $\qd^{\mathcal{C}}_l$ is $O(N_{\qd_l} N_{\mathcal{M}_l} (1 + \log(N_{\om})))$, where $N_{\qd_l}$ is the number of contact query kernels, $N_{\mathcal{M}_l}$ is the number of contact model kernels and $N_{\om}$ is the number of test object view kernels.

\subsection{Grasp Generation} 
Once query densities have been created for the new object for each contact model prototype, an initial set of grasps is generated (Figure~\ref{fig:flow}, Stage 3). Generation is by a series of random samples. We randomly sample, a grasp-view combination $(g,m)$ and then a hand-link $i$. This triple points to a contact-model cluster $\mathcal{C}_l$ and hence to a query density $\qd_l^{\mathcal{C}}$. A link pose $s_i$ is then sampled from that query density. Then a hand configuration $h_c$ is sampled from $\hc^{g}$. Together, the seed pose $s_i$ and the hand configuration $h_c$ define a complete grasp $h$, via forward kinematics, including the wrist pose $h_w$. This is an initial `seed' grasp, which will subsequently be refined. A large set $\mathcal{H}^1$ of such initial solutions is generated, where $h(j)=(h_w(j) , h_c(j))$ means the $j^{th}$ initial solution.

Having generated an initial solution set $\mathcal{H}^{1}$, stages of optimisation and selection are then interleaved to create a sequence of $K$ solution sets  $\mathcal{H}^{k}$ for $k=1 \ldots K$.


\subsection{Grasp Optimisation}
The final stage of the schema is optimisation of the candidate grasps (Figure~\ref{fig:flow}, Stage 4). The objective of grasp optimisation is, given a candidate equilibrium grasp and a reach to grasp model, to find a grasp that maximises the product of the likelihoods of the query densities and the hand configuration density
\begin{subequations}
\begin{align}
h^{*} & =  \argmax{h}  \mathcal{L}^{gm}(h) \\ 
         & = \argmax{h}  \mathcal{L}^g_\hc(h) \mathcal{L}^{gm}_\qd(h) \\
        & = \argmax{(h_w, h_c)}   \hc^{g}(h_c) \prod_{\qd_{l(i)}^{\mathcal{C}} \in \mathcal{Q}^{gm}} \qd_{l(i)}^{\mathcal{C}} \left(\mathrm{kin}_{i}^{\mathrm{for}}\left(h_w, h_c\right)\right)
\label{eq:grasping.product}
\end{align}
\end{subequations}
where ${\cal L}^{gm}(h)$ is the overall grasp likelihood and $\hc^g(h_c)$ is the hand configuration model~\eqref{eq:hc}. The query density $\qd_{l(i)}^{\mathcal{C}}$ is the query density for the cluster prototype $l(i)$ to which hand-link $i$ is mapped. The pose for hand-link $i$ is given by the forward kinematics of the hand, $\mathrm{kin}_{i}^{\mathrm{for}}\left(h_w, h_c\right)$. Finally, $\mathcal{Q}^{gm}$ is the set of instantiated query-densities for grasp-view pair $gm$.

Thus, whereas each initial grasp is generated using only a single query density, grasp optimisation requires evaluation of the grasp against all query densities. It is only in this improvement phase that all query densities must be used. Improvement is by simulated annealing (SA) \cite{kirkpatrick83optimizationby}. The SA temperature $T$ is declined linearly from $T_{1}$ to $T_{K}$ over the $K$ steps. In each time step $k$, one step of simulated annealing is applied to every grasp $h(j)$ in $\mathcal{H}^k$.

\begin{figure*}[t!]
\newcommand{\trainingw}{0.18\linewidth}
\includegraphics[width=\trainingw]{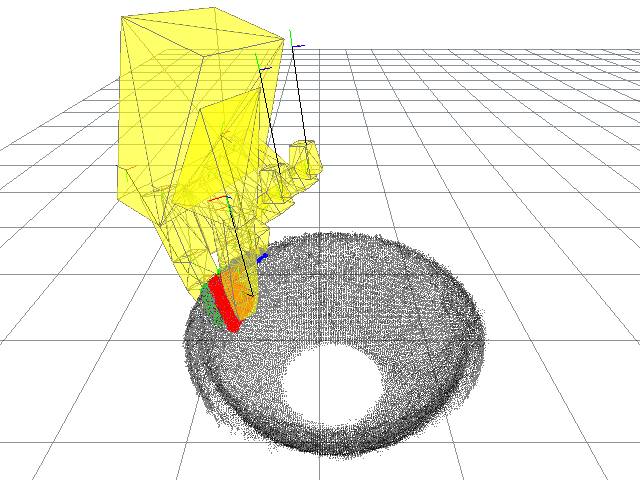}\hfill
\includegraphics[width=\trainingw]{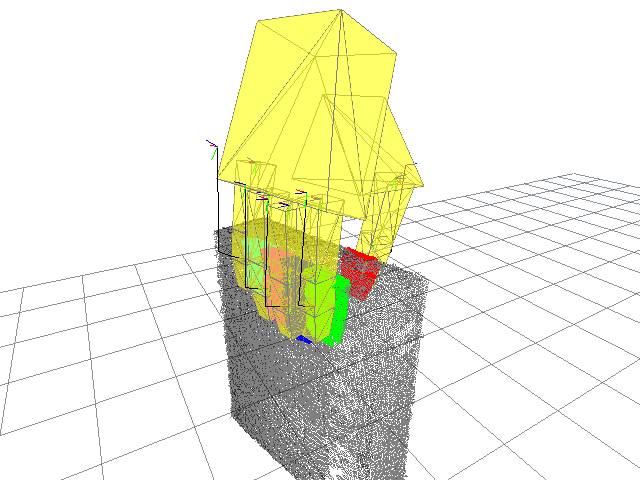}\hfill
\includegraphics[width=\trainingw]{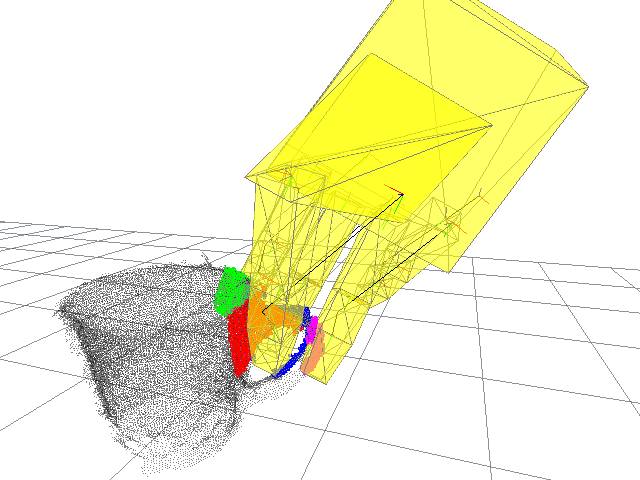}\hfill
\includegraphics[width=\trainingw]{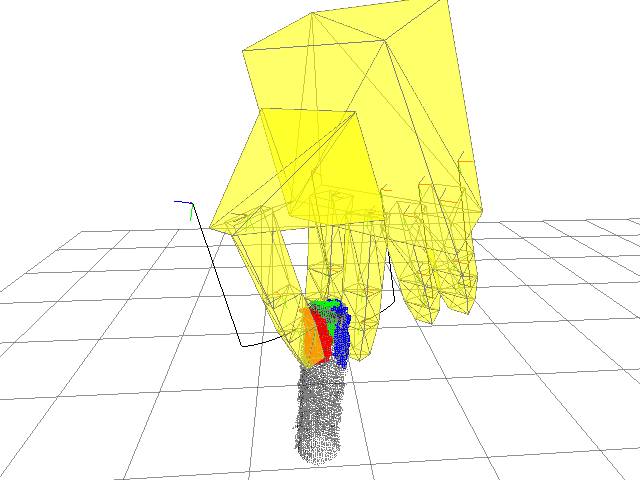}\hfill
\includegraphics[width=\trainingw]{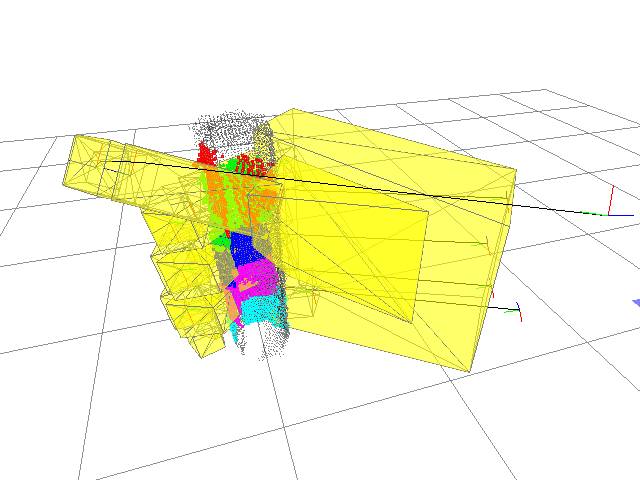}\hfill \\
\includegraphics[width=\trainingw]{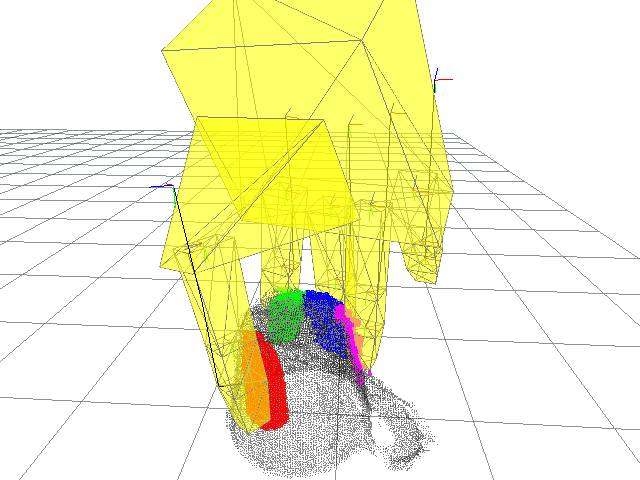}\hfill
\includegraphics[width=\trainingw]{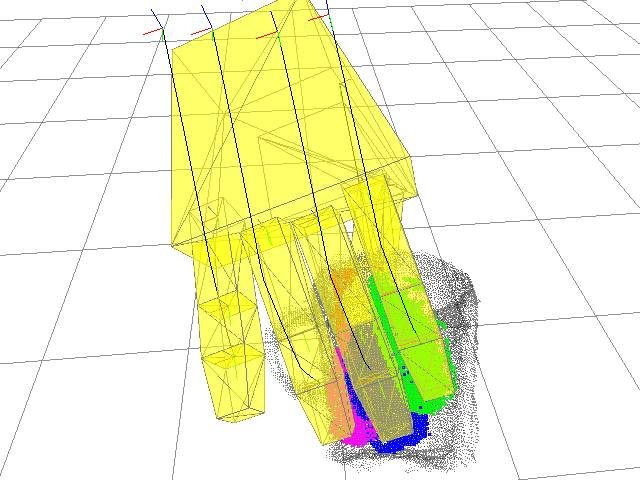}\hfill
\includegraphics[width=\trainingw]{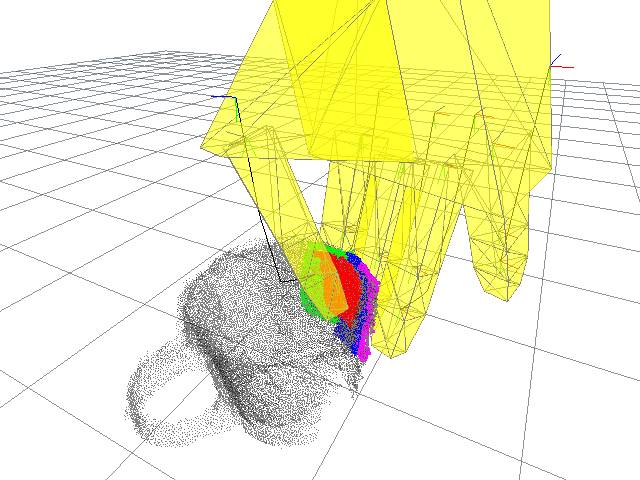}\hfill
\includegraphics[width=\trainingw]{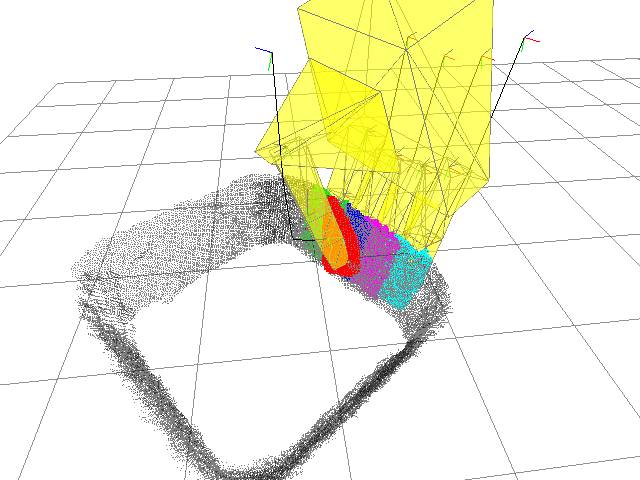}\hfill
\includegraphics[width=\trainingw]{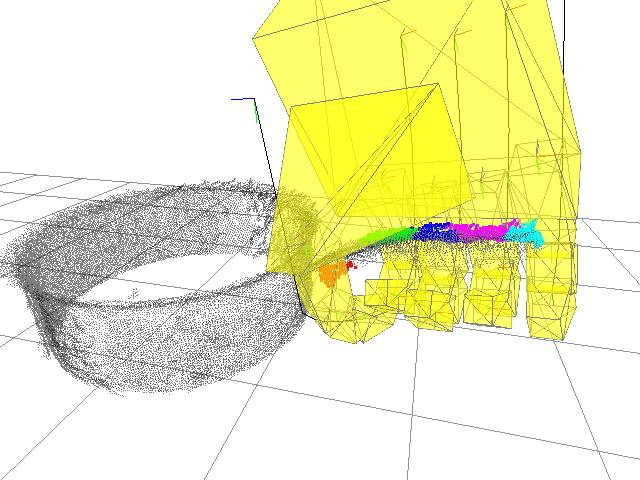}
\caption{The ten human demonstrated grasps. Top row (grasps from \cite{kopicki2015ijrr}), from left to right: {\em pinch with support}, {\em power-box}, {\em handle}, {\em pinch}, and {\em power-tube}. Bottom row (new training grasps), from left to right: {\em pinch-bottom}, {\em rim-side}, {\em rim}, {\em power-edge}, and {\em power-handle}. Top-row grasps were used in Experiment 1. Top-row and bottom-row grasps were used in Experiments 2 and 3. The grey lines show the sequence of finger tip poses on the demonstrated approach trajectory. The whole hand configuration is recorded for this whole approach trajectory. The initial pose and configuration we refer to as the pre-grasp position. For learning the contact models and the hand configuration model only the final hand pose (the yellow hand pose) is used. The point clouds are the result of registration of seven views with a wrist mounted depth camera taken during training. The training occurs with individual views. Coloured patches show contacts by finger, rather than individual hand-link.}
\label{fig:training}
\end{figure*}

\begin{figure*}[t]
\newcommand{\trainingw}{0.14\linewidth}
\includegraphics[width=\trainingw]{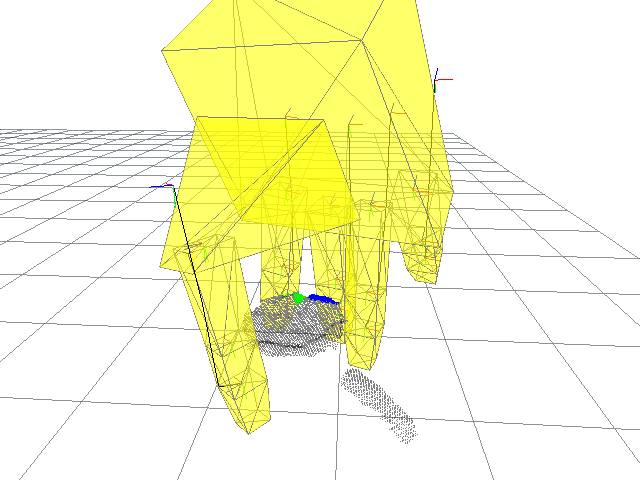}\hfill
\includegraphics[width=\trainingw]{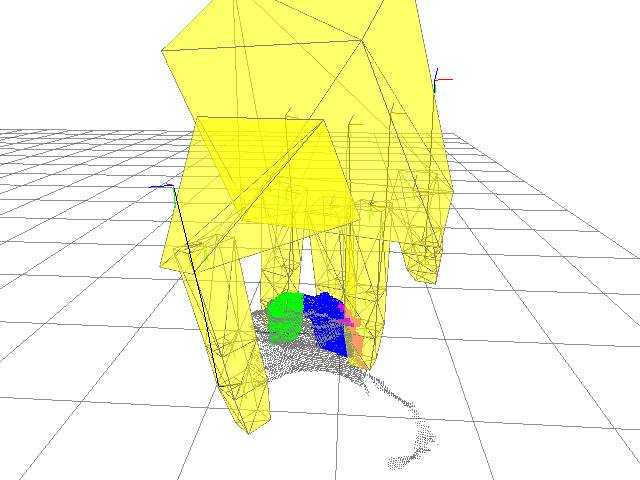}\hfill
\includegraphics[width=\trainingw]{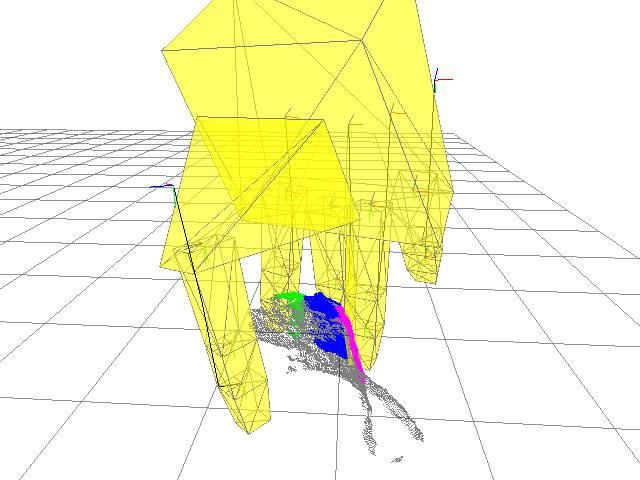}\hfill
\includegraphics[width=\trainingw]{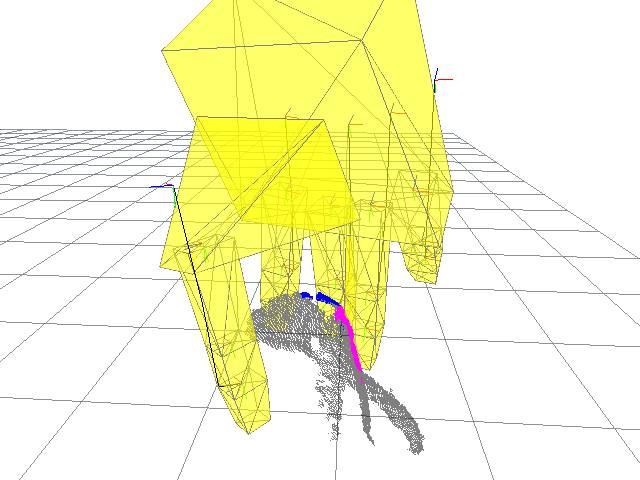}\hfill
\includegraphics[width=\trainingw]{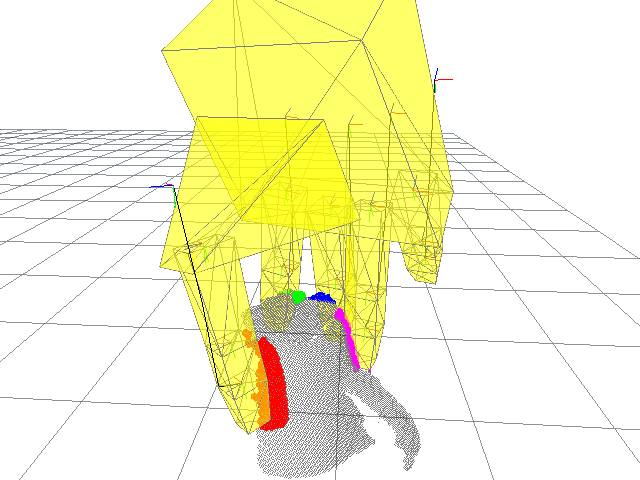}\hfill
\includegraphics[width=\trainingw]{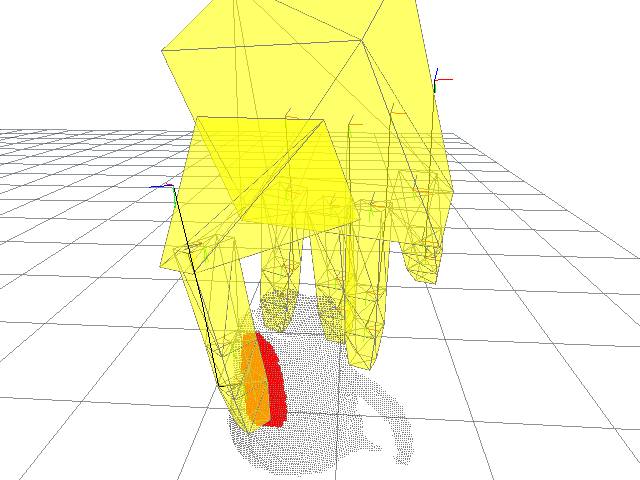}\hfill
\includegraphics[width=\trainingw]{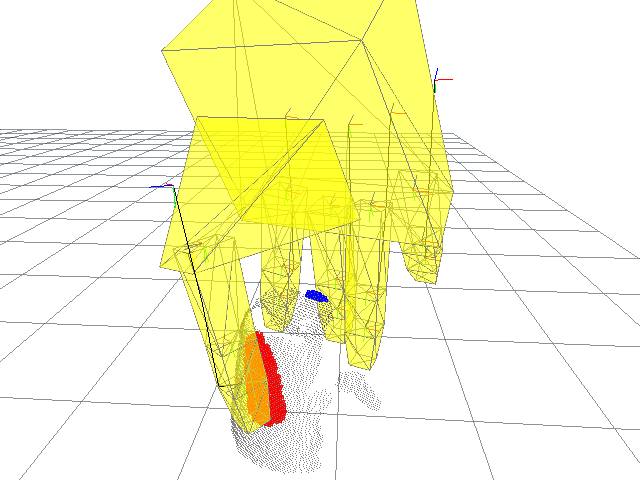}
\caption{Individual views, from which the view based models are trained, for the pinch-bottom grasp on an up-turned mug.}
\label{fig:view-training}
\end{figure*}
\subsection{Grasp Selection}
At predetermined selection steps (here steps 1 and 50 of annealing), grasps are ranked and only the most likely $10\%$ retained for further optimisation. During these selection steps the criterion in \eqref{eq:grasping.product} is augmented with an additional expert $\coll(h_w, h_c)$ penalising collisions for the entire reach to grasp trajectory in a soft manner. This soft collision expert has a cost that rises exponentially with the greatest degree of penetration through the object point cloud by any of the hand links. We thus refine \eq\ref{eq:grasping.product}:
\begin{subequations}
\begin{align}\label{eq:grasping.likelihood}
{\cal L}^{gm}(h) & =  {\cal L}_\coll(h) {\cal L}^g_\hc(h) {\cal L}^{gm}_\qd(h) \\
 & =  \coll(h_w, h_c) \hc^g(h_c)  \! \! \! \! \prod_{\qd_{l(i)}^{\mathcal{C}} \in \mathcal{Q}^{gm}} \! \! \! \! \qd_{l(i)}^{\mathcal{C}} \left(\mathrm{kin}_{i}^{\mathrm{for}}\left(h_w, h_c\right)\right)
\end{align}
\end{subequations}

where ${\cal L}^{gm}(h)$ is now factorised into three parts, which evaluate the collision, hand configuration and query density experts, all at a given hand pose $h$. A final refinement of the selection criterion is due to the fact that the number of hand-links in contact during a grasp varies across grasps and views. Thus the number of query densities $N^{gm}_\qd$, $N^{g'm'}_\qd$ also varies, and so the values of ${\cal L}^{gm}$ and ${\cal L}^{g'm'}$ cannot be compared directly. Given the grasp with the maximum number of involved links $N^{\max}_\qd$, we therefore normalise the likelihood value~\eqref{eq:grasping.likelihood} with
\begin{equation}
\left\|{\cal L}^{gm}(h)\right\| = {\cal L}_\coll(h) {\cal L}^g_\hc(h) \left({\cal L}^{gm}_\qd(h)\right)^{\frac{N^{\max}_\qd}{N^{gm}_\qd}}.
\label{eq:grasping.likelihood.norm}
\end{equation}
It is this normalised likelihood $\left\|{\cal L}^{gm}\right\|$ that is used to rank all the generated grasps across all the grasp-view pairs during selection steps. After simulated annealing has yielded a ranked list of optimised grasp poses, they are checked for reachability given other objects in the workspace, and unreachable poses are pruned. 

\subsection{Grasp Execution}
The remaining best scoring hand pose $h^{*}$, evaluated with respect to \eqref{eq:grasping.likelihood.norm}, is then used to generate a reach to grasp trajectory.  This is the command sequence that is executed on the robot.

\label{sec:results}
\section{Experimental Study}

This section is structured as follows. First, the creation of  challenging data set is described. Second, the algorithmic variants tested are enumerated. Then, three experiments are presented in turn. Each of these varies in the size of the training set. Experiment 1 trains with five grasps and Experiment 2 with ten. Experiment 3 introduces training from robot generated grasps. A discussion examines each of the hypotheses in the light of the results.

\subsection{Test set creation}
In preparation for the experimental evaluation, a challenging test set was created. This used 40 novel test objects (Figure~\ref{fig:objects}). The test cases were object-pose pairs relative to a single, fixed viewpoint. The object-pose combinations were chosen to be particularly challenging by using a pose that reduced the typical surface recovery from the fixed view. Some objects were employed in several poses, yielding a total of 49 object-pose pairs. Because both object and pose are controlled this means that we can test algorithms using a paired comparisons statistical methodology. This can yield statistically significant results for small numbers of test grasps.

The question arises as to whether it can be validated that this new data-set is indeed challenging. A previous single-view data set was also generated in \citet{kopicki2015ijrr}, using many of the same objects, but without deliberately challenging poses. The performance of the algorithm presented in \citet{kopicki2015ijrr} on this original single view test set was 77.8\% (35/45). Therefore, to verify the challenging nature of the new test data-set, that algorithm was tested on the new single-view dataset. To make the two test-sets comparable the same set of five training grasps, presented in \citet{kopicki2015ijrr}, was used.
Testing on the new dataset the performance of the algorithm reduced to 59.2\% (29/49). Since we hypothesized that the harder data set should produce a lower success rate we applied a one-tailed $\chi^2$ test, using Fisher's exact test, which gave a p-value of 0.043. Thus the difference between the success rates is statistically significant and is unlikely to have been caused by chance. We therefore accept the hypothesis that the new data set of object-pose pairs is more challenging than the previous data set.

\subsection{Algorithmic variations}

\begin{table}[t]
\begin{center}
\caption{Algorithmic variations for Experiment 2 \label{tab:algs}}
\begin{tabular}{|c|c|c|c|c|} \hline
Alg & View-based & New Eval & Merging & Features  \\ \hline
A1 & No  & No  & No  & Curv   \\
A2 & Yes & No  & No  & Curv   \\
A3 & Yes & Yes  & No & Curv   \\
A4 & Yes & Yes & Yes & Curv    \\
A5 & Yes & Yes & No & FPFH \\
A6 & Yes & Yes & Yes & FPFH \\
\hline
\end{tabular}
\end{center}
\end{table}

The paper has presented three main innovations. These are: (i) a view based representation; (ii) a method for merging contact models; (iii) a new evaluation method for calculating the likelihood of a generated grasp. In addition, the surface descriptor may be either principal curvatures or fast point feature histograms. It is clearly desirable to evaluate which of these innovations is most effective, and to study how well they work in combination. The sixteen possible combinations, however, are too many to evaluate properly on a real robot. Therefore, six different combinations were tried, each one introducing a new innovation on top of the others. These six `algorithmic variations' are listed in Table~\ref{tab:algs}. The algorithm reported in \citet{kopicki2015ijrr} is variation A1. Note that variants A1 and A2 are not to be confused with Algorithm 1 (contact model clustering) and Algorithm 2 (query density computation), which are components of all variants.

The algorithms presented depend on a number of parameters, which have been presented earlier in the text. It would not be possible to systematically tune these parameters using grid search, but a small number of informal experiments were used to select the values used here. The same parameter settings were used for all algorithmic variants. It is entirely possible that better settings exist. The values used are presented in Table~\ref{tab:params}.

\begin{table}
\begin{center}
\caption{Parameters of the grasp learning and inference algorithms.\label{tab:params}}
\begin{tabular}{|c|} \hline
Receptive field \\ 
$\delta= 0.01, \lambda=50.0$ \\ \hline
Contact model (curvature) \\
$\sigma_r=10, \sigma_p=0.005$ (linear), $\sigma_q=0.5$ (angular) \\ \hline
Contact model selection \\
$\eta=0.2, \zeta=3$ \\ \hline
Clustering Contact Models \\
$\xi=1.0, w_{lin}=1.0, w_{ang}=0.01$ \\ \hline
Hand Configuration Model \\
$N_C=1000$ (number of kernels), $\alpha=100.0, \beta=1.0$ \\ \hline
Query density \\
$N_Q=5000$ (number of kernels), $\phi=1.0$ \\ \hline
Grasp Generation \\
$\mathcal{H}^1=50000$ (number of initial solutions) \\
$K=500$, selection steps are at $k=1$, $k=50$\\
$T_1=0.005, T_{500}=0.05$ \\ \hline
\end{tabular}
\end{center}
\end{table}

\subsection{Experiment 1}
This experiment tests the hypothesis that, even without an enlarged set of training grasps, the combined innovations, present in A4 and A6, will improve the grasp success rate. Thus, variations A1, A4 and A6 were trained with the five grasps from~\citet{kopicki2015ijrr}. A paired comparisons experiment with all 49 test cases was executed. This led to a grasp success rate of 59.2\% (as reported above) for A1, a success rate of 75.5\% (37/49) for A6, and 77.6\% (38/49) for A4. Although the success rates for both A4 and A6 are higher than that for A1, using the two-tailed McNemar test for the difference between two proportions on paired data these differences are not statistically significant (p=0.1175 for A6:A1, p=0.0665 for A4:A1). 

\subsection{Experiment 2}
This experiment tests hypotheses H2 and H3. H2 is the hypothesis that view-based grasp modelling enables better generation of grasps for thick objects. H3 is the hypothesis that the success rate progressively increases as innovations 1 to 3 are added. Experiment 2 also provides evidence to test hypotheses H4 and H5, that the grasp success rate will improve as training data is added, and do so faster if all innovations are deloyed. The training set was therefore increased to 10 grasps. Figure~\ref{fig:objects} shows the objects used for training. To test H2 and H3, six algorithmic variations A1:A6 were tested, as detailed in Table~\ref{tab:algs}. These progressively add the three innovations. Algorithm A1 is the version described in \cite{kopicki2015ijrr}. A2 is A1 plus view based organisation of grasp models. Algorithm A3 is A2 plus improved evaluation of grasp likelihood by density comparison. Algorithm A4 is A3 plus contact model merging. All of algorithms A1-A4 use principal curvatures as the features. As a final step, we also test the robustness of the method to changes in the surface descriptors used. Variants A5 and A6 are the equivalents of A3 and A4 respectively, but use FPFH as the surface descriptor instead of curvatures. 

The success rates for each algorithmic variation are shown in Table~\ref{tab:exp2}. As the innovations are added the success rate rises. 
Applying McNemar's test for the difference in proportions for paired data, A4 and A6 dominate A1 and A2, and these differences are highly statistically significant. A full table of p-values is shown in Table~\ref{tab:pvalues}.

\begin{table}
\begin{center}
\caption{Experiments 1, 2 and 3: Grasp success rates for algorithm variations A1 (Vanilla) to A6, and for A1+AT and A4+AT (Autonomous Training). Numbers in brackets indicate the number of training examples used. \label{tab:exp2}}
\begin{tabular}{|c|c|c|c|c|c|} \hline
Alg & \# succ & \% succ & Alg & \# succ & \% succ \\ \hline
A1(5)   & 29  & 59.2 & A4(5)   & 38  & 77.6 \\
A6(5)   & 37  & 75.5 & A1(10) &  27 & 55.1 \\
A2(10) &  28 & 57.1 & A3(10) &  34 & 69.4 \\
A4(10) &  40 & 81.6 & A5(10) &  35 & 71.4 \\
A6(10) &  40 & 81.6 & A1+AT &  31 & 63.3 \\
A4+AT & 43 & {\bf 87.8} & & & \\
\hline
\end{tabular}
\end{center}
\end{table}

\begin{table}
\begin{center}
\caption{p-values for statistically significant pairwise differences between algorithms for Experiments 1, 2 and 3. Stronger algorithms are on the left. Format is Alg(X), where X is the number of training examples, Alg+AT means autonomous training.\label{tab:pvalues}}
\begin{tabular}{|c|c|c|c|} \hline
Alg pair & p-value & Alg pair & p-value \\ \hline
 A4(5):A1(10)     & 0.0153 &  A4(10):A1(5) & 0.0218 \\ \hline
 A6(5):A1(10)     & 0.0162 & A6(10):A1(5) & 0.0153 \\ 
 \hline
 A4(10):A1(10) & 0.0009 & A4+AT:A1(10)&  0.0002 \\ \hline
 A6(10):A1(10) & 0.0036 &  A4+AT:A2(10)&  0.0013 \\ \hline
 A4(10):A2(10)  & 0.0033 &  A4+AT:A3(10)&  0.0265 \\ \hline
 A6(10):A2(10)   & 0.0033 & A4+AT:A5(10) &  0.0433 \\  \hline
 A4+AT:A1(5)    & 0.0056  & A4+AT:A1+AT & 0.0095 \\
\hline
\end{tabular}
\end{center}
\end{table}

\begin{figure*}
\newcommand{\testingw}{0.12\linewidth}
\newcommand{\leftcrop}{12cm}
\newcommand{\botcrop}{5cm}
\newcommand{\rightcrop}{12cm}
\newcommand{\topcrop}{10cm}
\mbox{\bf \large \hspace{1.4cm} A1+AT \hspace{2.85cm} A4+AT \hspace{2.85cm} A1+AT \hspace{2.85cm} A4+AT }
\mbox{\bf \large \hspace{1.2cm} (Failures) \hspace{2.3cm} (Successes) \hspace{2.3cm} (Failures) \hspace{2.3cm} (Successes) } \\
\includegraphics[width=\testingw,trim={12cm 5cm 13cm 10cm},clip]{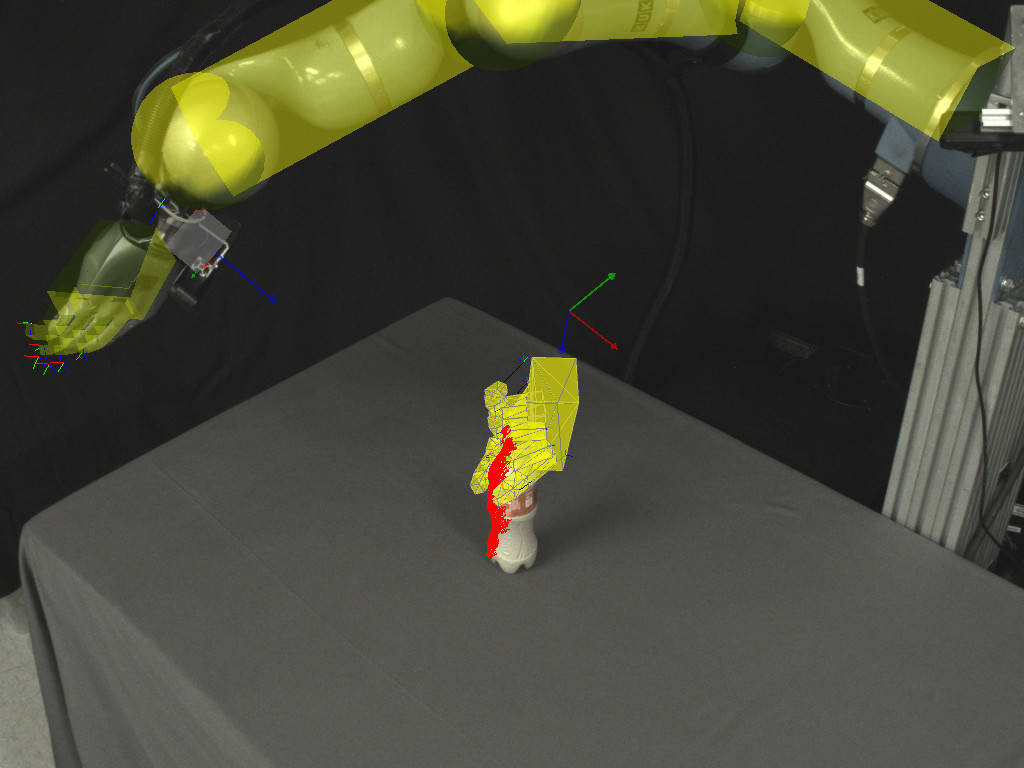} 
\includegraphics[width=\testingw,trim={12cm 5cm 13cm 10cm},clip]{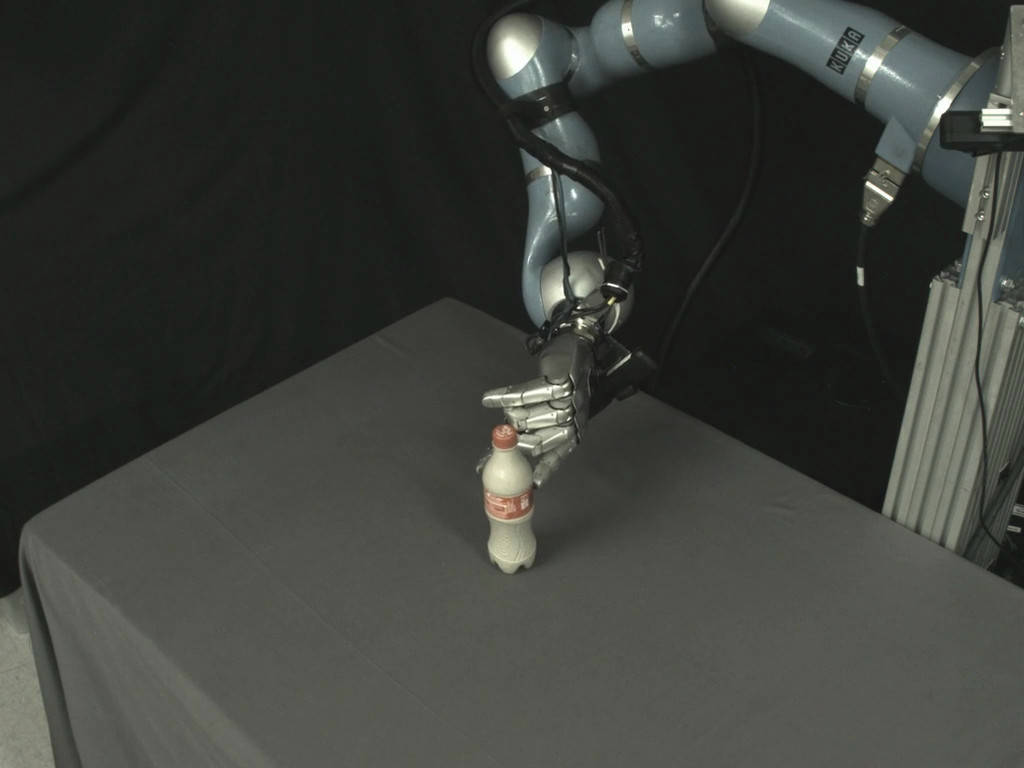}
\includegraphics[width=\testingw,trim={12cm 5cm 13cm 10cm},clip]{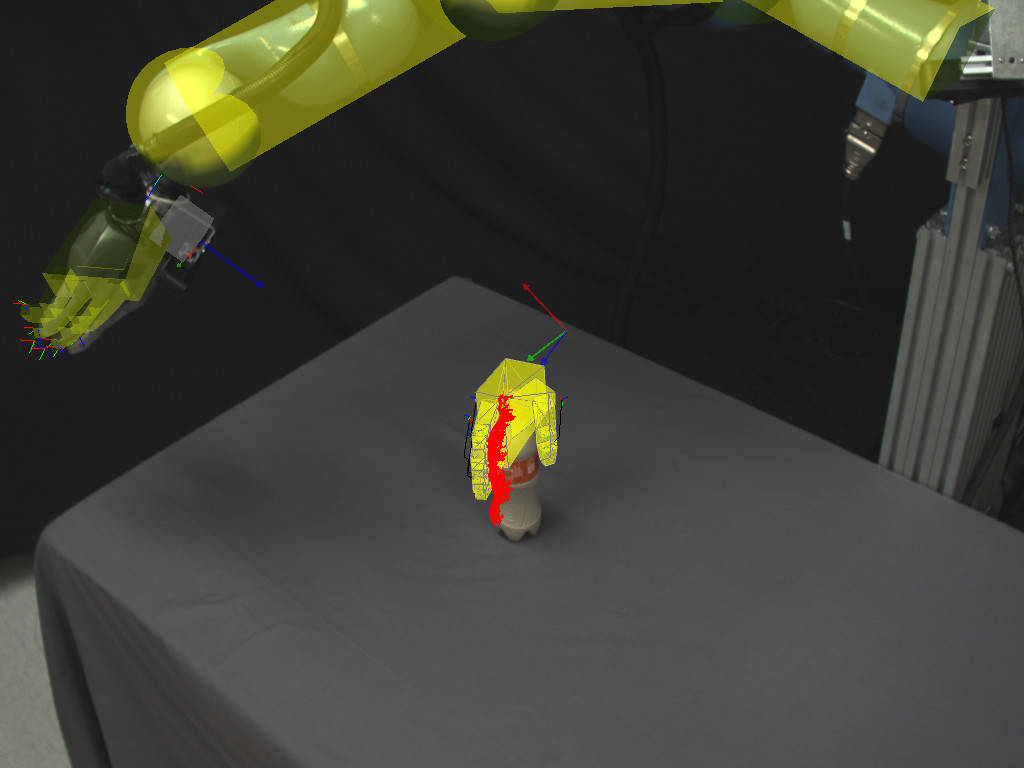}
\includegraphics[width=\testingw,trim={23.3cm 9.6cm 23.8cm 18.77cm},clip]{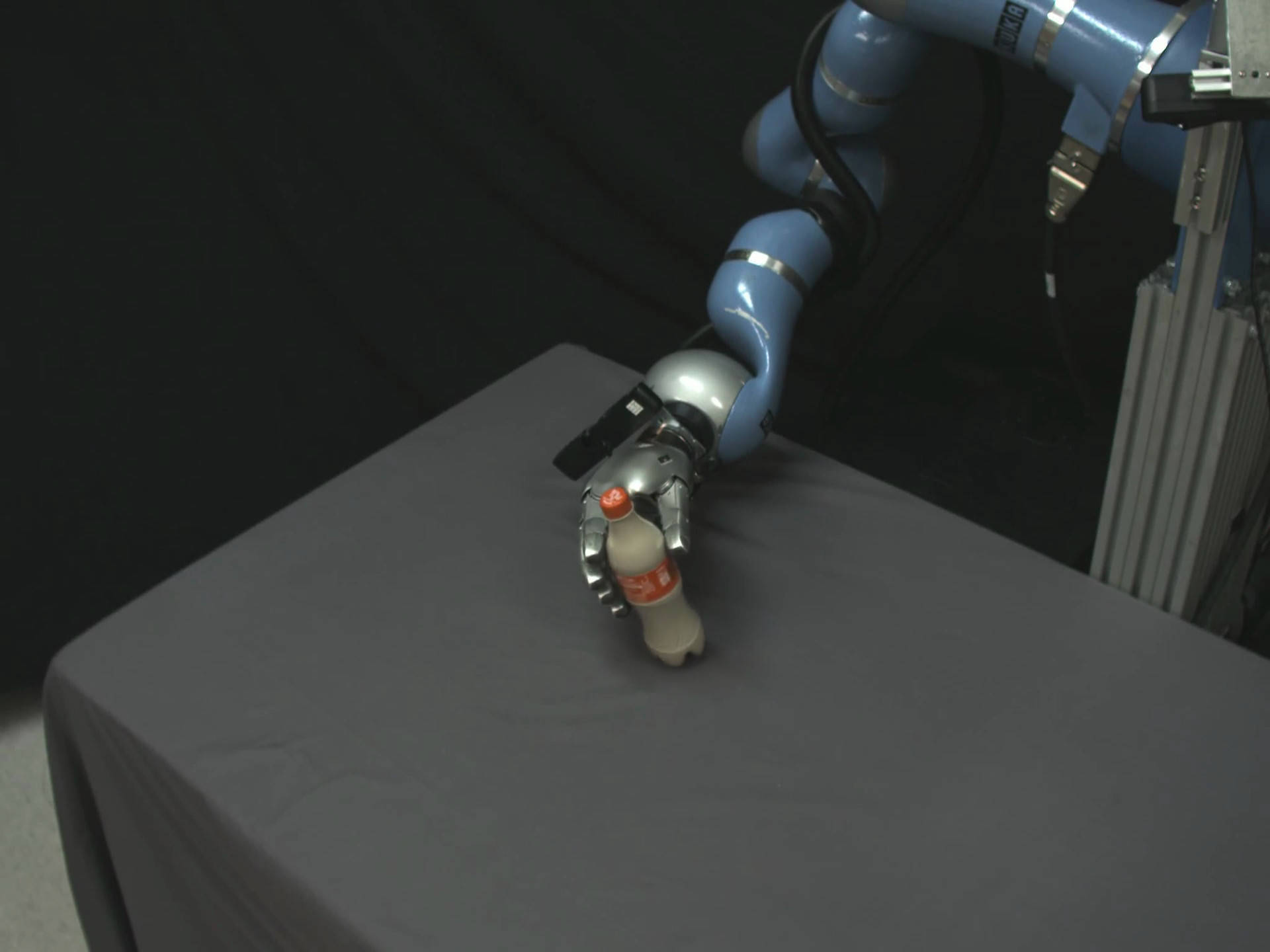}
 \includegraphics[width=\testingw,trim={12cm 5cm 13cm 10cm},clip]{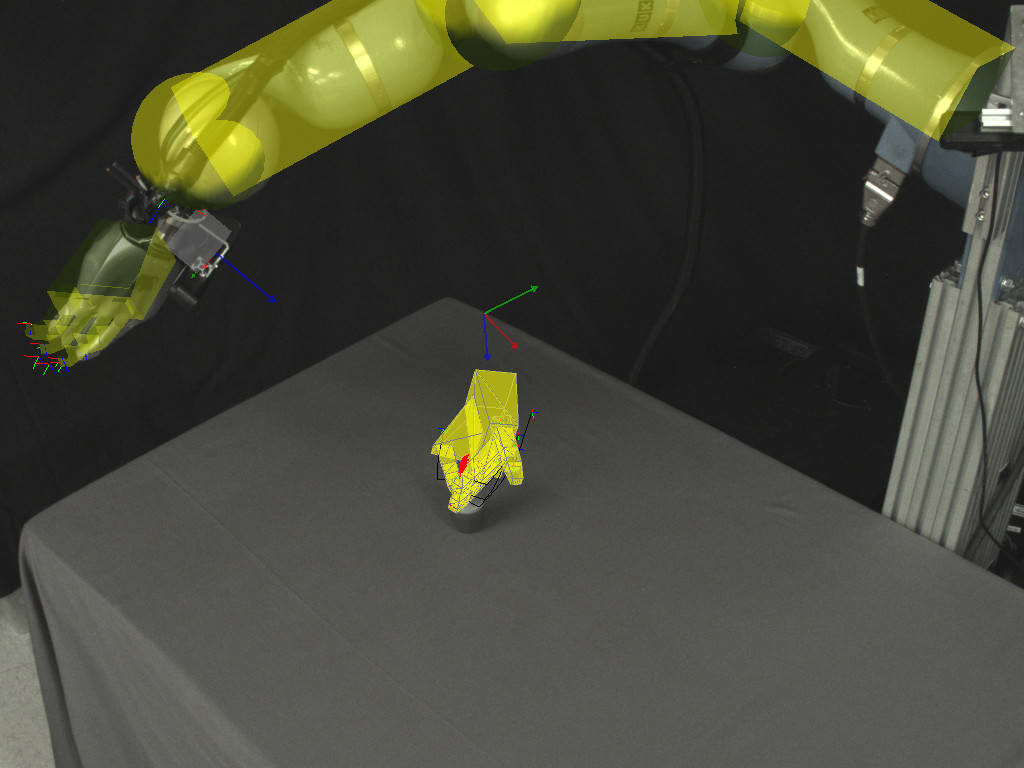}
\includegraphics[width=\testingw,trim={12cm 5cm 13cm 10cm},clip]{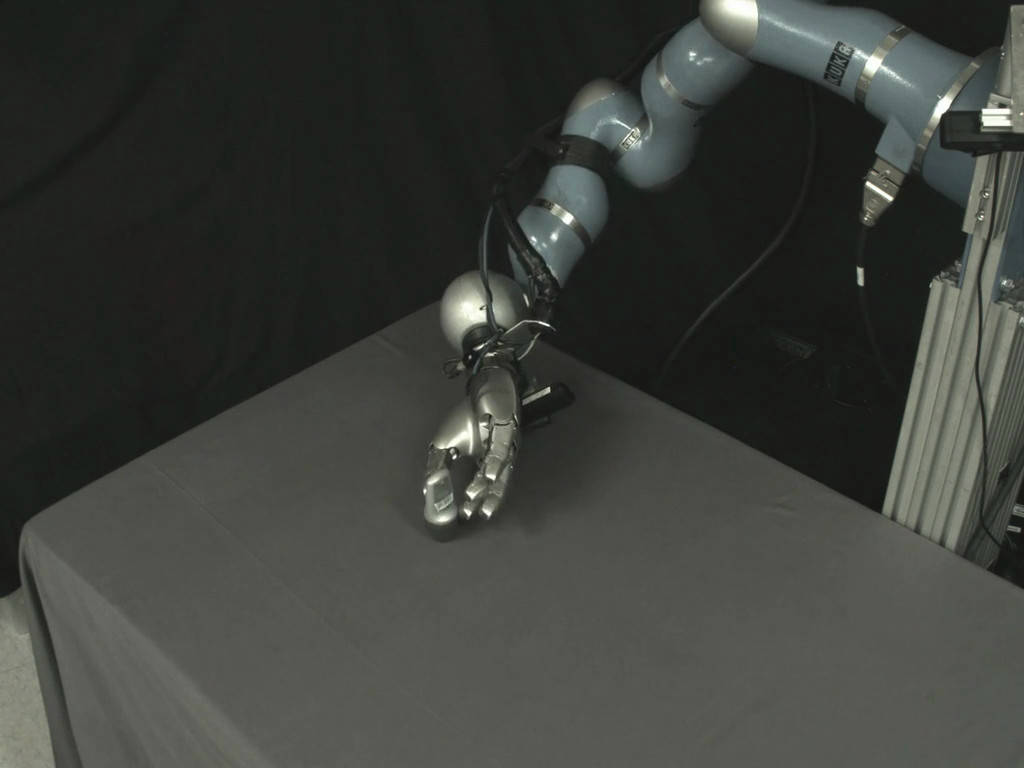}
\includegraphics[width=\testingw,trim={12cm 5cm 13cm 10cm},clip]{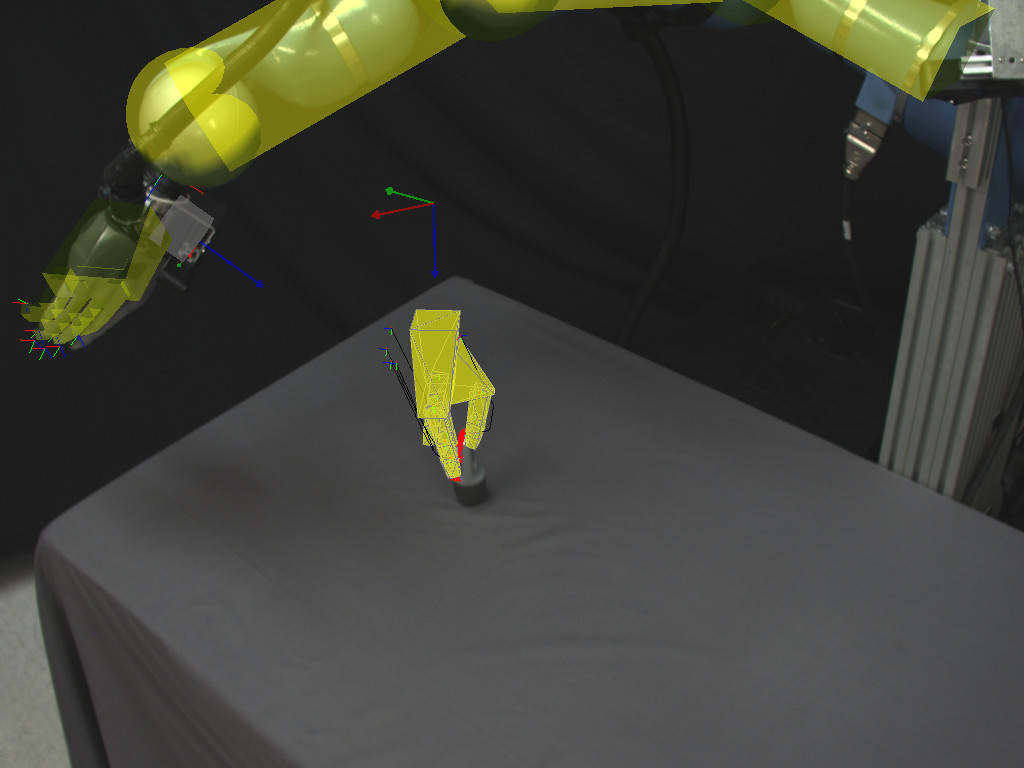}
\includegraphics[width=\testingw,trim={23.3cm 9.6cm 23.8cm 18.77cm},clip]{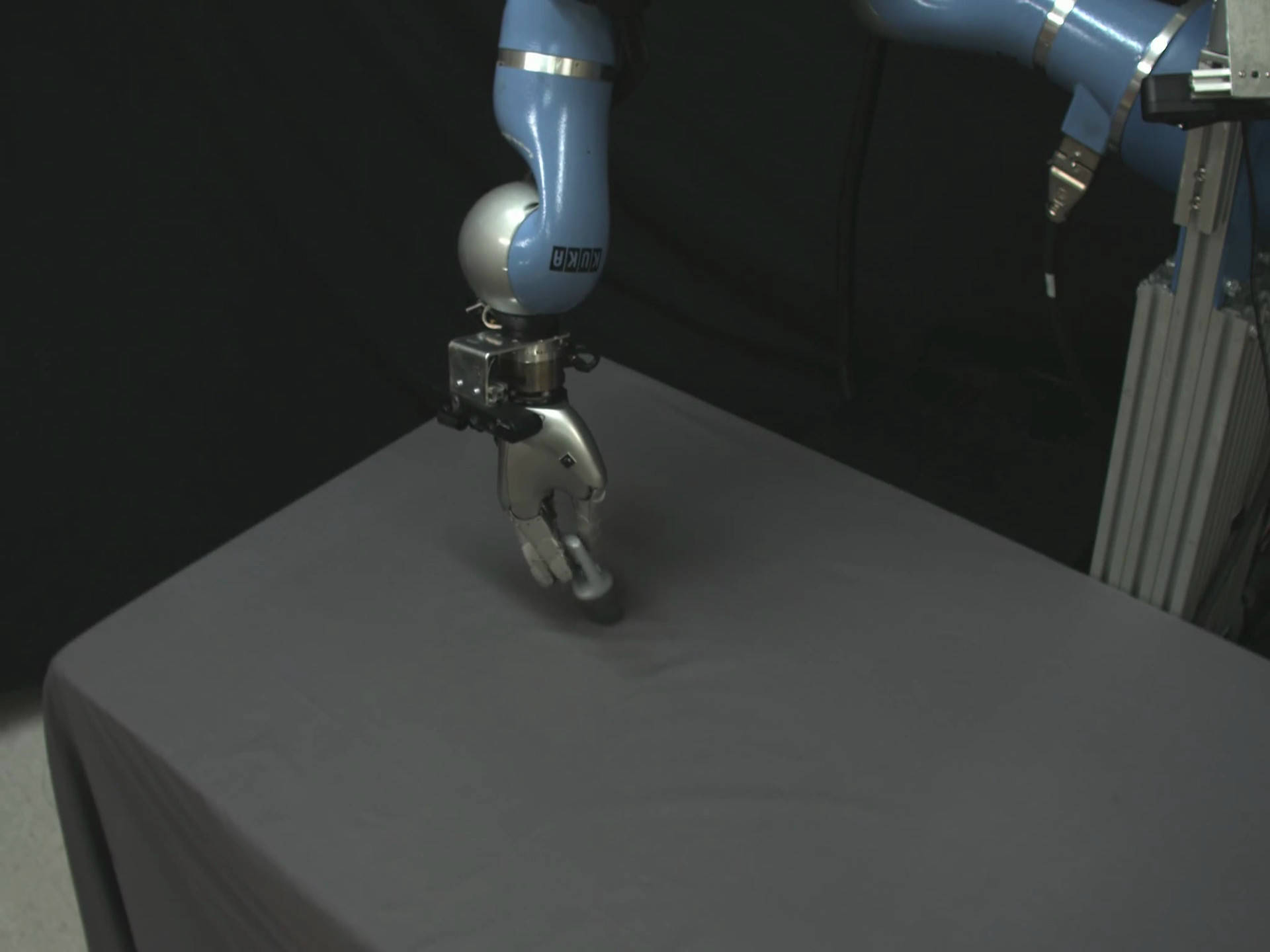} \\
\includegraphics[width=\testingw,trim={10cm 5cm 15cm 10cm},clip]{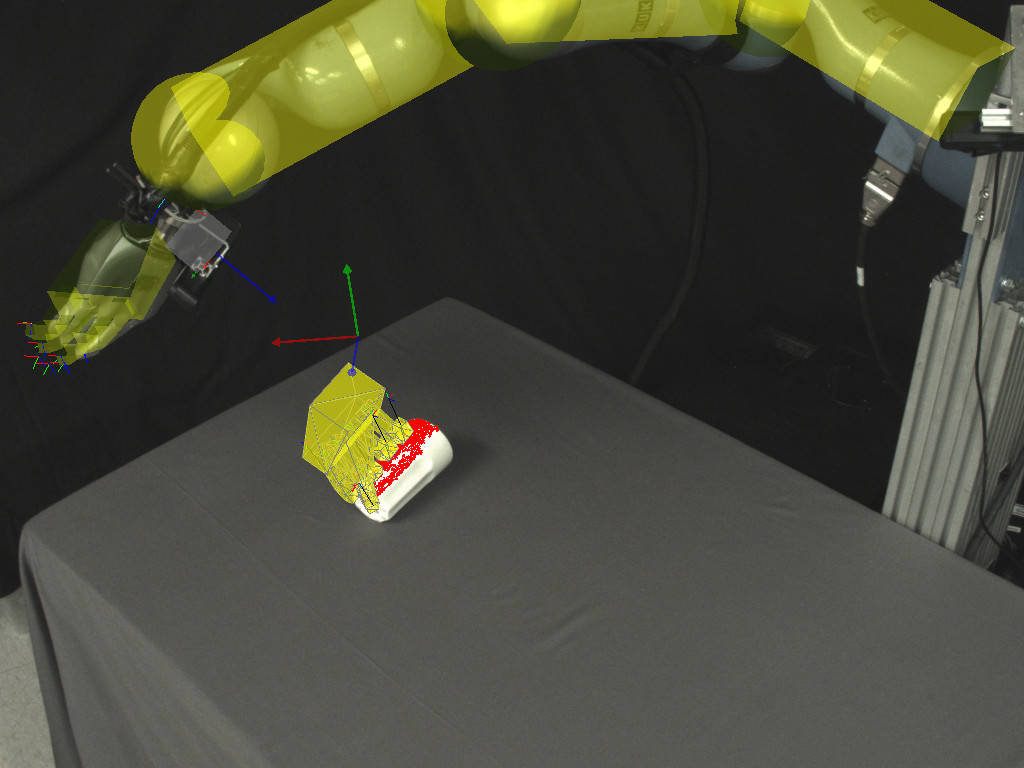} 
\includegraphics[width=\testingw,trim={10cm 5cm 15cm 10cm},clip]{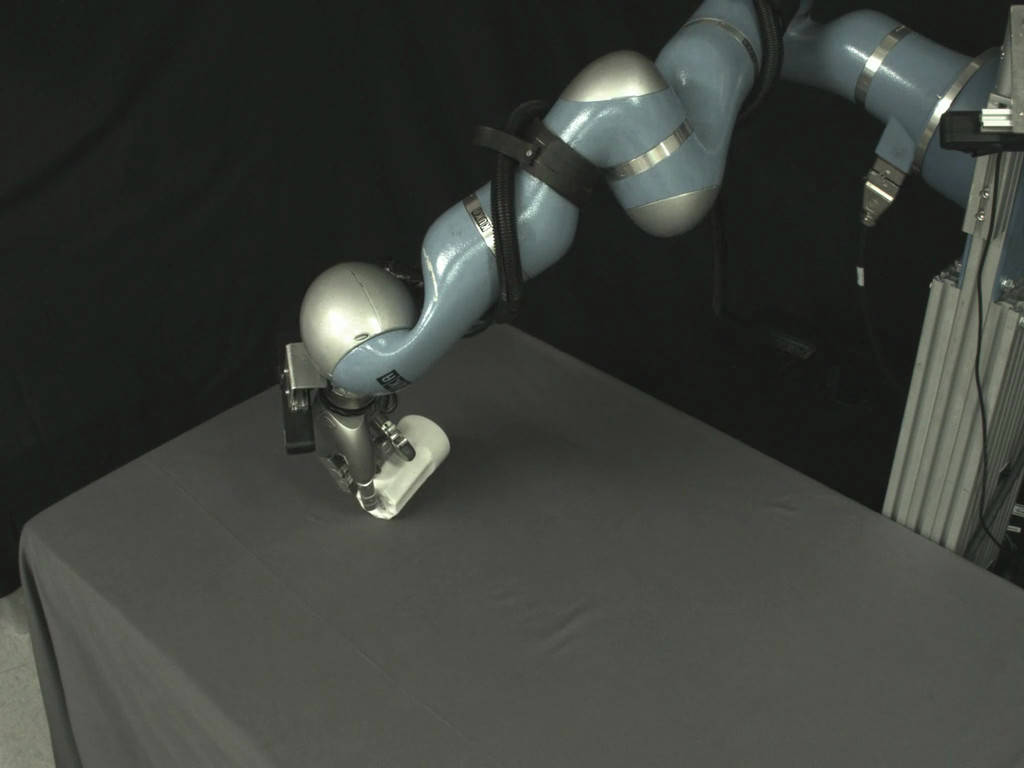}
\includegraphics[width=\testingw,trim={10cm 5cm 15cm 10cm},clip]{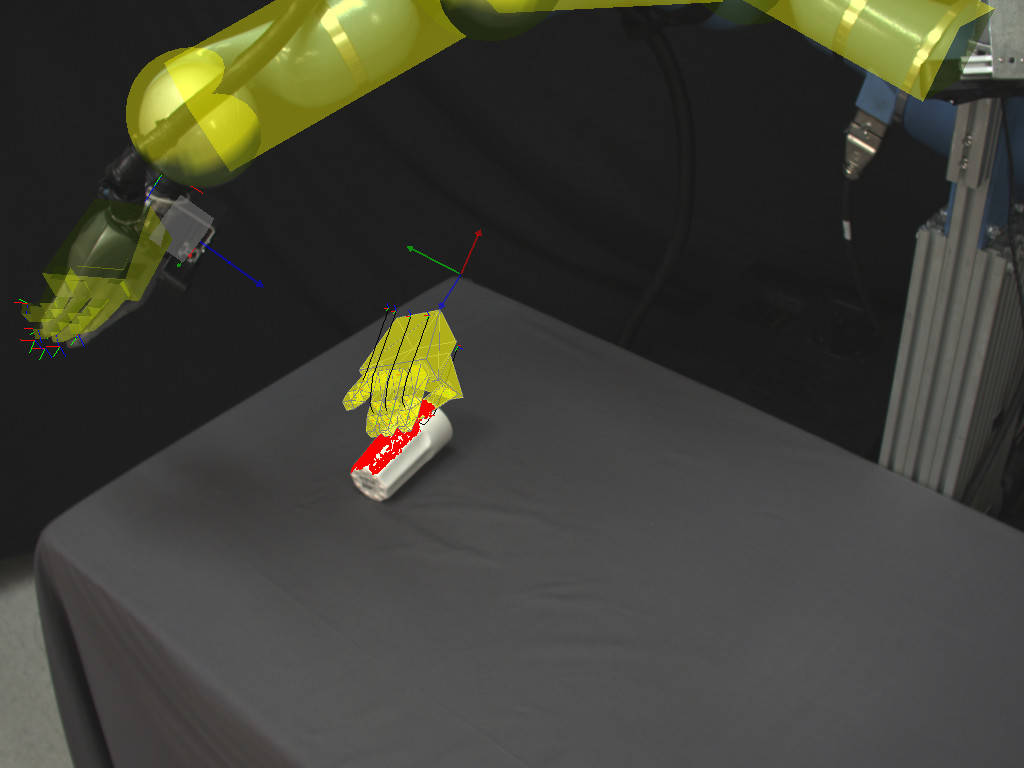} 
\includegraphics[width=\testingw,trim={19.3cm 9.6cm 27.8cm 18.77cm},clip]{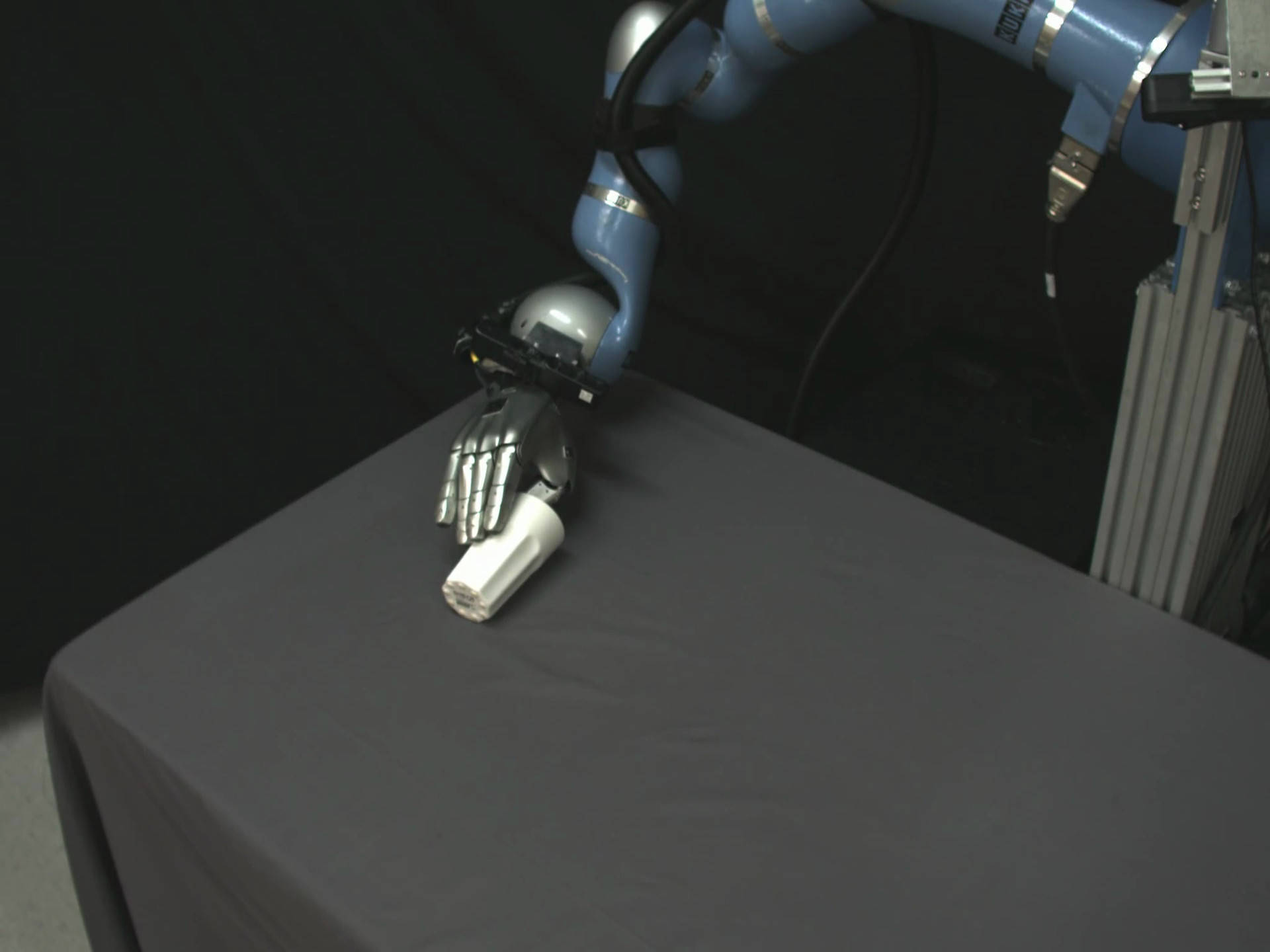}
\includegraphics[width=\testingw,trim={12cm 5cm 13cm 10cm},clip]{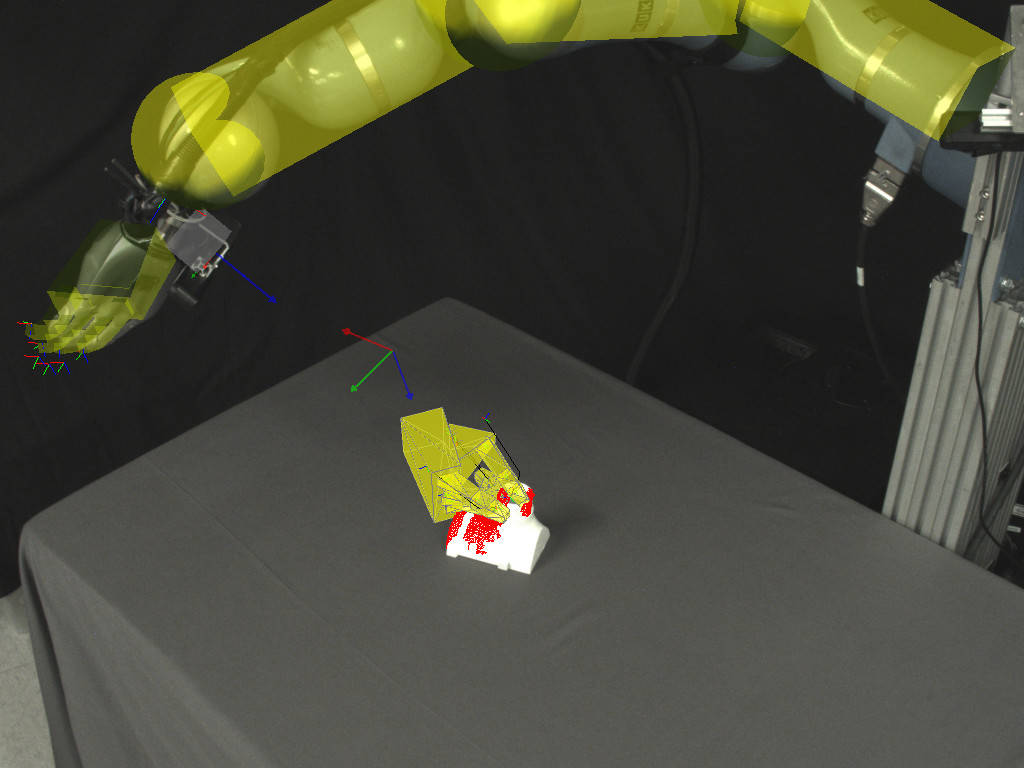} 
\includegraphics[width=\testingw,trim={12cm 5cm 13cm 10cm},clip]{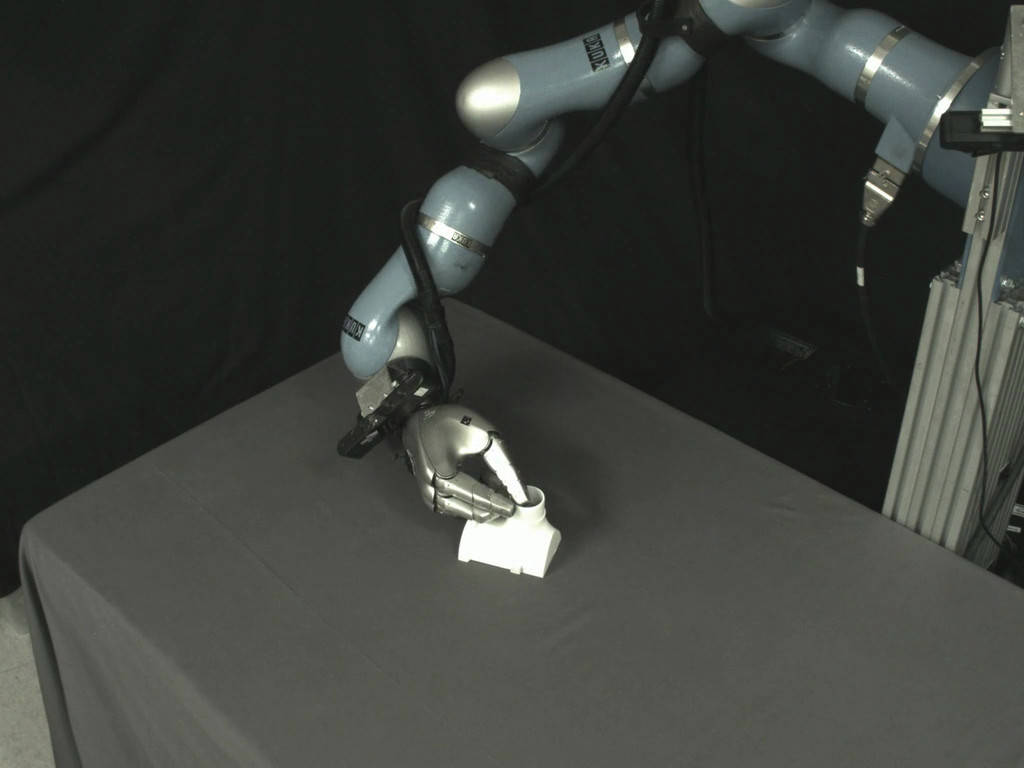}
\includegraphics[width=\testingw,trim={12cm 5cm 13cm 10cm},clip]{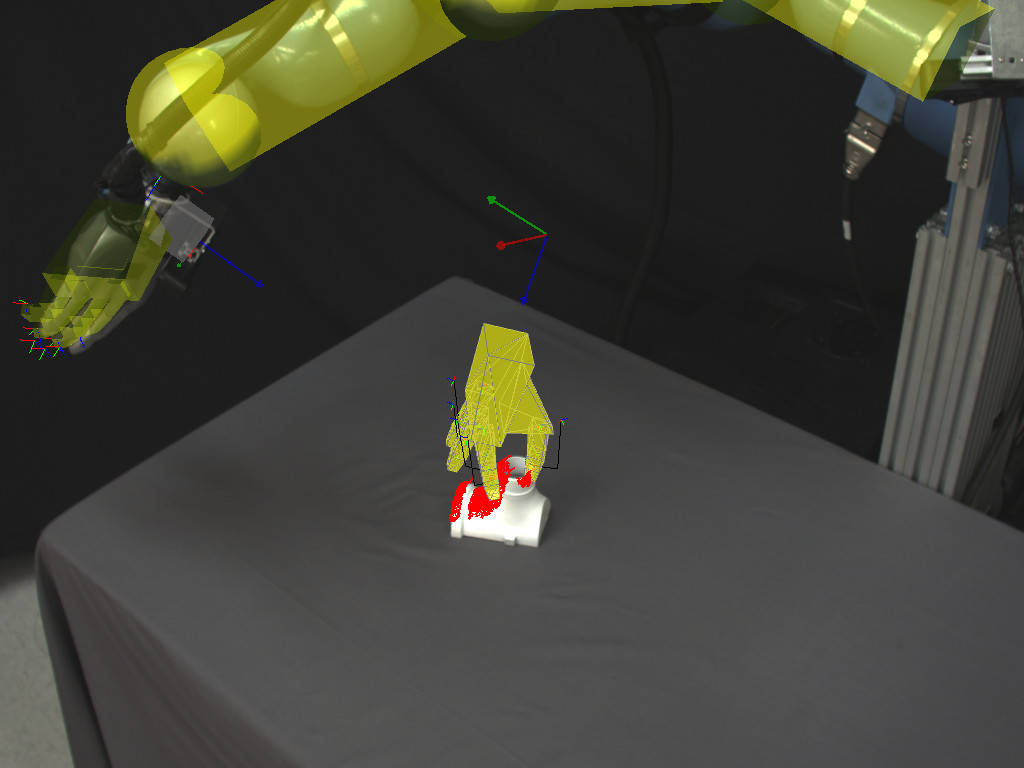}
\includegraphics[width=\testingw,trim={23.3cm 9.6cm 23.8cm 18.77cm},clip]{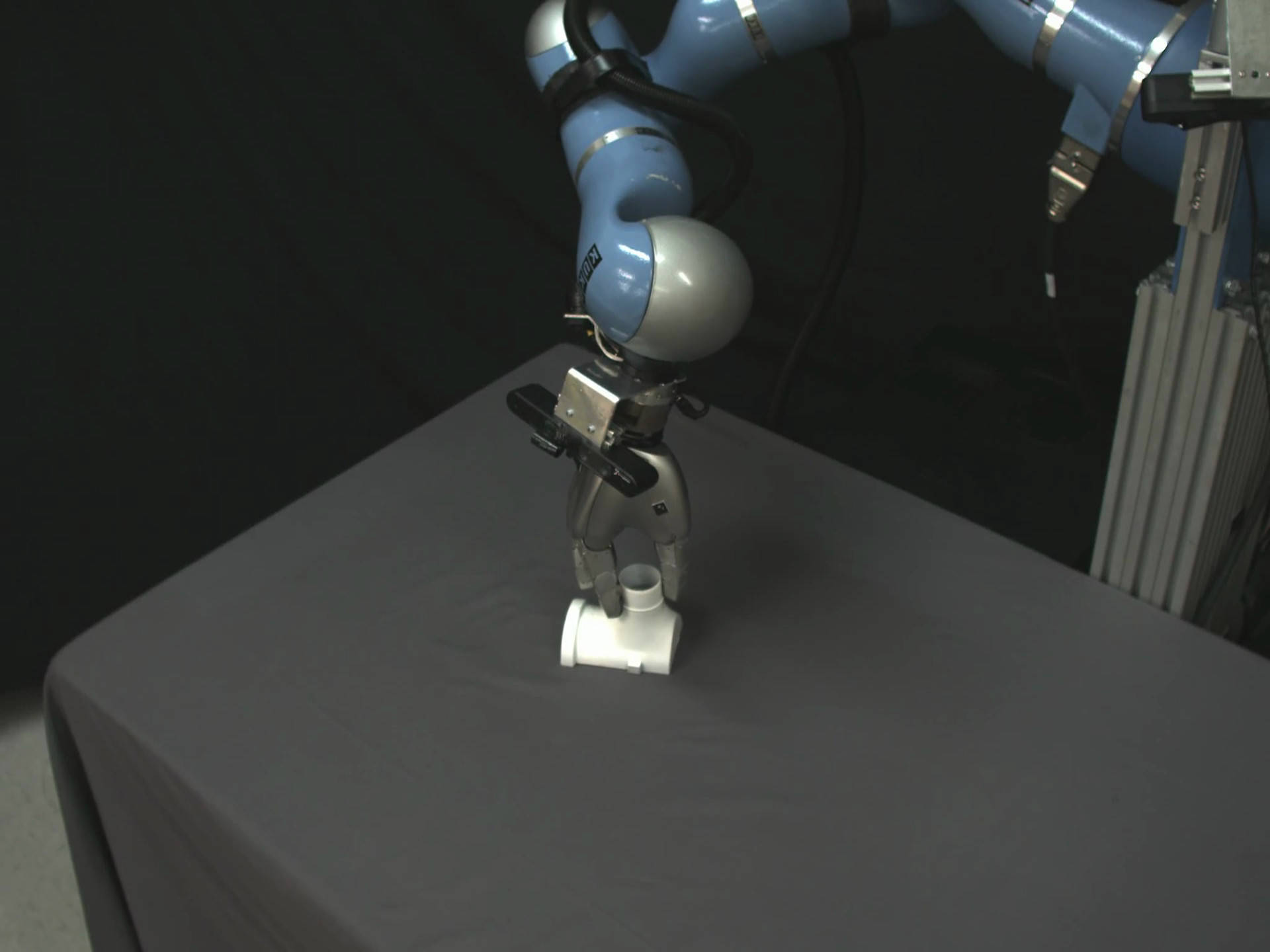} \\
\includegraphics[width=\testingw,trim={10cm 5cm 15cm 10cm},clip]{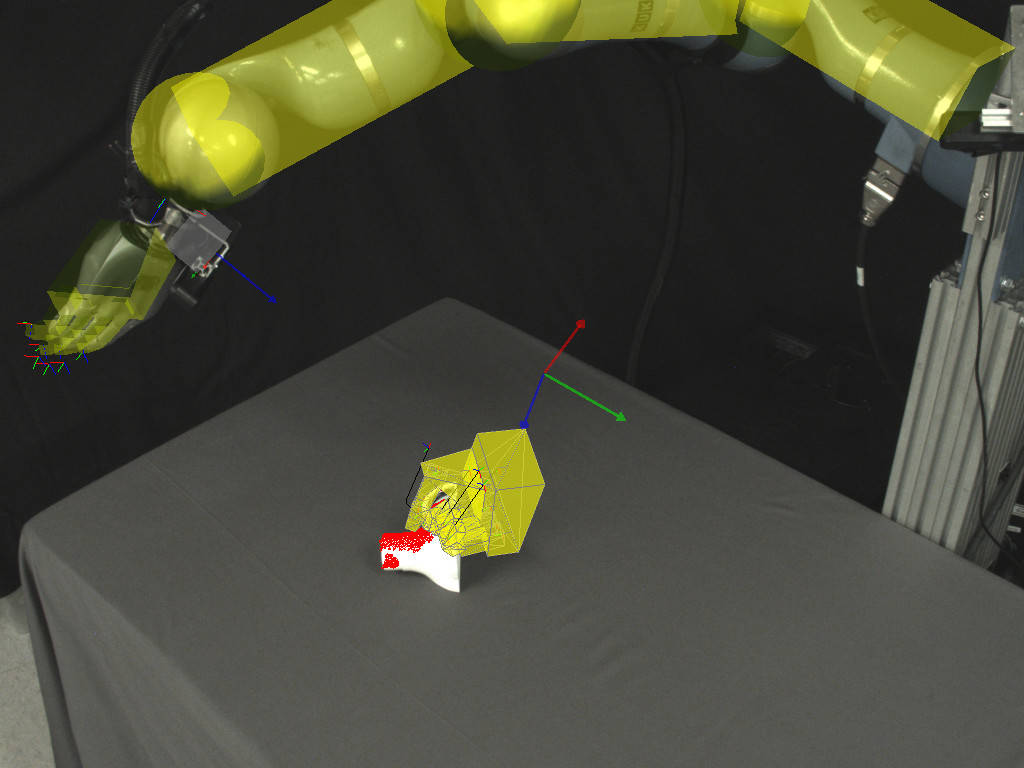} 
\includegraphics[width=\testingw,trim={10cm 5cm 15cm 10cm},clip]{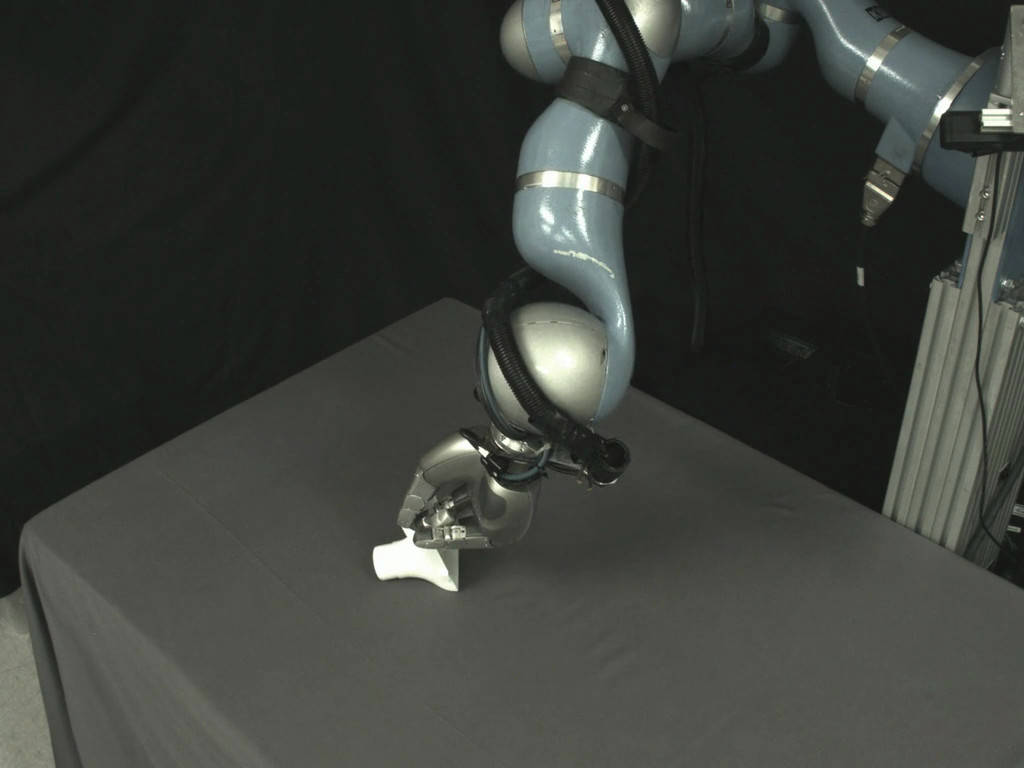}
\includegraphics[width=\testingw,trim={10cm 5cm 15cm 10cm},clip]{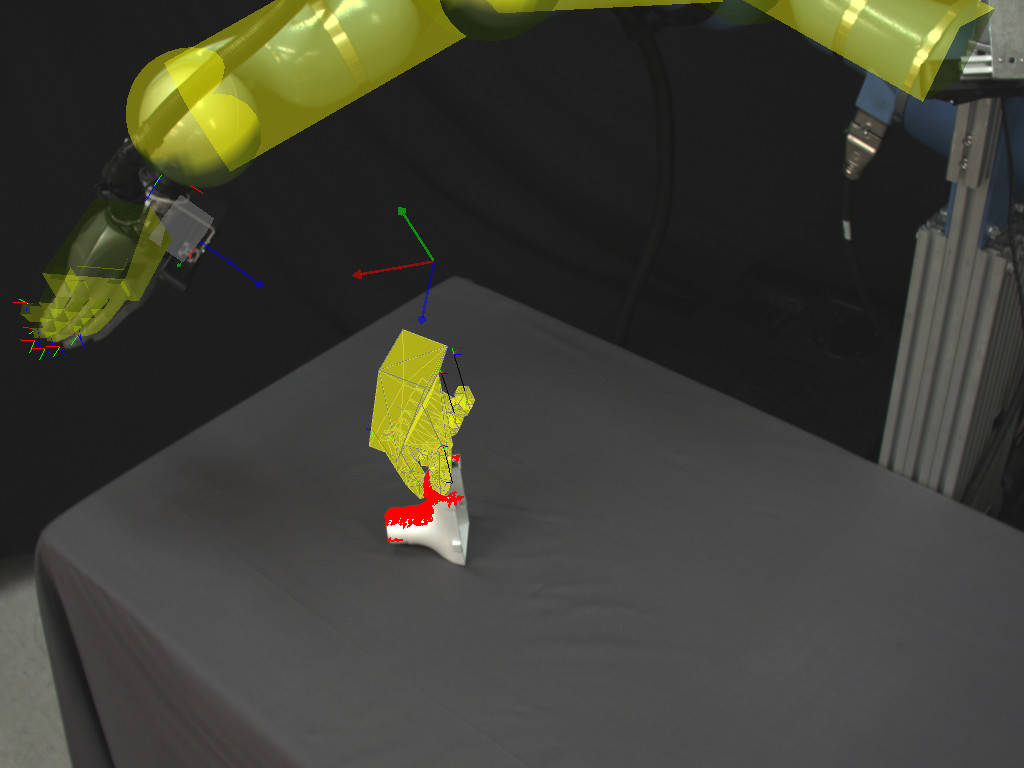}
\includegraphics[width=\testingw,trim={19.3cm 9.6cm 27.8cm 18.77cm},clip]{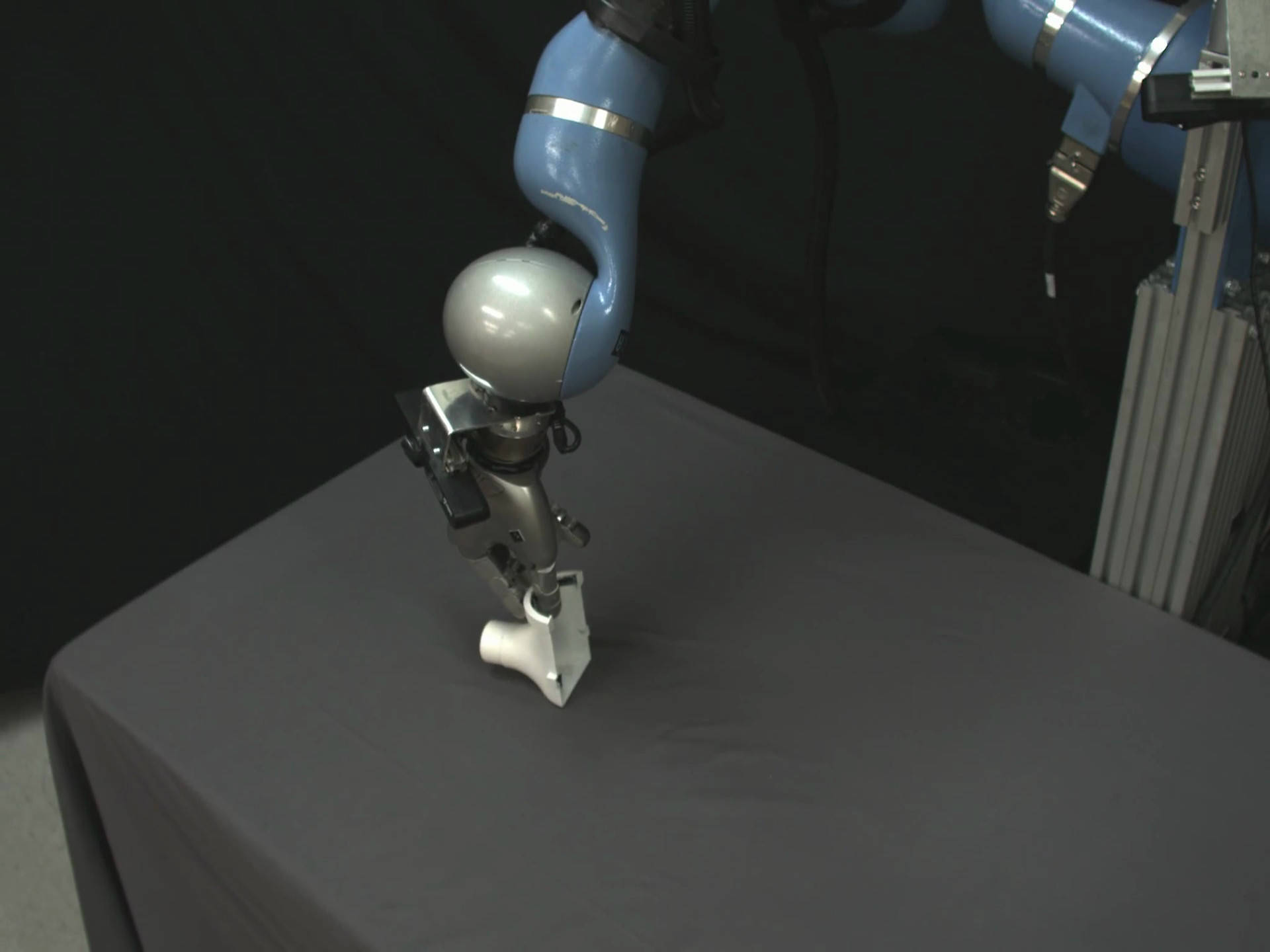}
\includegraphics[width=\testingw,trim={12cm 5cm 13cm 10cm},clip]{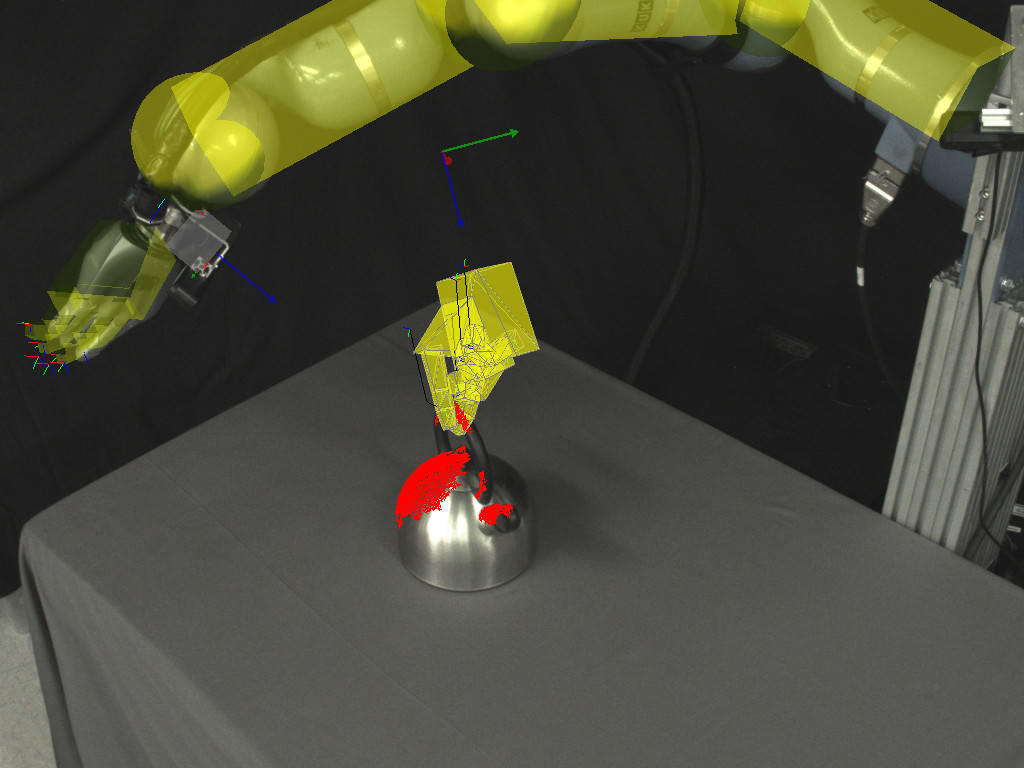} 
\includegraphics[width=\testingw,trim={12cm 5cm 13cm 10cm},clip]{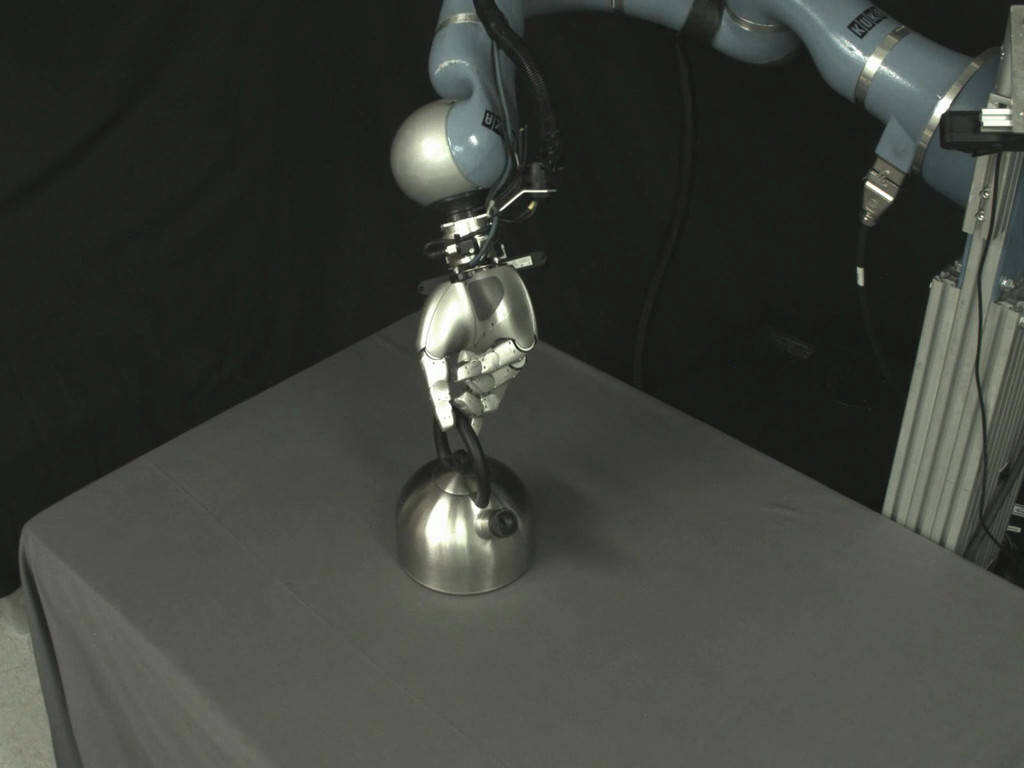}
\includegraphics[width=\testingw,trim={12cm 5cm 13cm 10cm},clip]{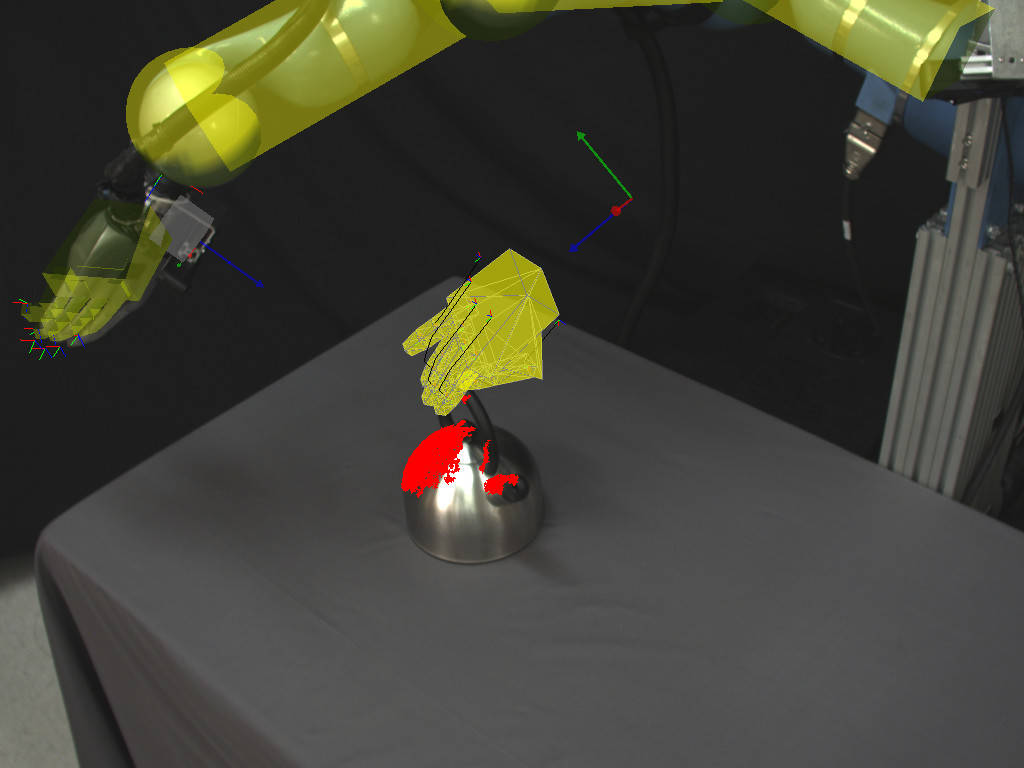}
\includegraphics[width=\testingw,trim={23.3cm 9.6cm 23.8cm 18.77cm},clip]{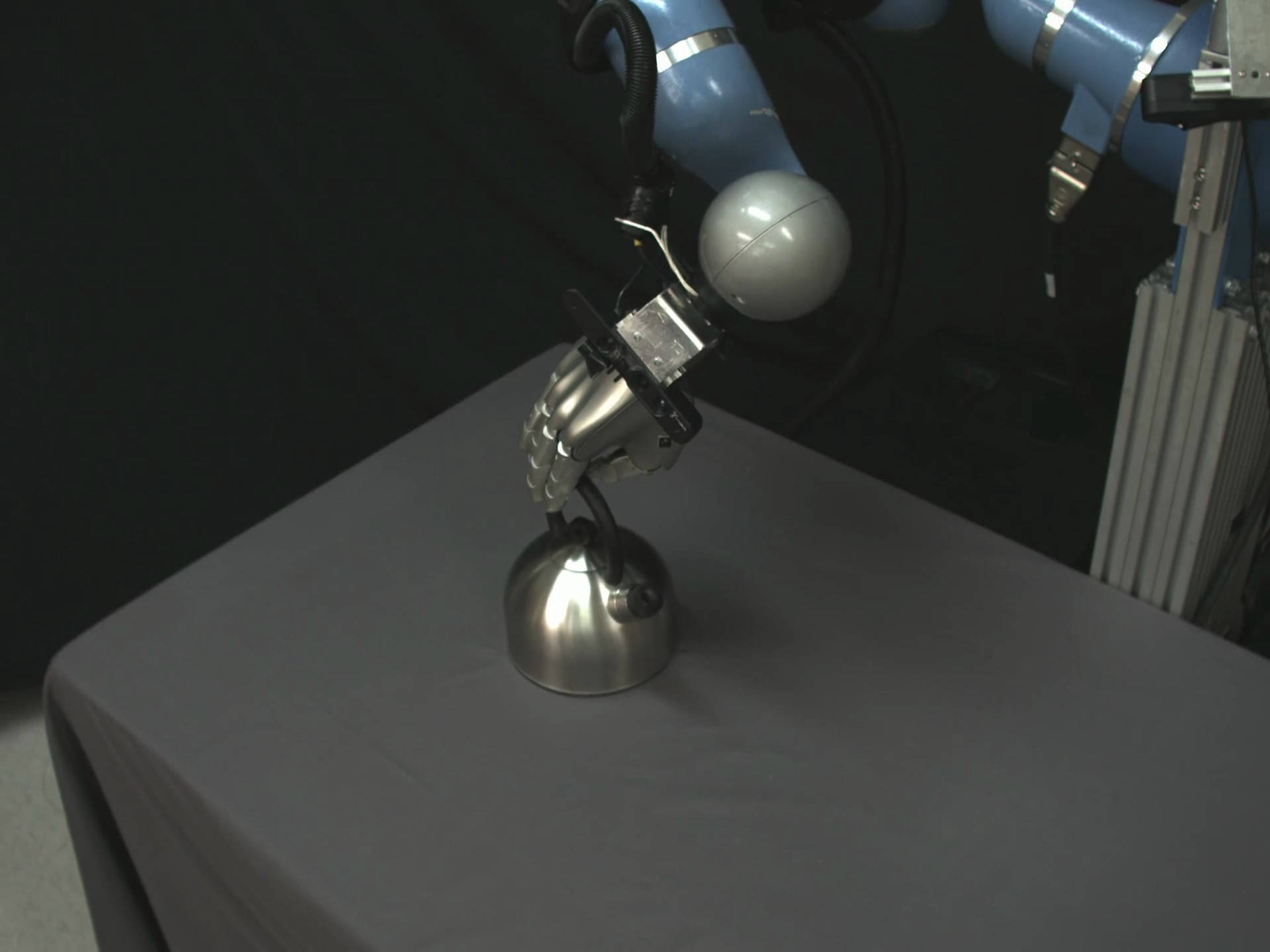} \\
\includegraphics[width=\testingw,trim={10cm 5cm 15cm 10cm},clip]{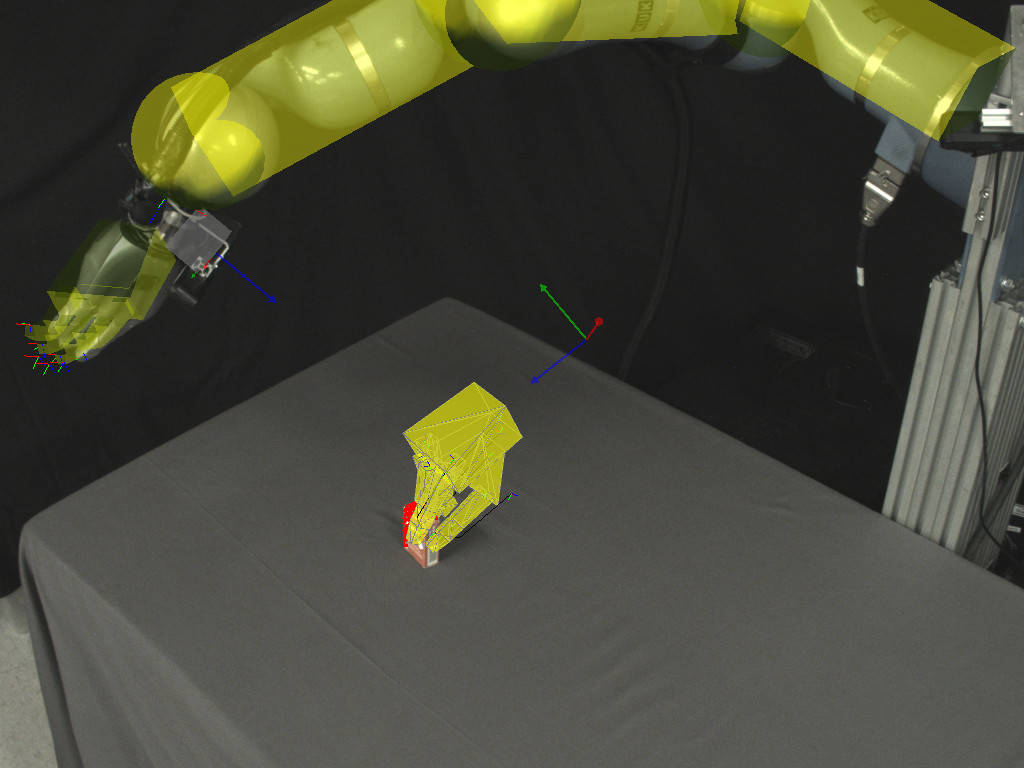}
\includegraphics[width=\testingw,trim={10cm 5cm 15cm 10cm},clip]{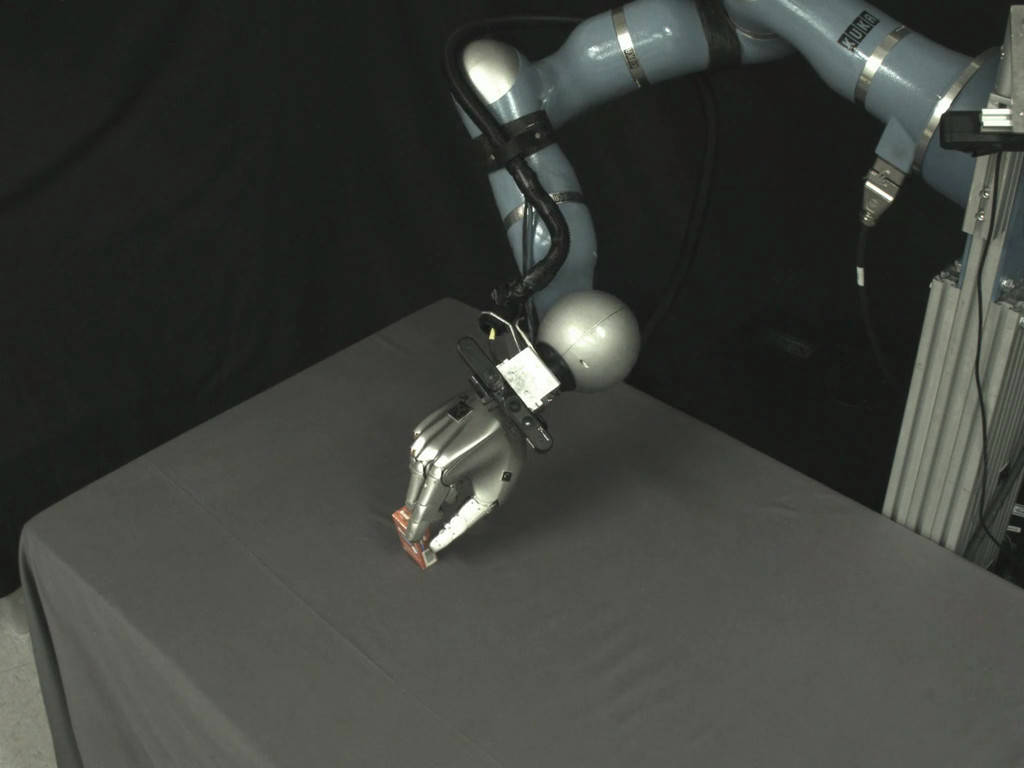}
\includegraphics[width=\testingw,trim={10cm 5cm 15cm 10cm},clip]{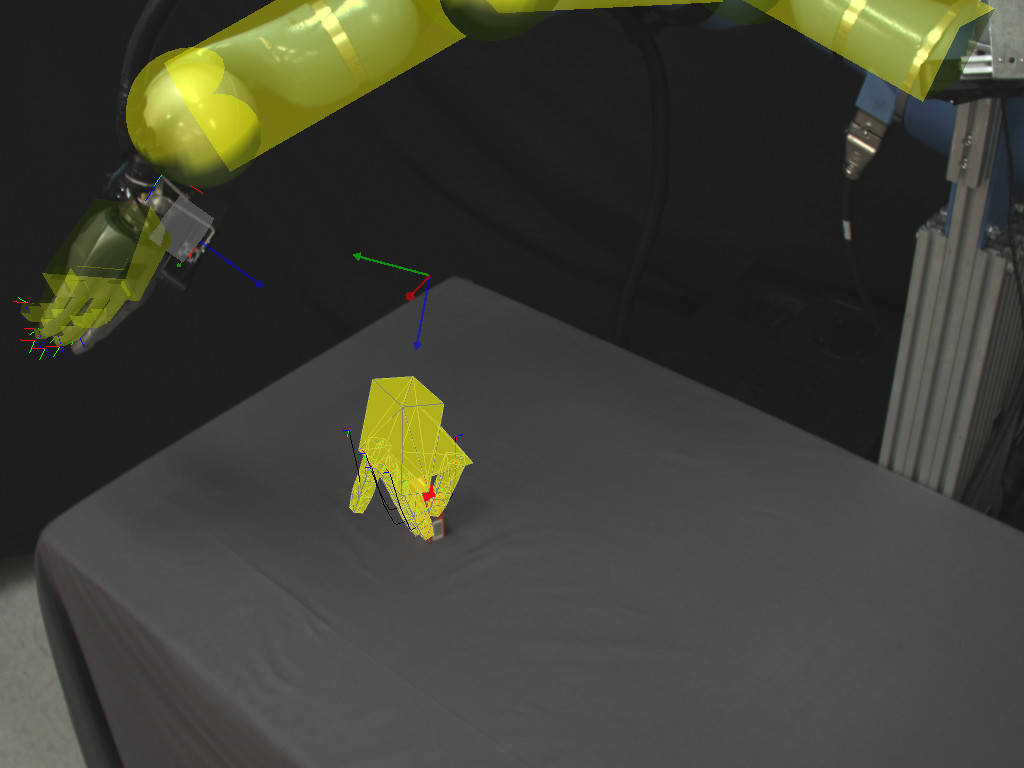}
\includegraphics[width=\testingw,trim={19.3cm 9.6cm 27.8cm 18.77cm},clip]{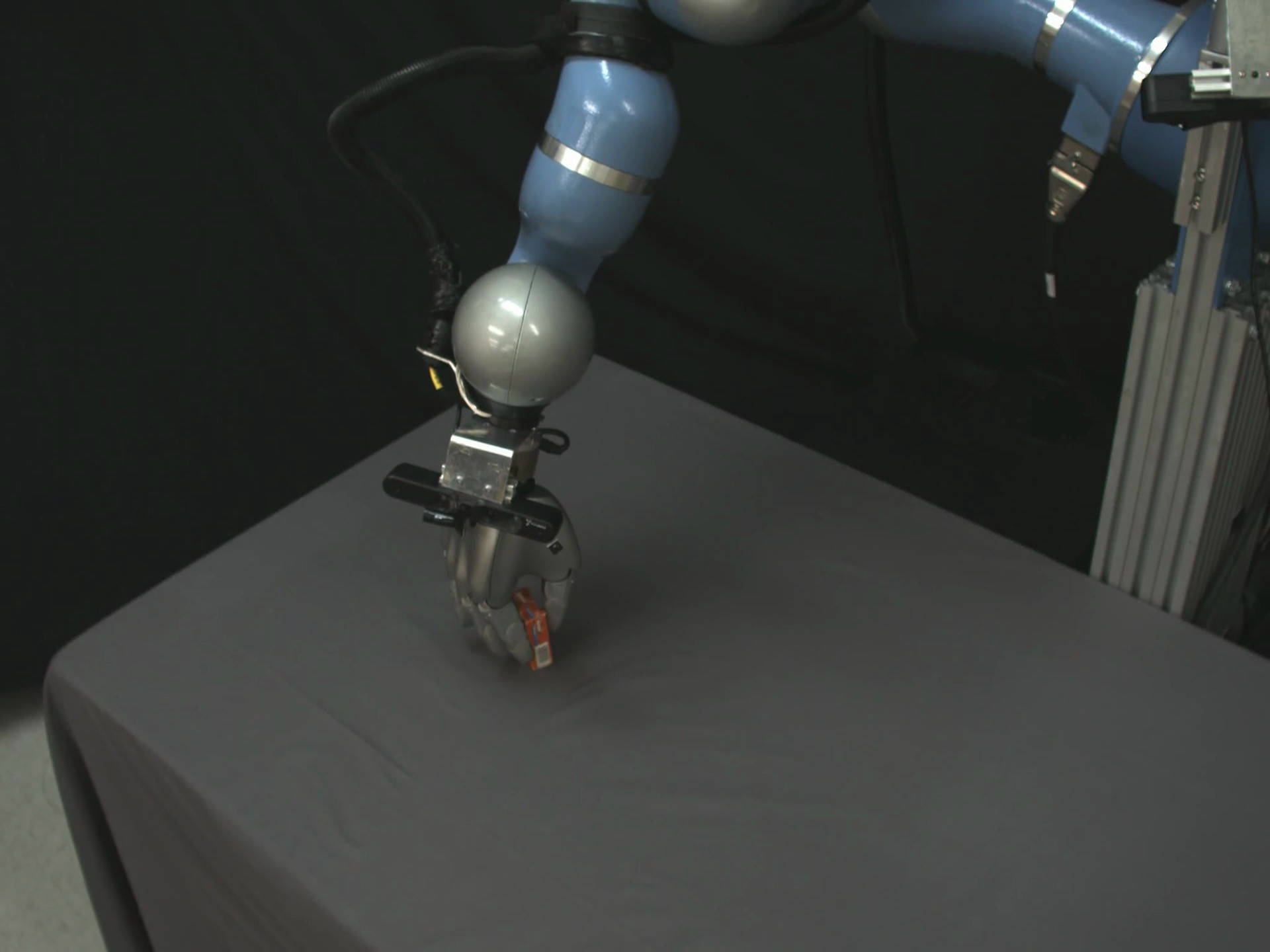}
\includegraphics[width=\testingw,trim={12cm 3cm 13cm 12cm},clip]{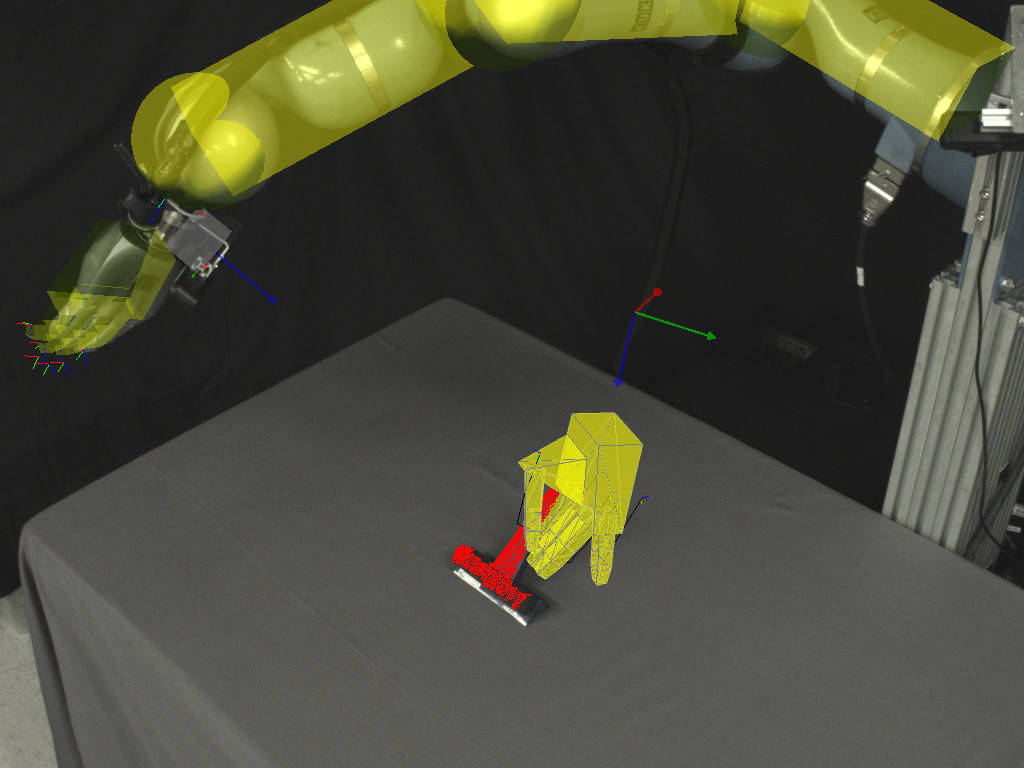} 
\includegraphics[width=\testingw,trim={12cm 3cm 13cm 12cm},clip]{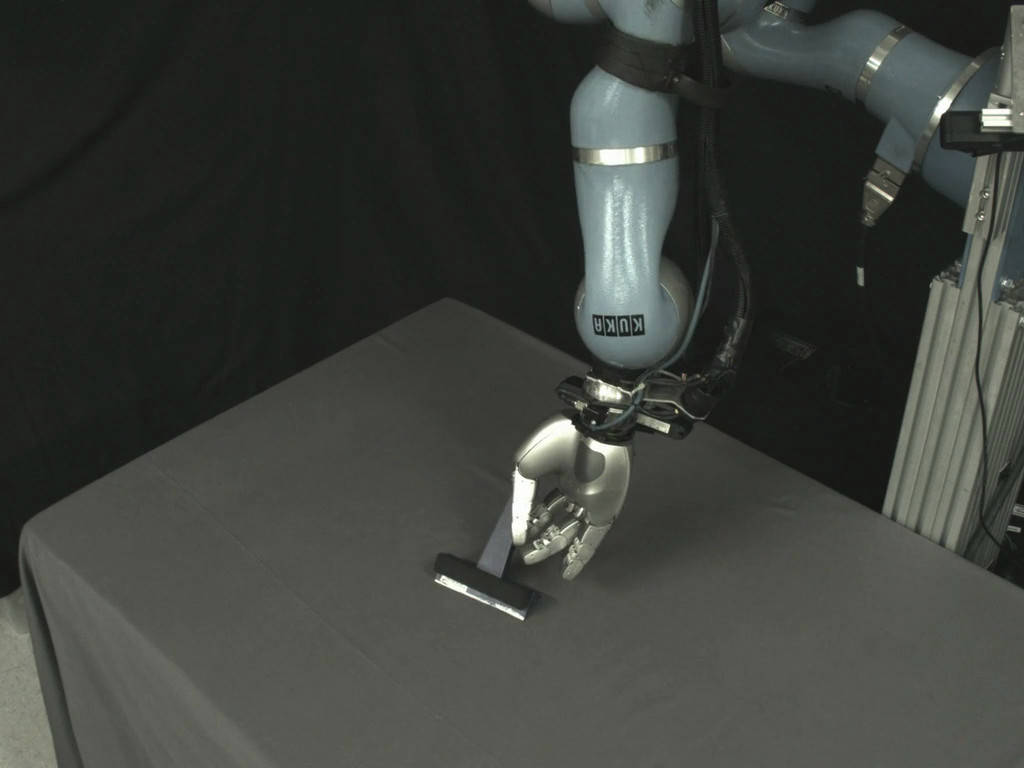}
\includegraphics[width=\testingw,trim={12cm 3cm 13cm 12cm},clip]{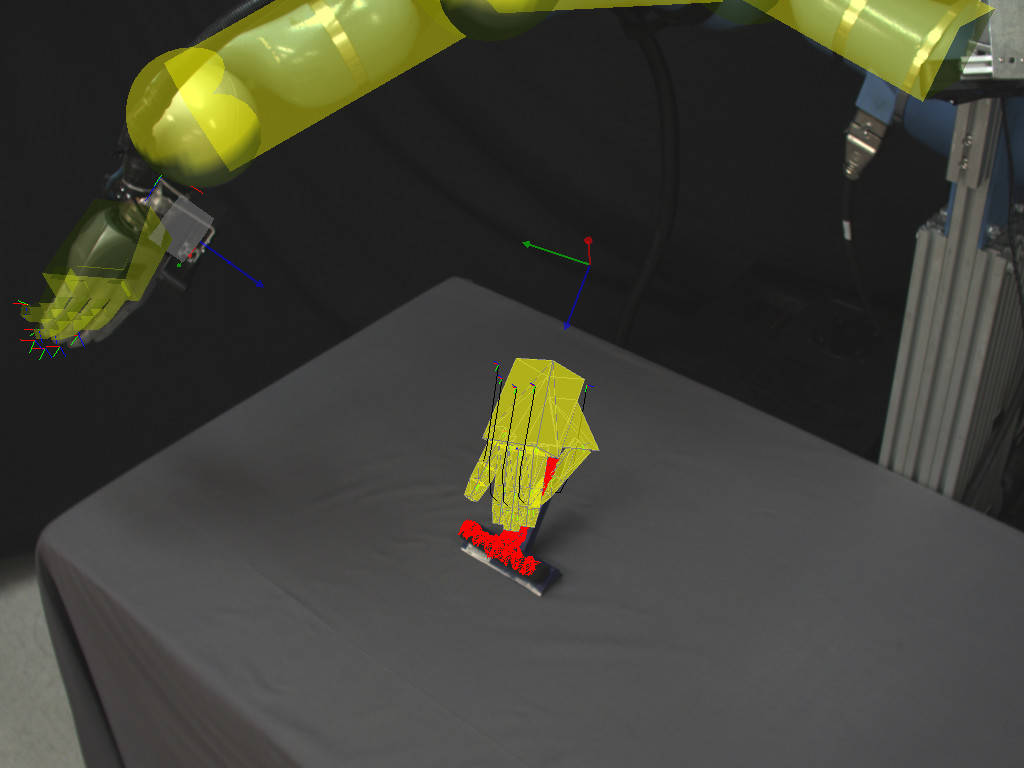}
\includegraphics[width=\testingw,trim={23.3cm 6.1cm 23.8cm 22.27cm},clip]{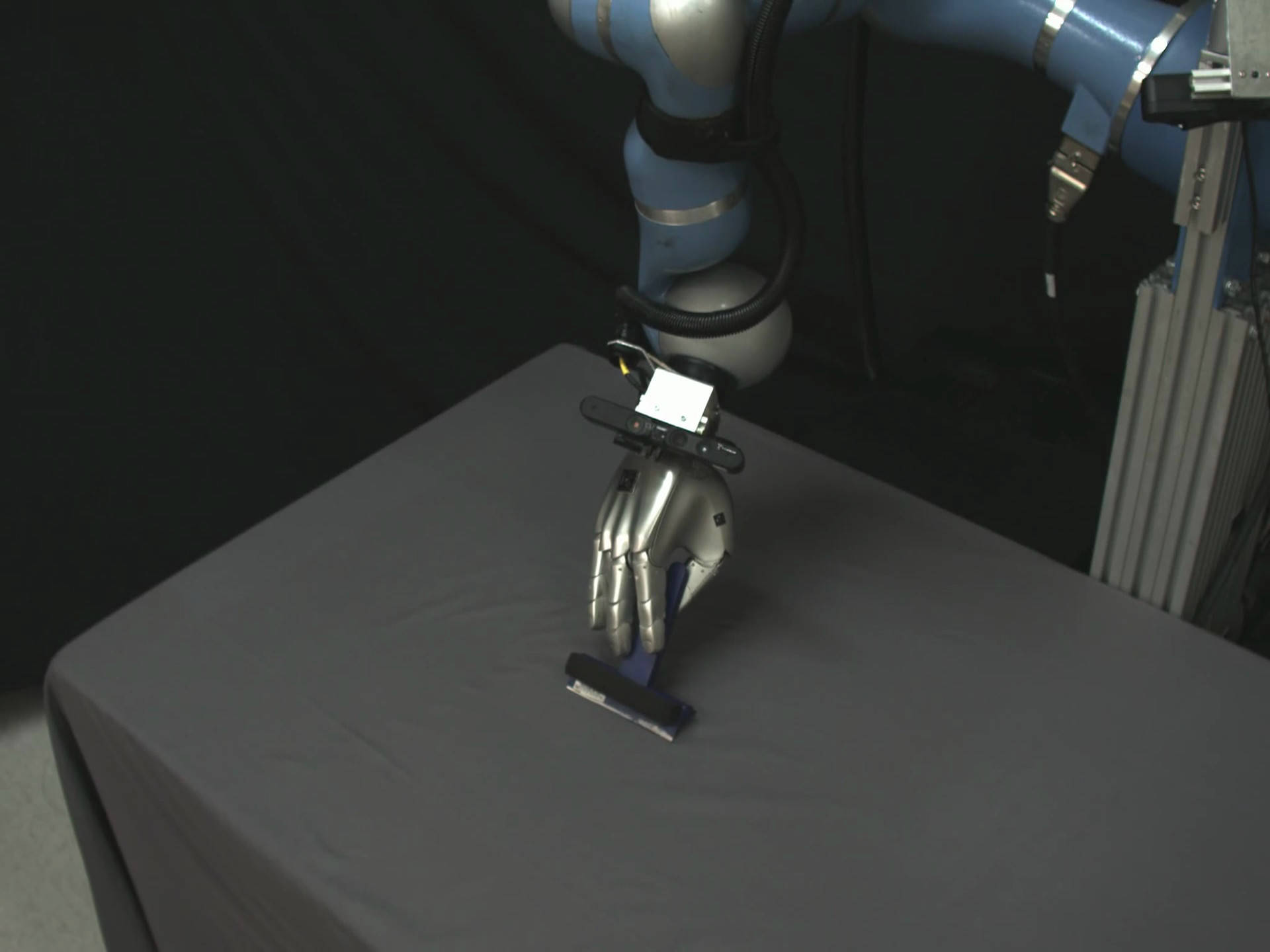} \\
\includegraphics[width=\testingw,trim={12cm 5cm 13cm 10cm},clip]{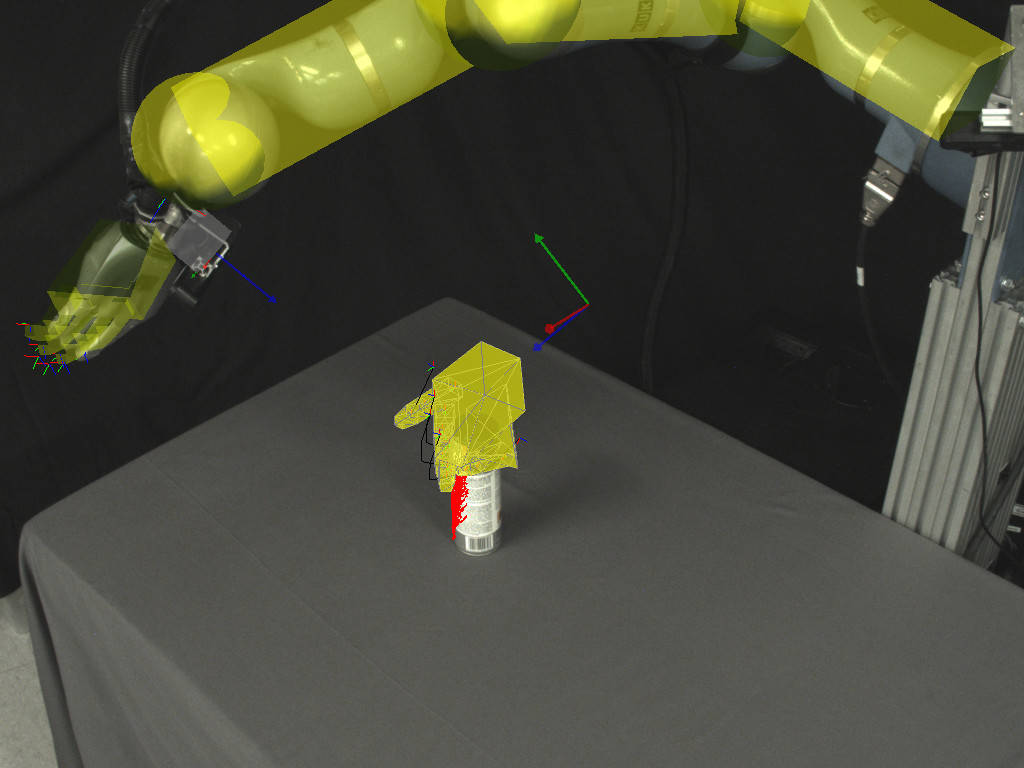} 
\includegraphics[width=\testingw,trim={12cm 5cm 13cm 10cm},clip]{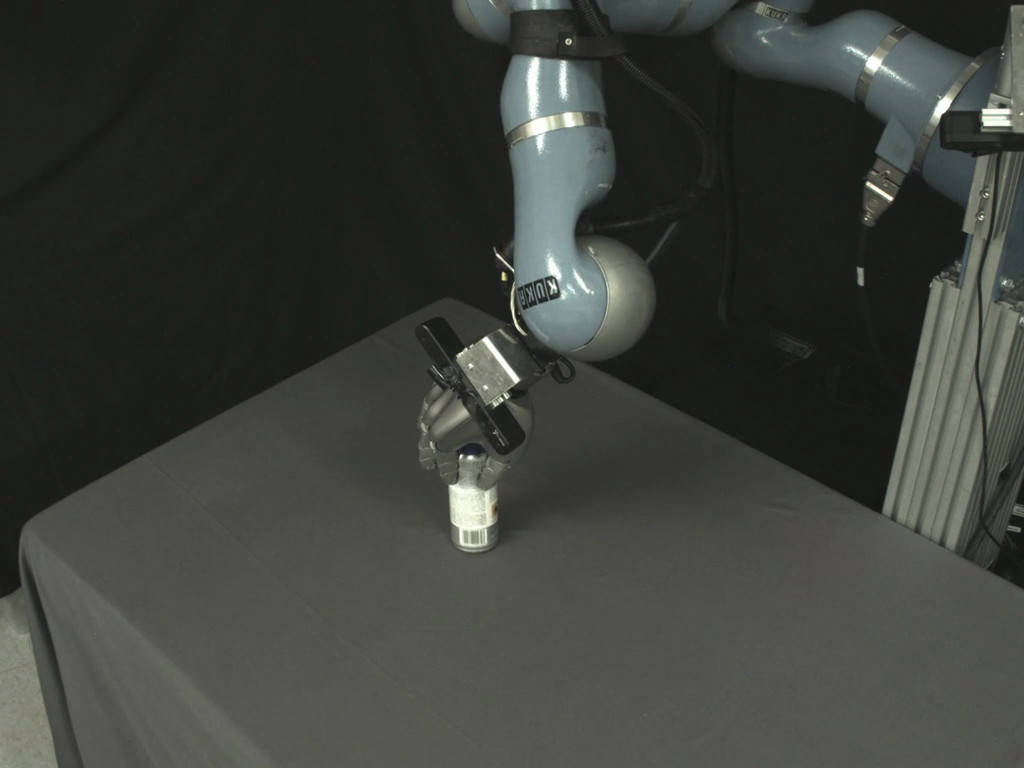}
\includegraphics[width=\testingw,trim={12cm 5cm 13cm 10cm},clip]{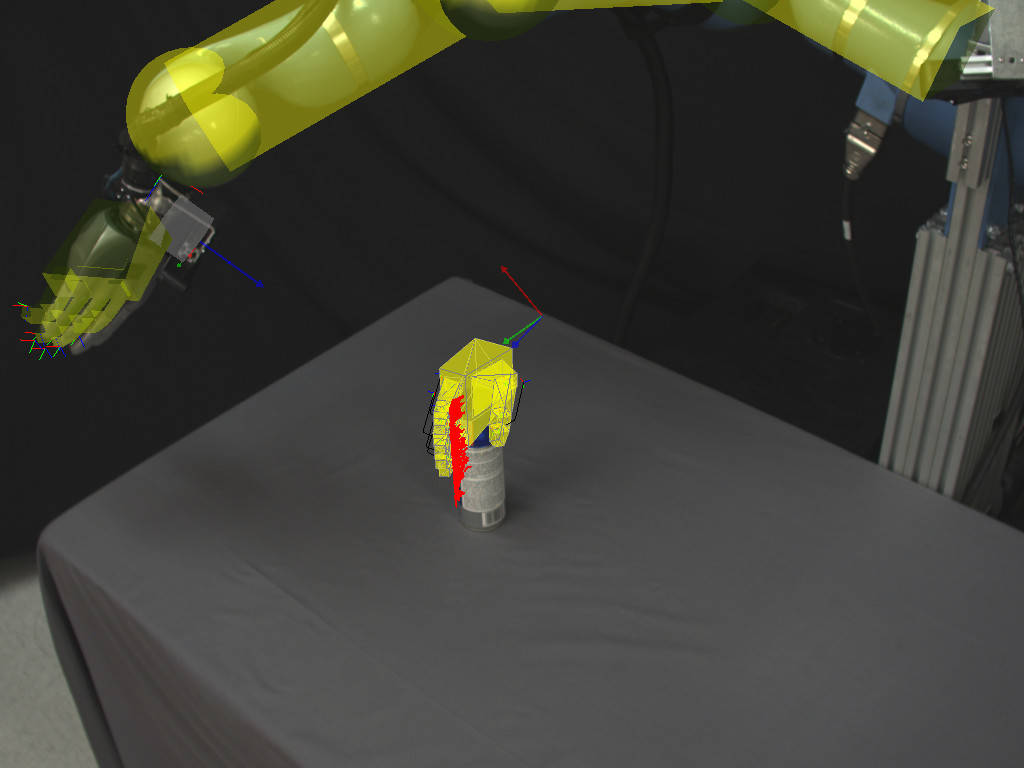}
\includegraphics[width=\testingw,trim={23.3cm 9.6cm 23.8cm 18.77cm},clip]{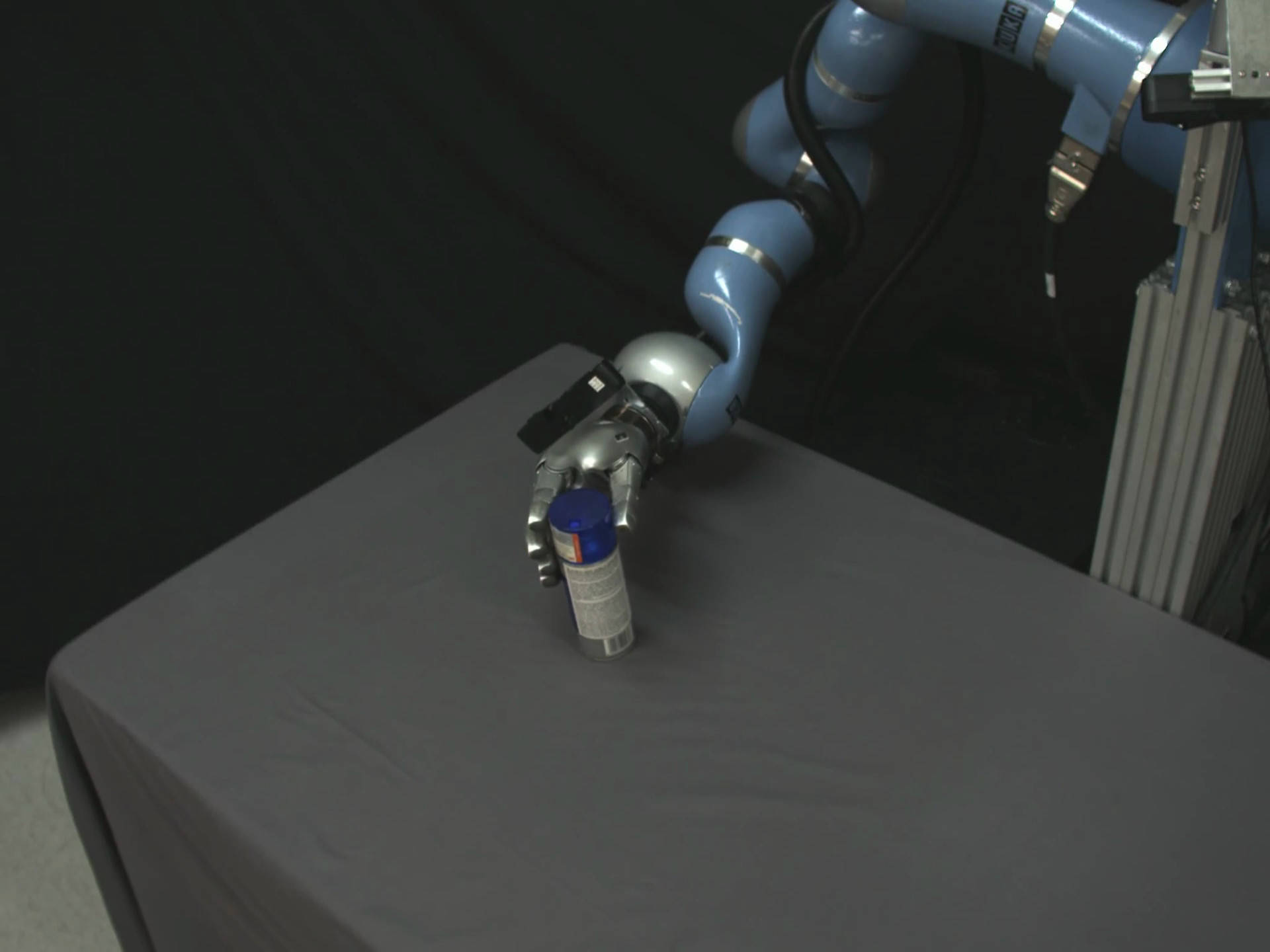}
\includegraphics[width=\testingw,trim={12cm 5cm 13cm 10cm},clip]{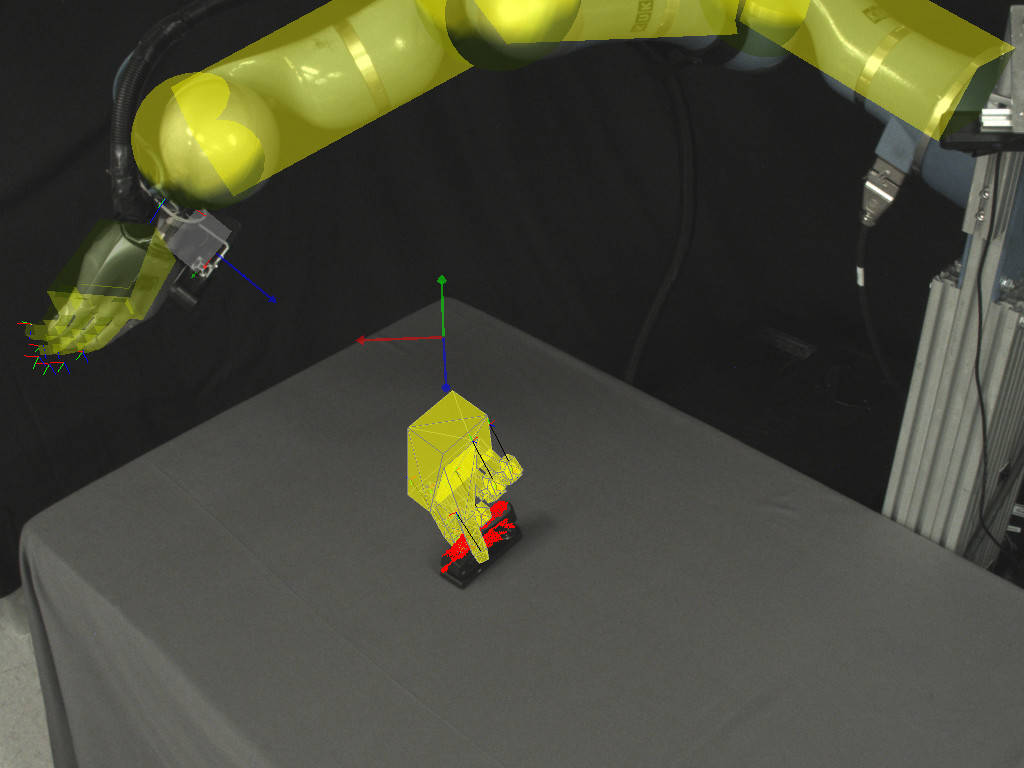} 
\includegraphics[width=\testingw,trim={12cm 5cm 13cm 10cm},clip]{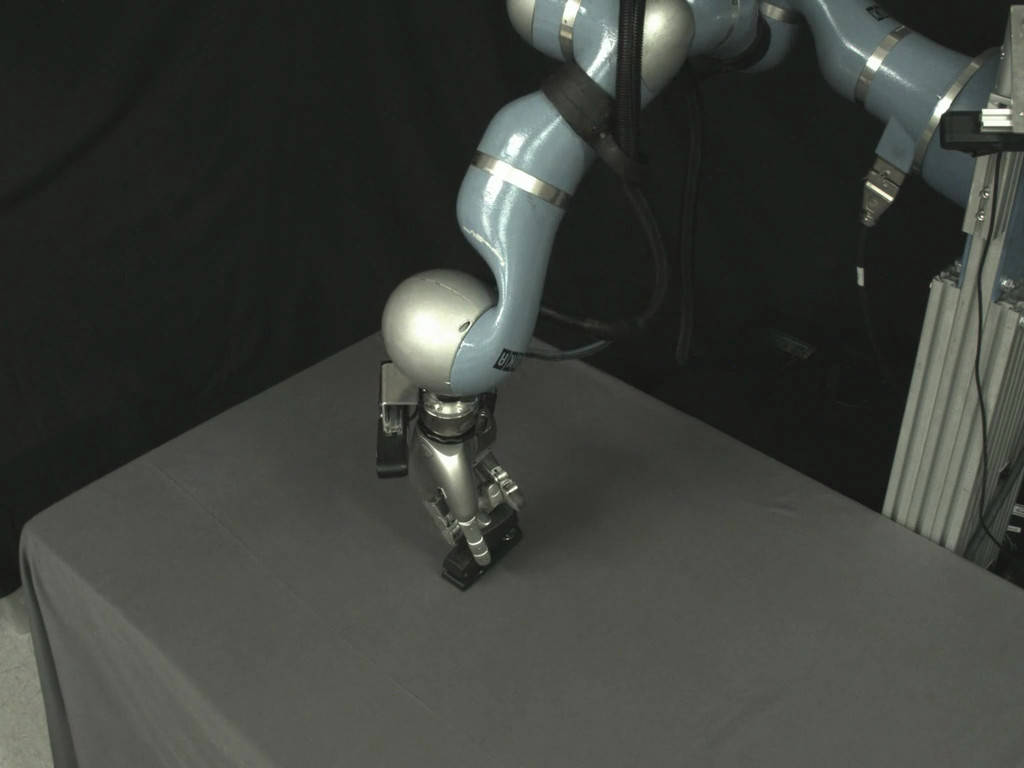}
\includegraphics[width=\testingw,trim={12cm 5cm 13cm 10cm},clip]{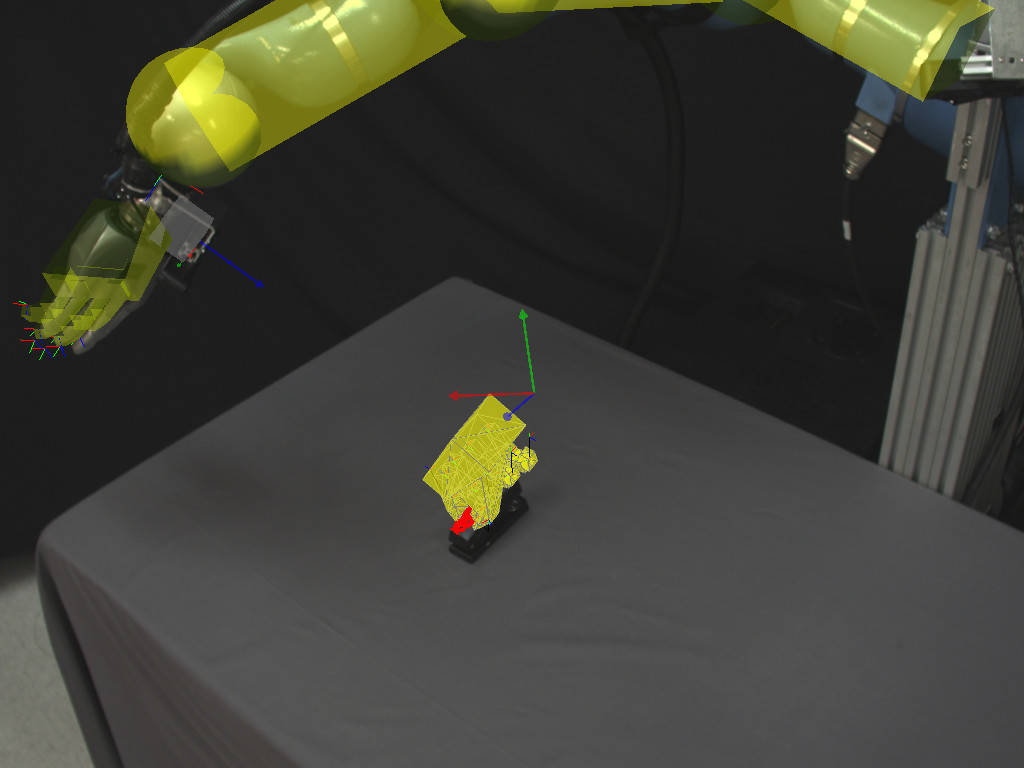}
\includegraphics[width=\testingw,trim={23.3cm 9.6cm 23.8cm 18.77cm},clip]{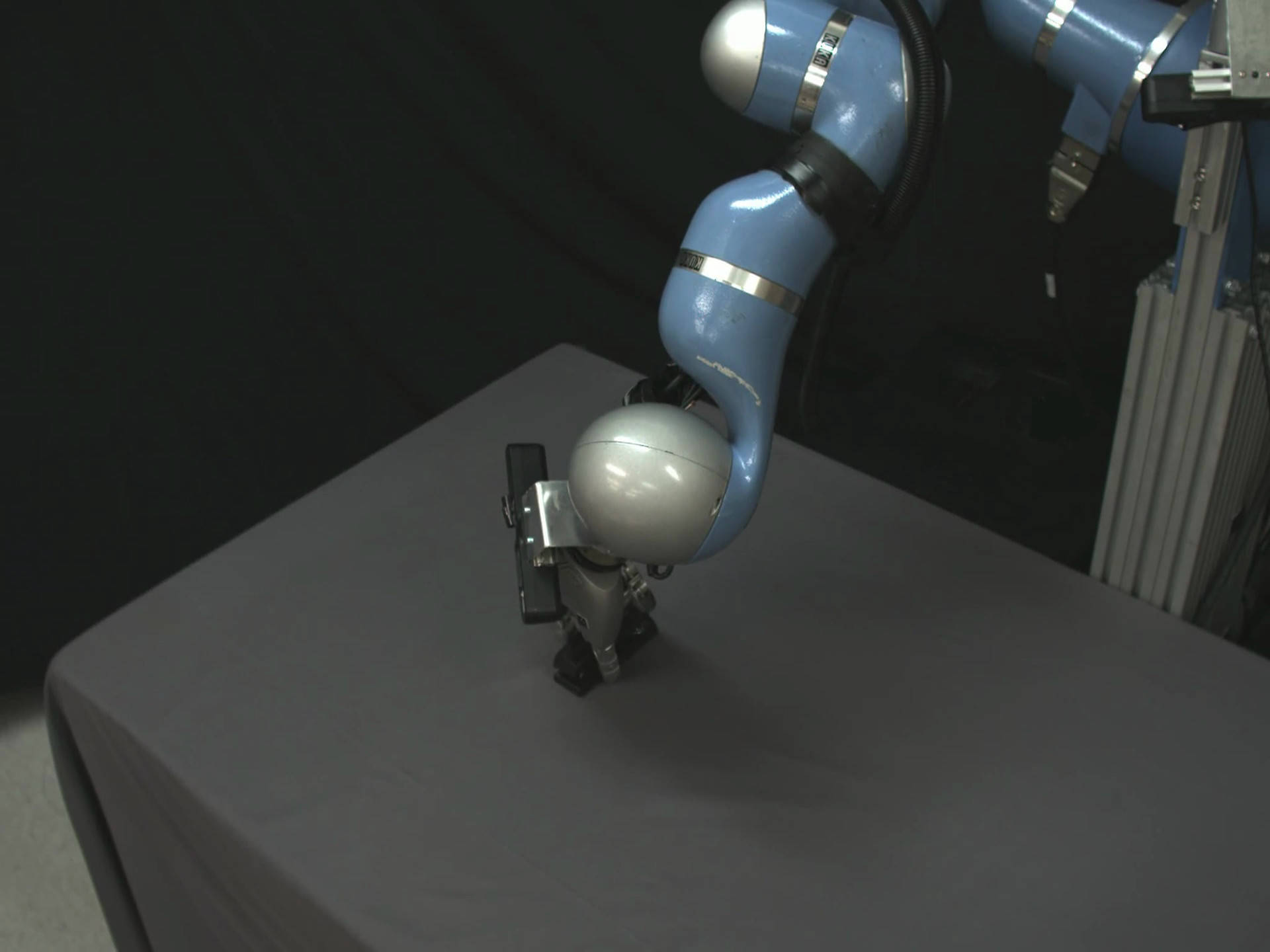}
\caption{\label{fig:graspscomp1}Comparison of grasps executed by algorithms A1+AT and A4+AT. These are 10 of the 15 cases where A1+AT failed and A1+AT succeeded. Columns are labelled by the variant.}
\end{figure*}

\begin{figure*}
\newcommand{\testingw}{0.12\linewidth}
\mbox{\bf \large \hspace{1.4cm} A1+AT \hspace{2.85cm} A4+AT \hspace{2.85cm} A1+AT \hspace{2.85cm} A4+AT } 
\mbox{\bf \large \hspace{1.2cm} (Successes) 
\hspace{2.3cm} (Failures) \hspace{2.3cm} (Failures) \hspace{2.3cm} (Failures) } \\
\includegraphics[width=\testingw,trim={12cm 6cm 13cm 10cm},clip]{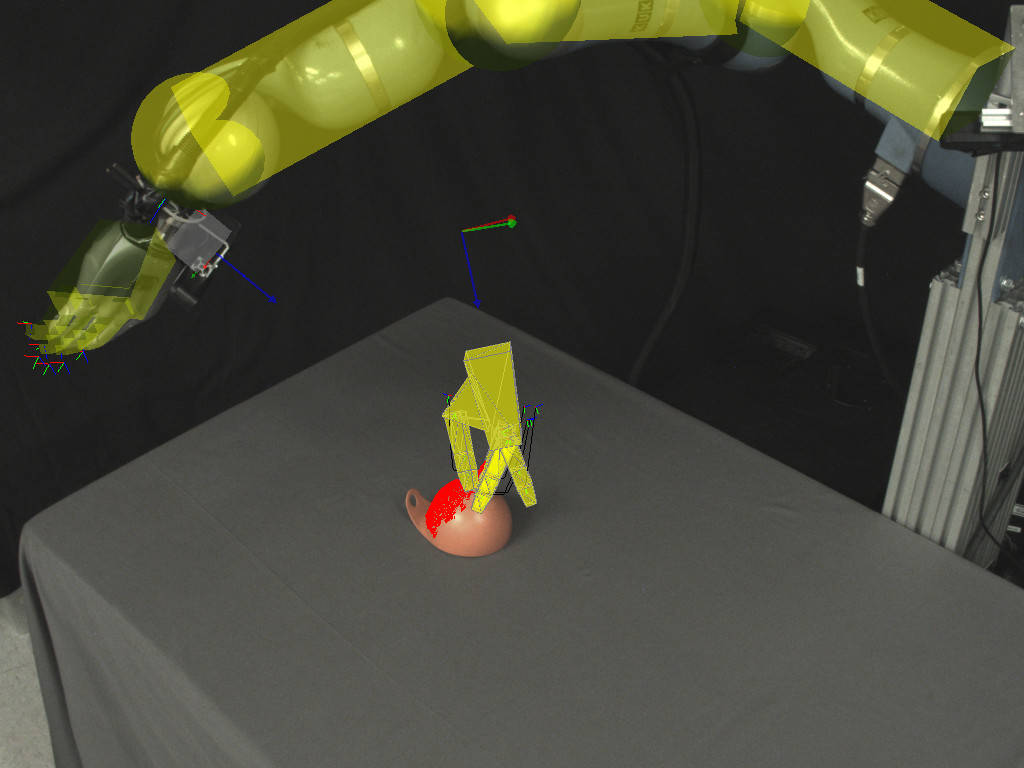} 
\includegraphics[width=\testingw,trim={12cm 6cm 13cm 10cm},clip]{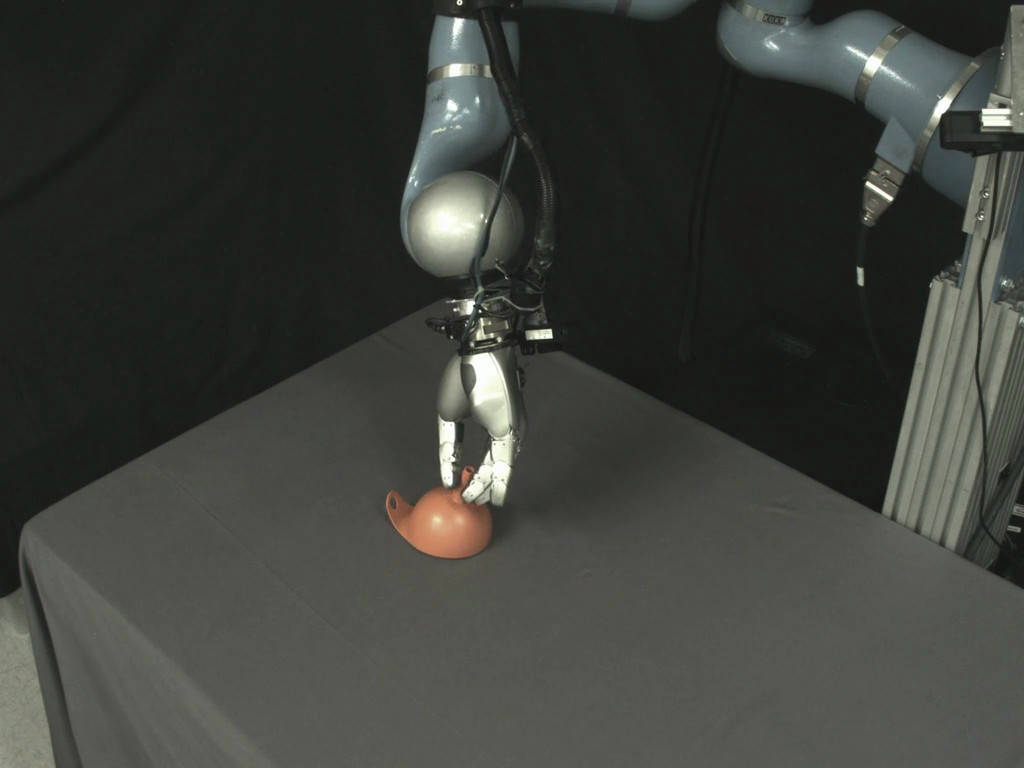}
\includegraphics[width=\testingw,trim={12cm 6cm 13cm 10cm},clip]{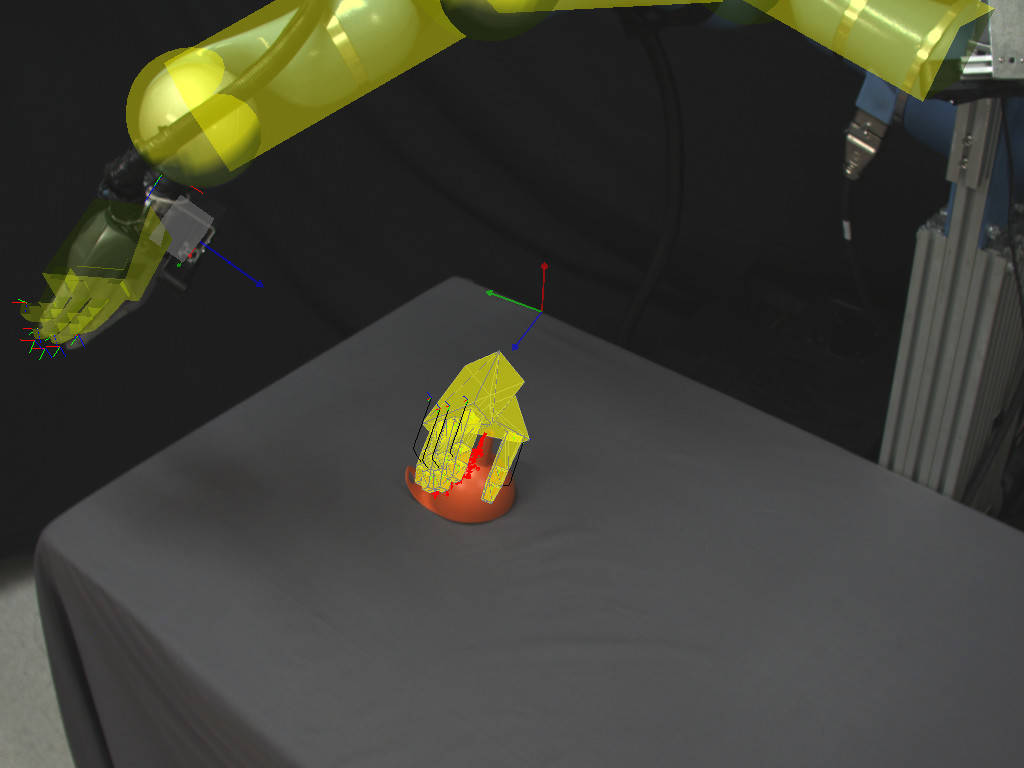} 
\includegraphics[width=\testingw,trim={23.3cm 10.5cm 23.8cm 19.7cm},clip]{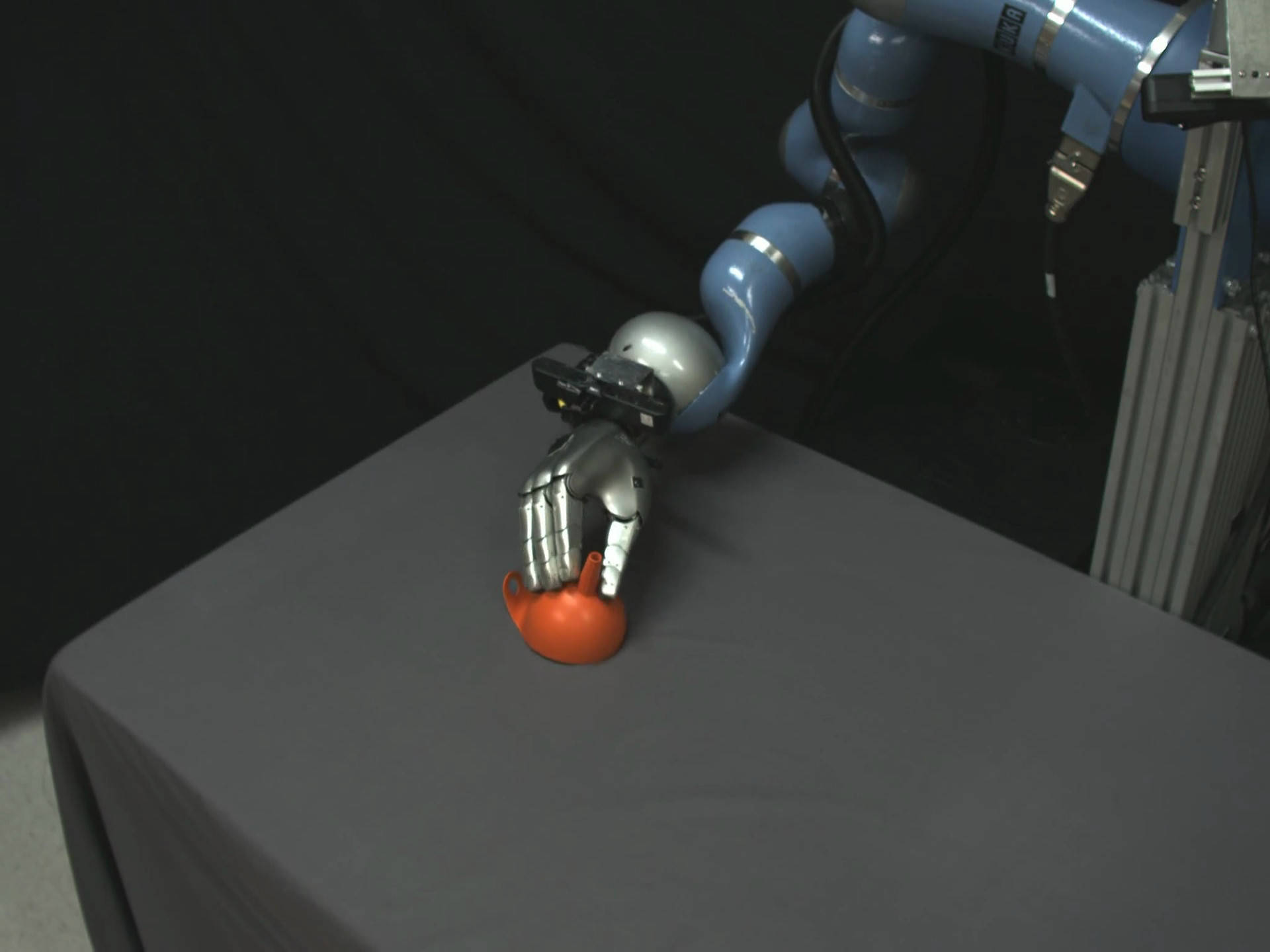}
\includegraphics[width=\testingw,trim={9cm 5cm 15cm 10cm},clip]{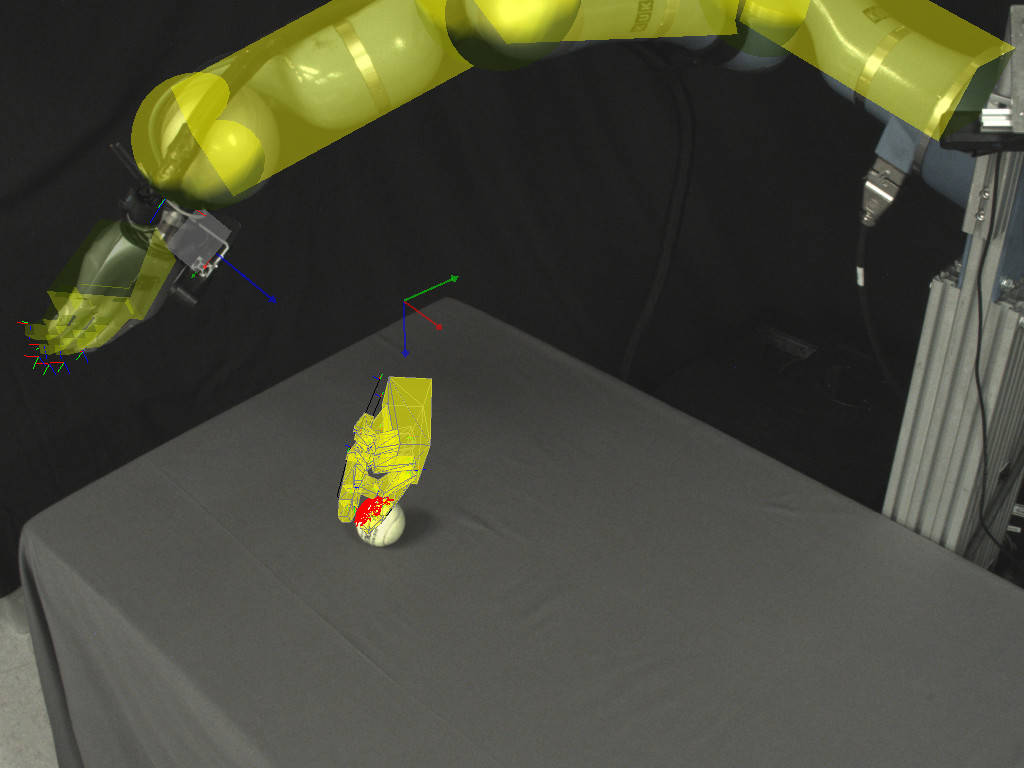} 
\includegraphics[width=\testingw,trim={9cm 5cm 15cm 10cm},clip]{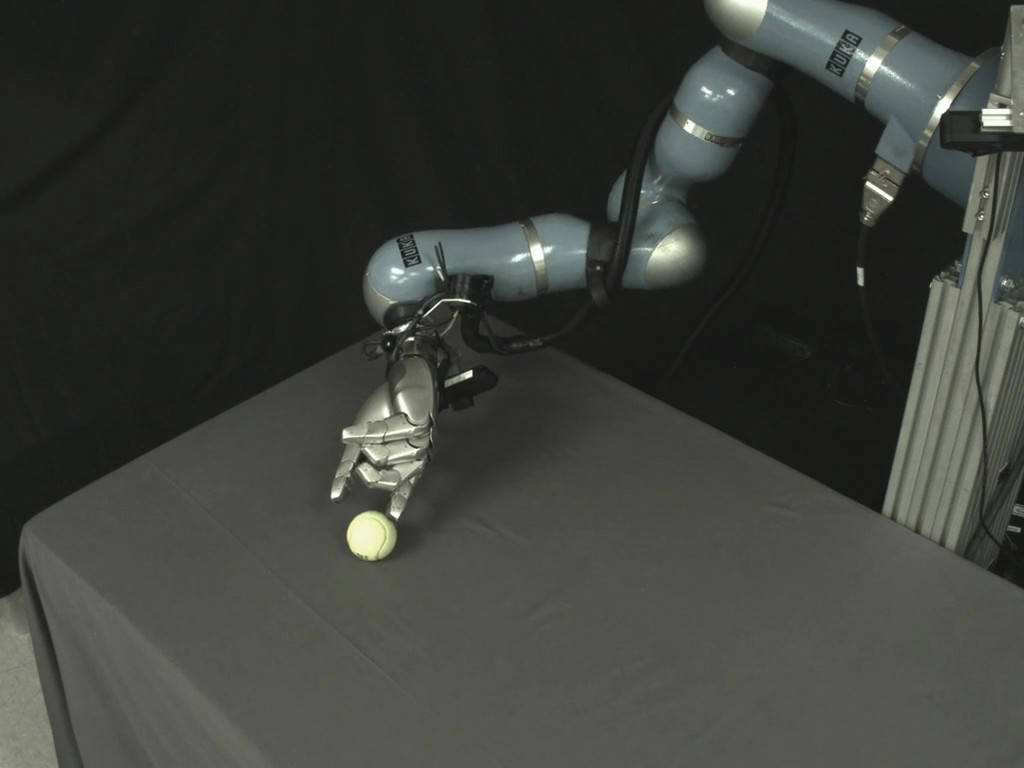}
\includegraphics[width=\testingw,trim={9cm 5cm 15cm 10cm},clip]{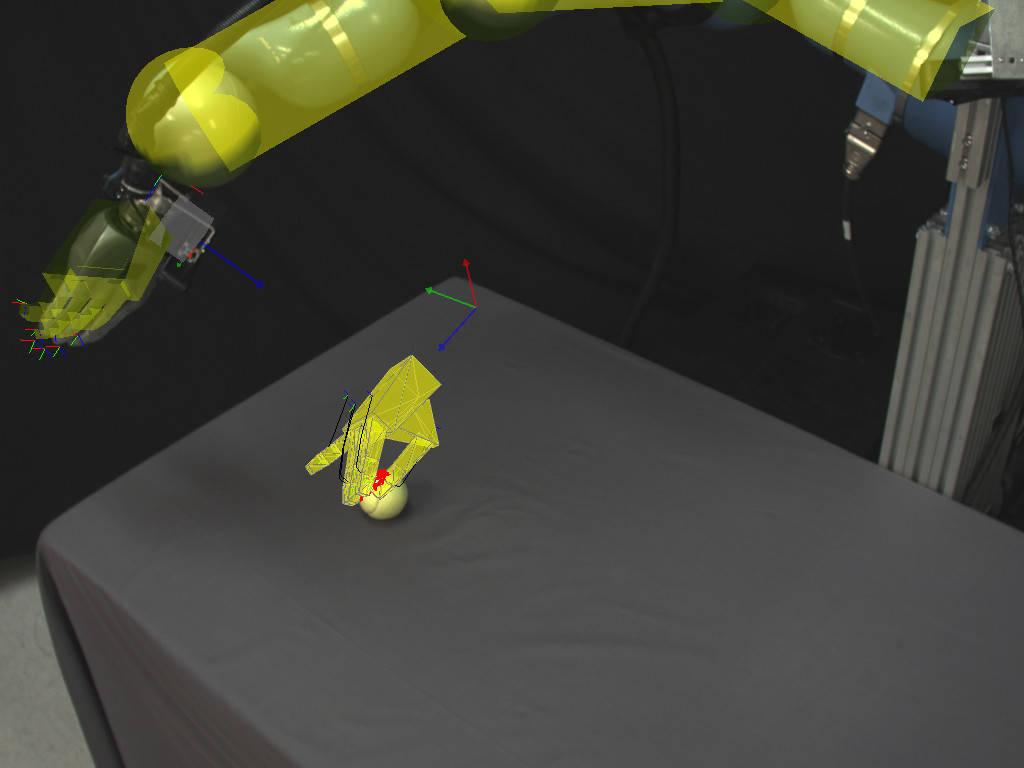} 
\includegraphics[width=\testingw,trim={17.4cm 9.6cm 27.8cm 18.77cm},clip]{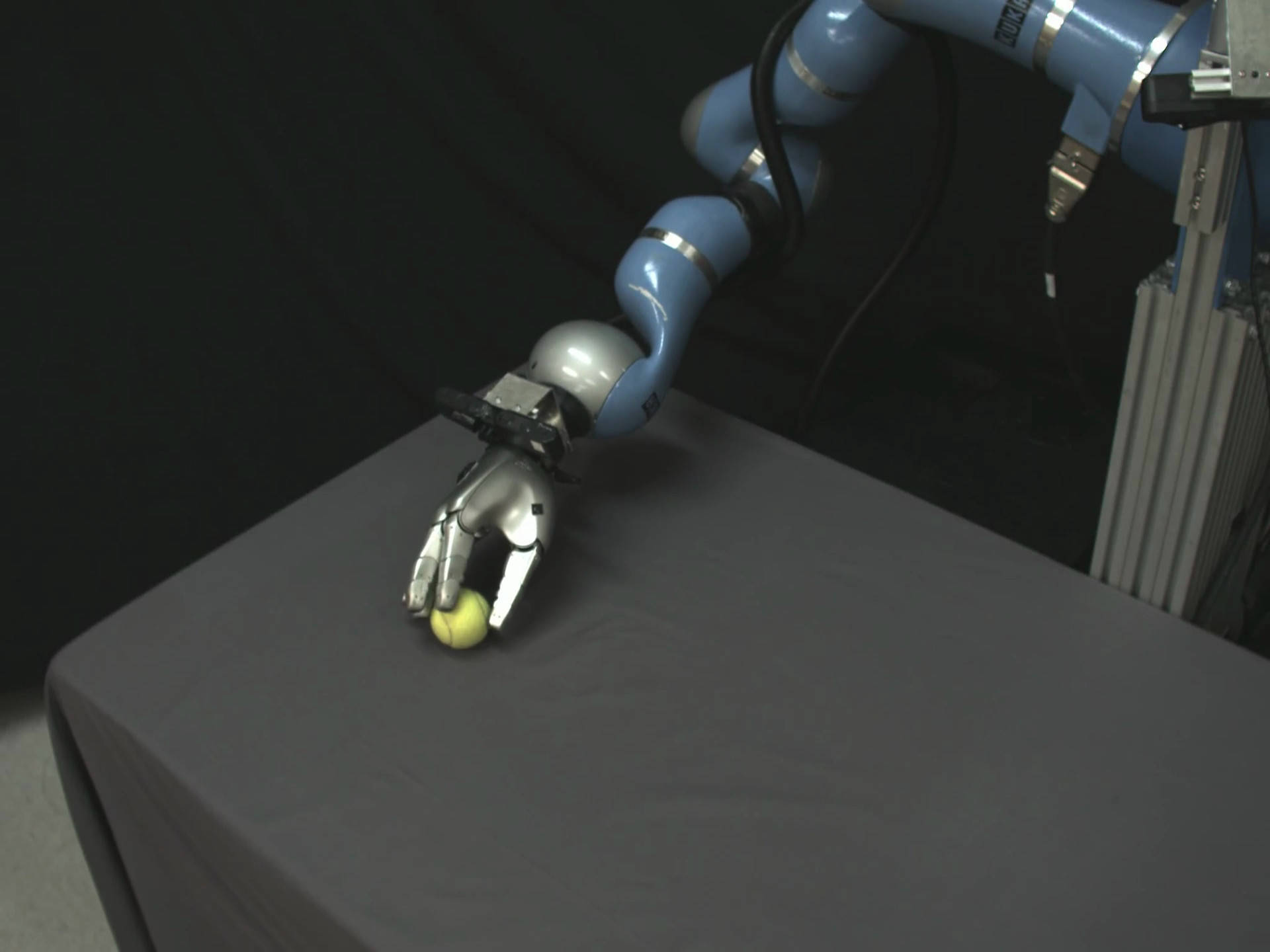}\\
\includegraphics[width=\testingw,trim={12cm 5cm 13cm 10cm},clip]{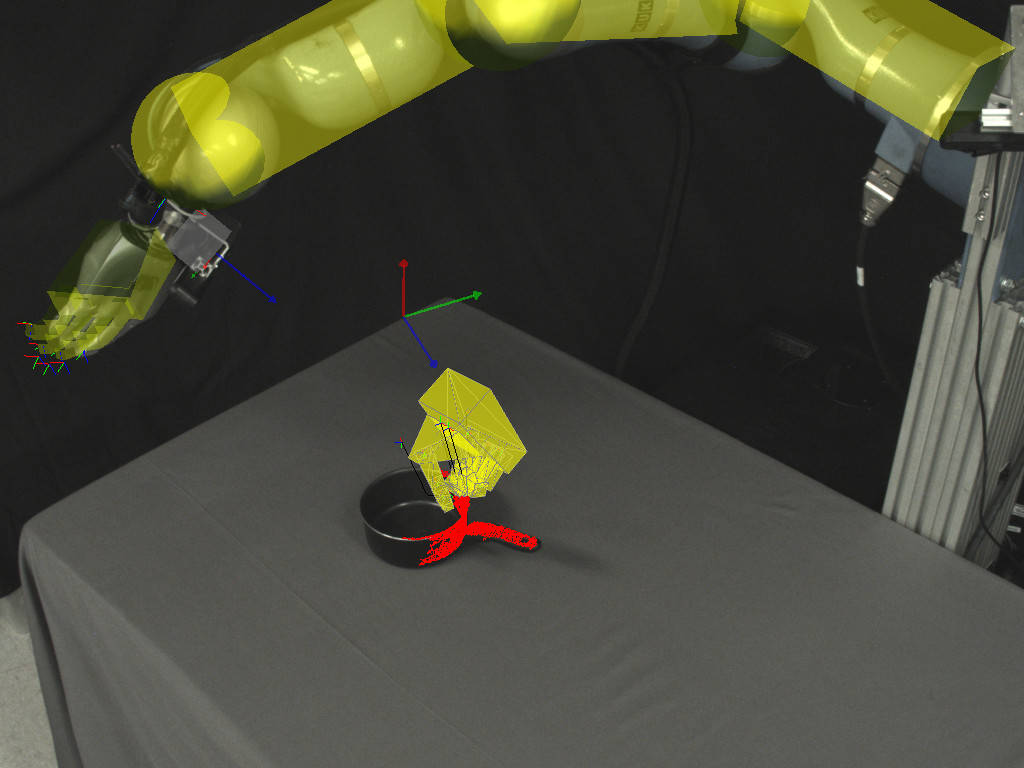} 
\includegraphics[width=\testingw,trim={12cm 5cm 13cm 10cm},clip]{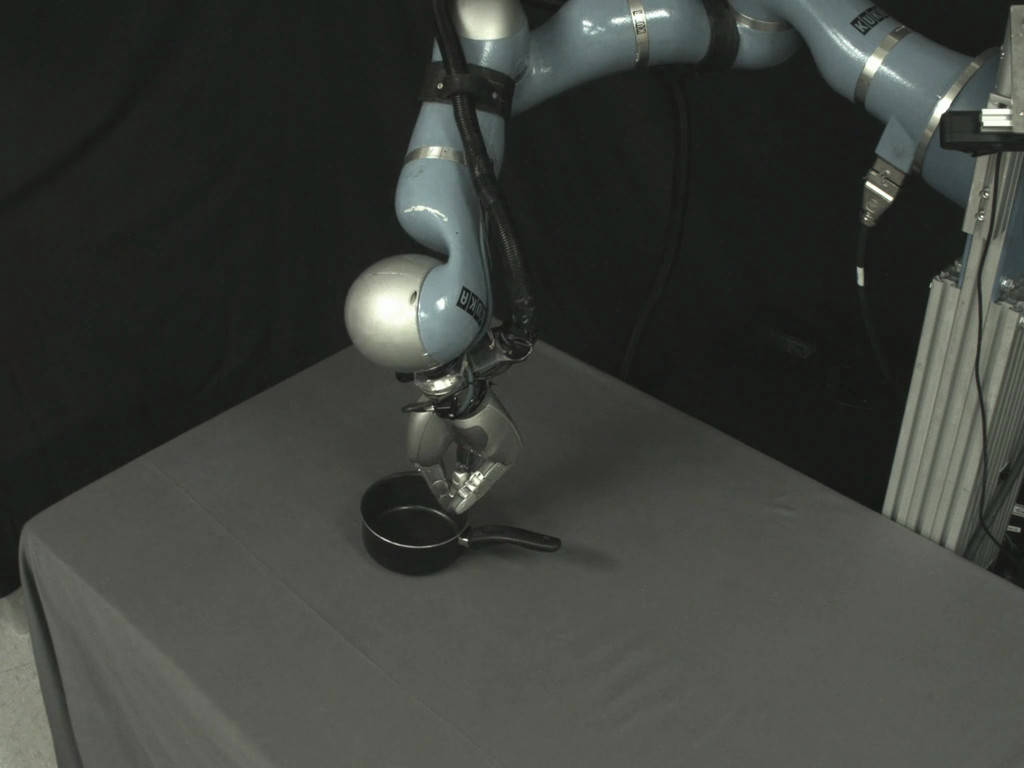}
\includegraphics[width=\testingw,trim={12cm 5cm 13cm 10cm},clip]{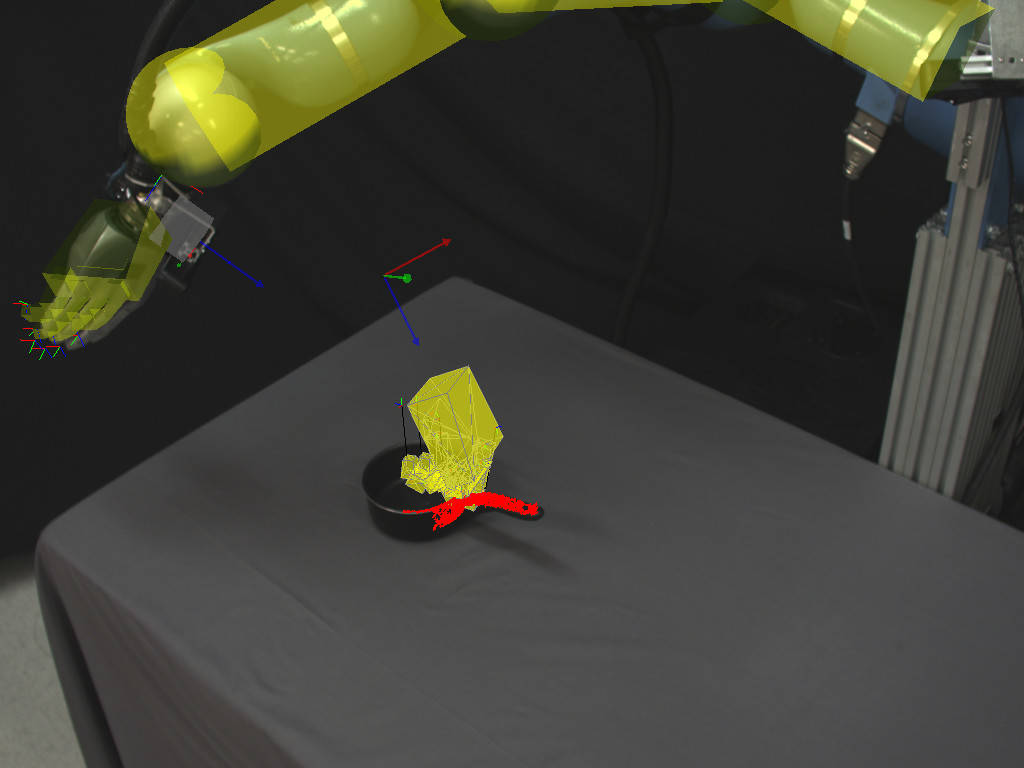}
\includegraphics[width=\testingw,trim={20.3cm 9.6cm 26.8cm 18.77cm},clip]{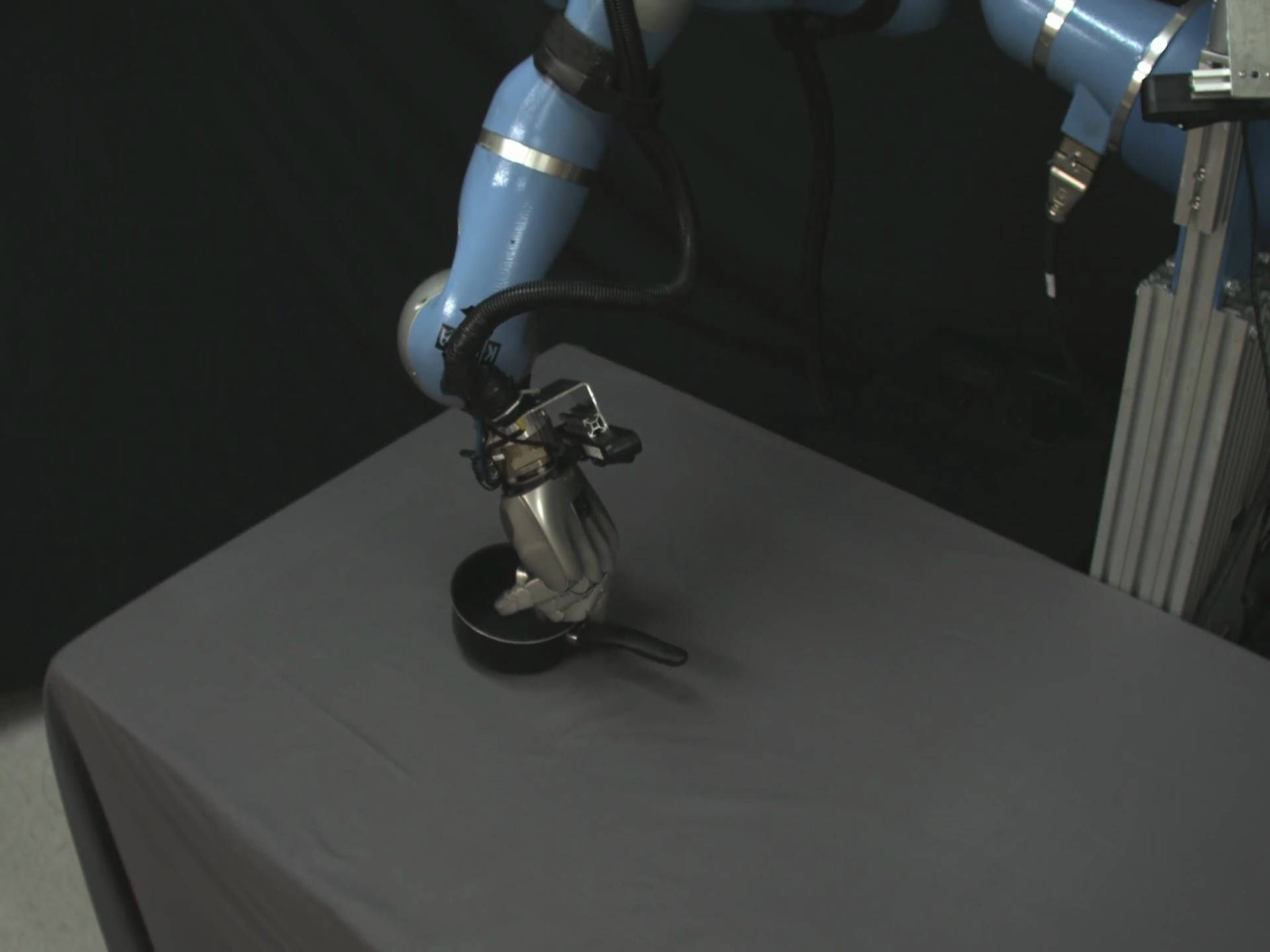}
\includegraphics[width=\testingw,trim={12cm 5cm 13cm 10cm},clip]{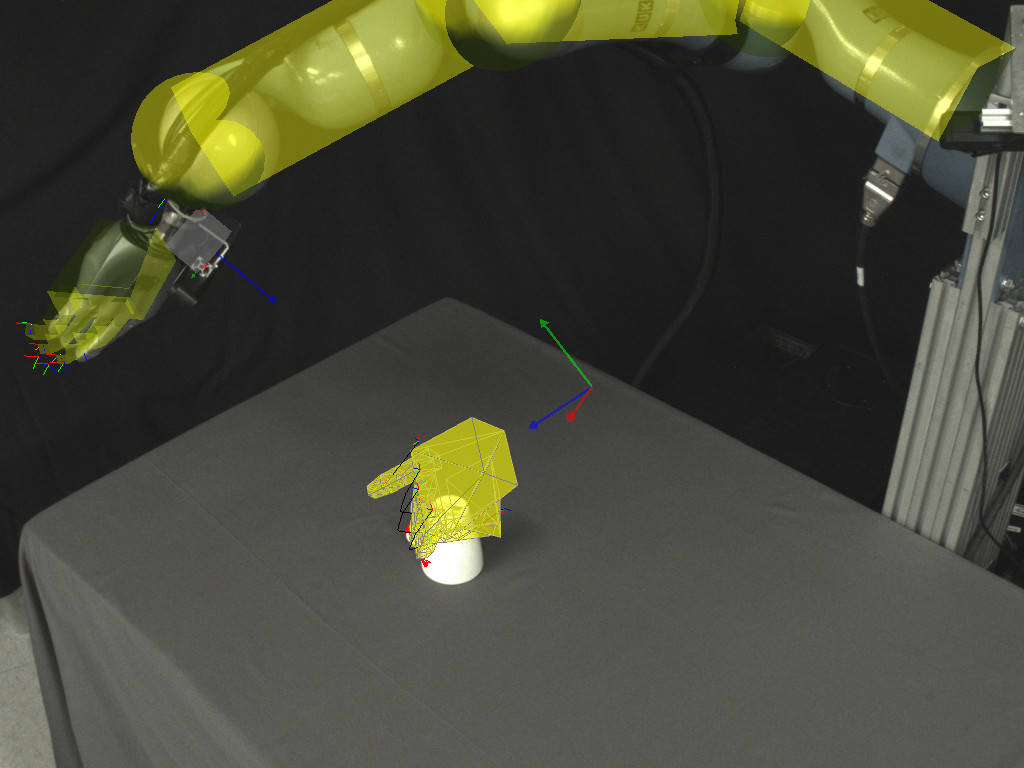}
\includegraphics[width=\testingw,trim={12cm 5cm 13cm 10cm},clip]{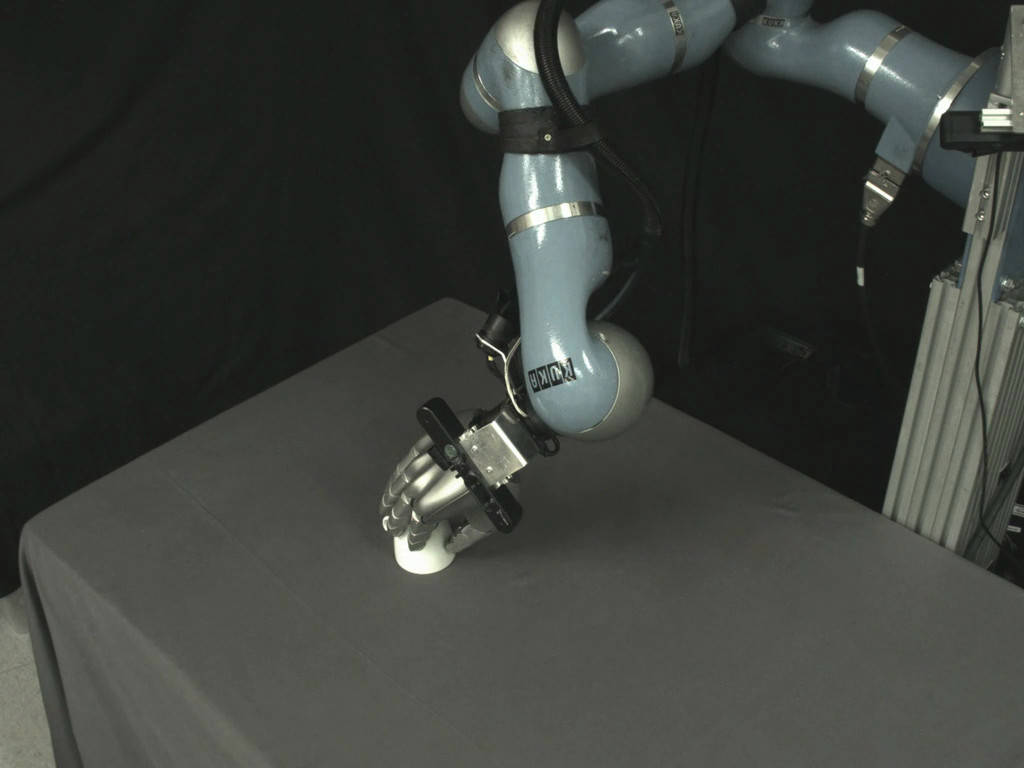}
\includegraphics[width=\testingw,trim={12cm 5cm 13cm 10cm},clip]{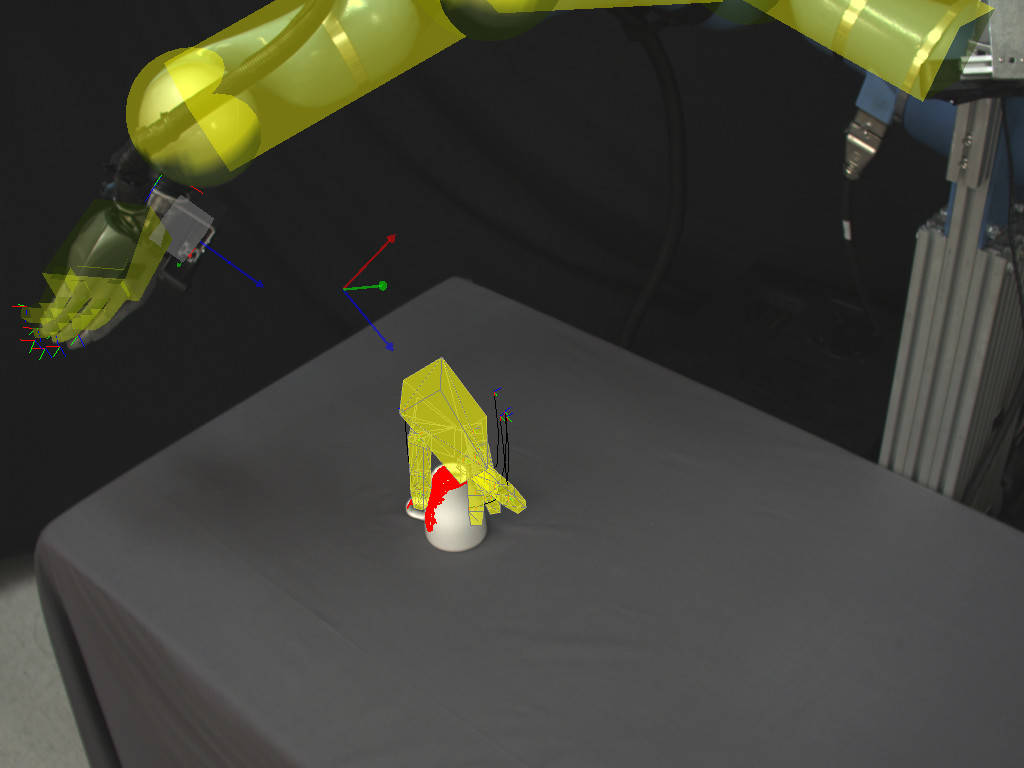}
\includegraphics[width=\testingw,trim={21.3cm 9.6cm 25.8cm 18.77cm},clip]{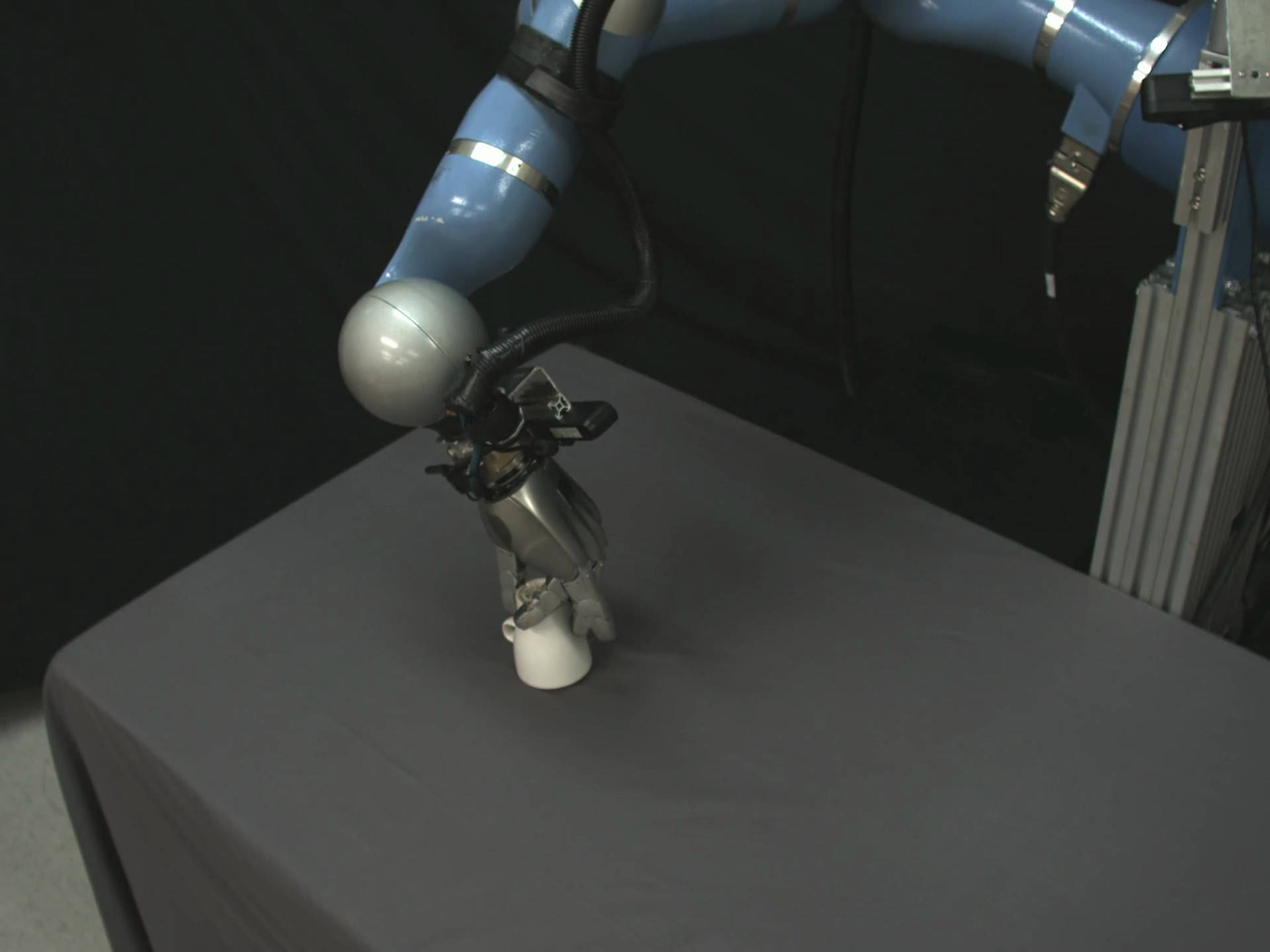} \\
\includegraphics[width=\testingw,trim={12cm 5cm 13cm 10cm},clip]{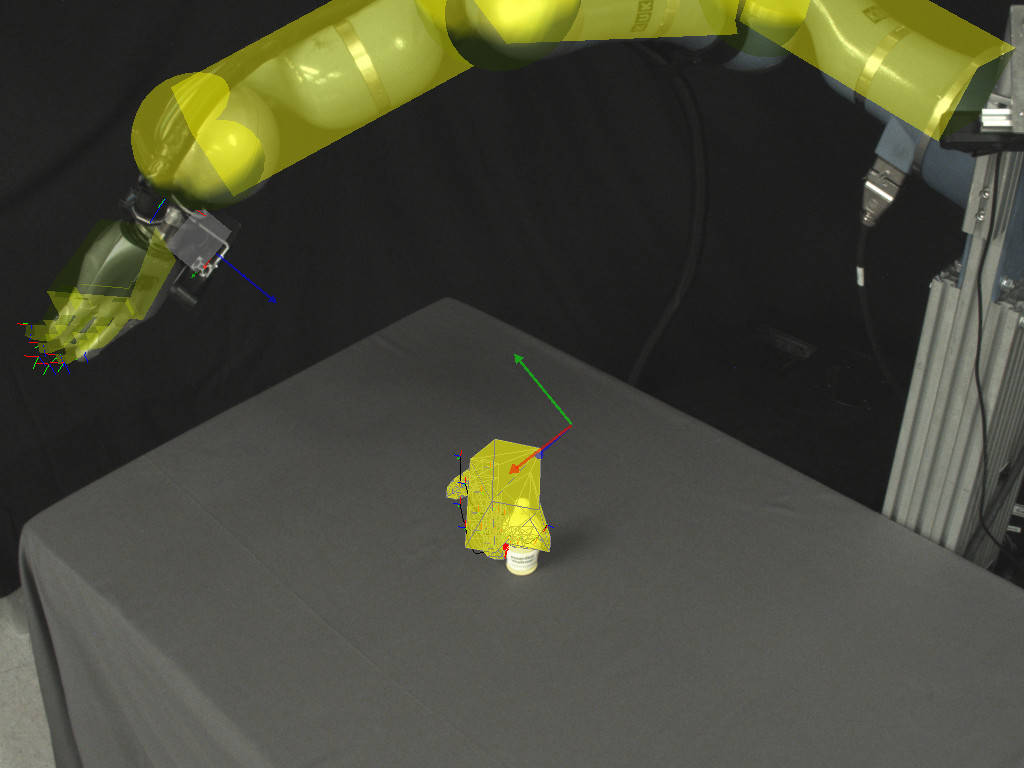} 
\includegraphics[width=\testingw,trim={12cm 5cm 13cm 10cm},clip]{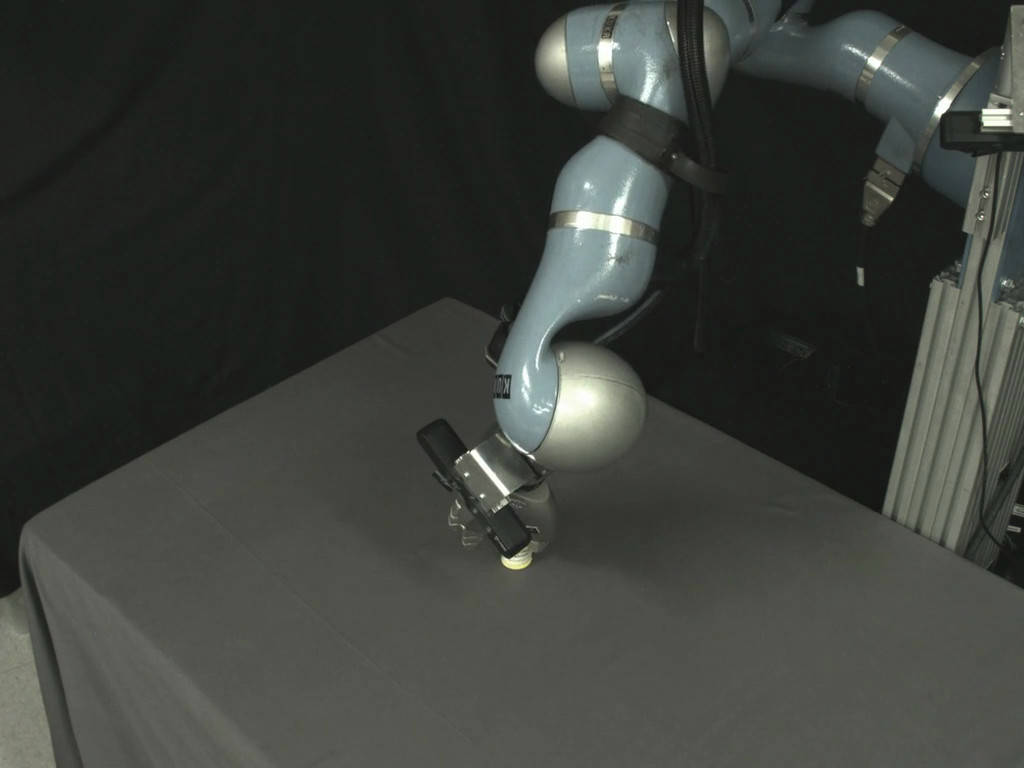}
\includegraphics[width=\testingw,trim={12cm 5cm 13cm 10cm},clip]{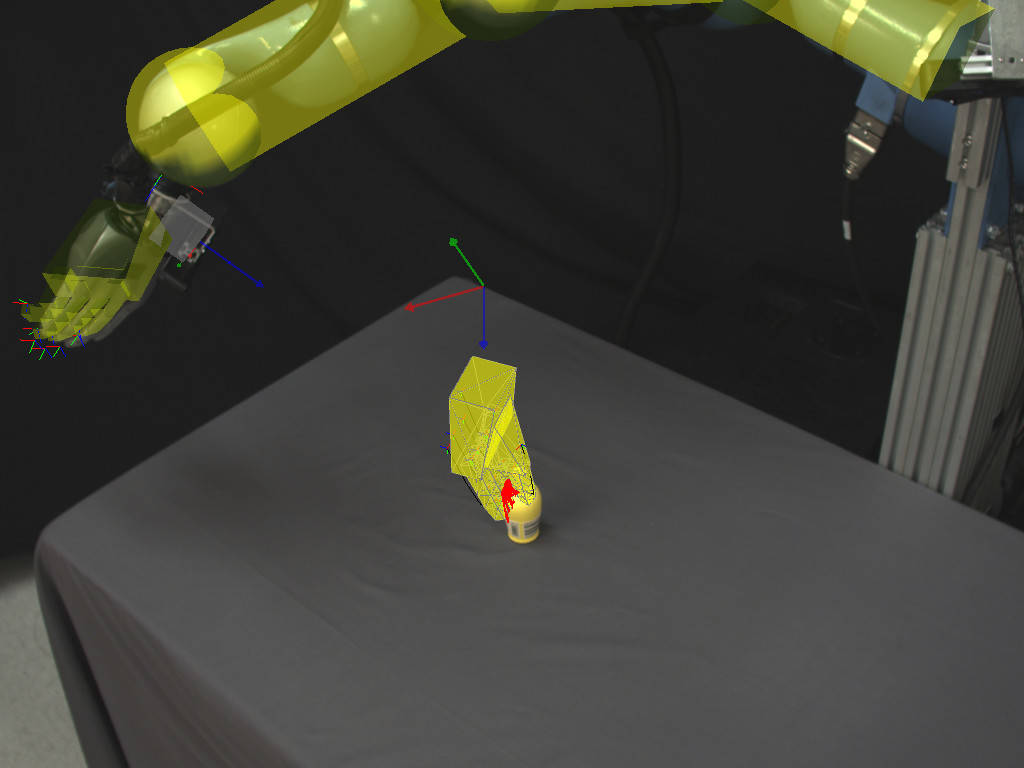}
\includegraphics[width=\testingw,trim={23.3cm 9.6cm 23.8cm 18.77cm},clip]{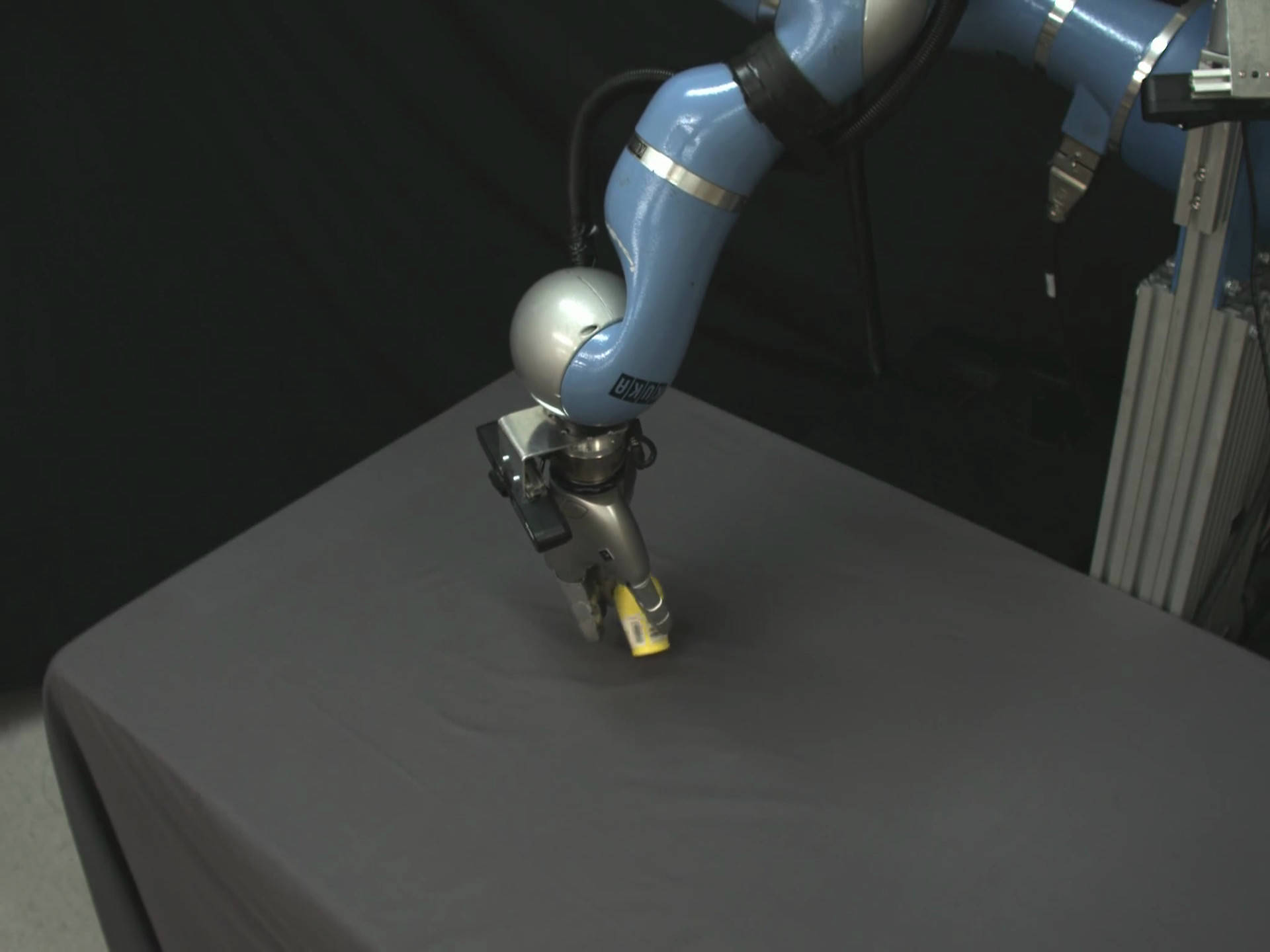}
\caption{\label{fig:graspscomp2}Comparison of grasps executed by algorithms A1+AT and A4+AT. The funnel, lemon juice bottle and small saucepan are the 3 cases where A1+AT succeeded and A1+AT failed. The upside-down mug and the tennisball are 2 of the 3 cases where both A1+AT and A4+AT failed. Columns are labelled by the variant.}
\end{figure*}

\subsection{Experiment 3}
\begin{figure}
\begin{center}
\begin{tabular}{cc}
\includegraphics[width=1.0\linewidth]{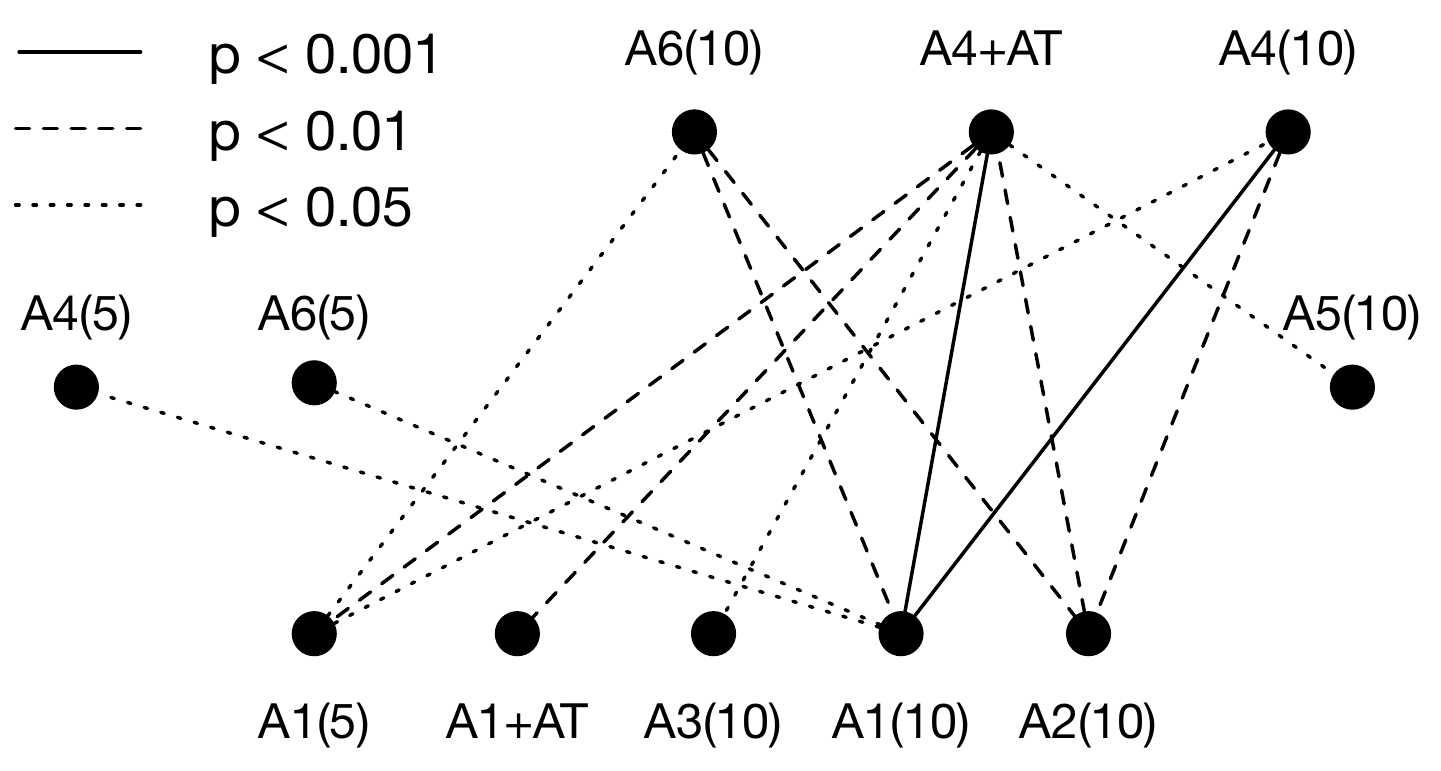}
\end{tabular}
\caption{A partial order dominance diagram for Experiments 1, 2 and 3. Algorithms are banded in rows by their success rate. More successful algorithms are higher up. \label{fig:dom}}
\end{center}
\end{figure}

This experiment provides further evidence to test hypotheses H4 and H5. Specifically, it tests what happens to the grasp success rate when the training data continues to grow.  Because human demonstration is time consuming, growing the training data can be achieved by using grasps generated autonomously by the robot as additional training data. This autonomous training (AT) allows the algorithm to scale. This was implemented here using a leave-one-out training regime. The robot was trained with all successful grasps excluding the test object.\footnote{The training grasps varied from 49 and 50. We trained with 40 successful grasps from A4 in Experiment 2 and 10 demonstrated grasps. If the test grasp-object-pose had been successful in Experiment 2 it was removed from the training set hence there were 49 training examples. We used the same training set for A1+AT and A4+AT.} Thus, the algorithm is trained with 10 demonstrated grasps, and up to 40 successful autonomously generated grasps from Experiment 2. The testing regime rotates the test object-pose pair through the complete set of 49 object-pose pairs, thus making it possible to conduct a paired-comparisons test against Experiment 2. This autonomous training regime was tested using the baseline variant A1 and variant A4. We refer to these variants with autonomous training as A1+AT and A4+AT respectively. Grasps are shown in the multimedia extension. Since there was no appreciable difference between A4 and A6 in Experiment 2 we did not create an additional variant for A6. For A1+AT the success rate rose to 63.3\% (31) and for A4+AT the success rate was 87.8\% (43/49). A two-tailed McNemar test for paired comparisons data shows that the differences between A4+AT and several other variants (A1, A2, A3, A5) are statistically significant (Table~\ref{tab:pvalues}, Figure~\ref{fig:dom}).

\begin{figure}
\begin{center}
\begin{tabular}{cc}
\includegraphics[width=0.95\linewidth]{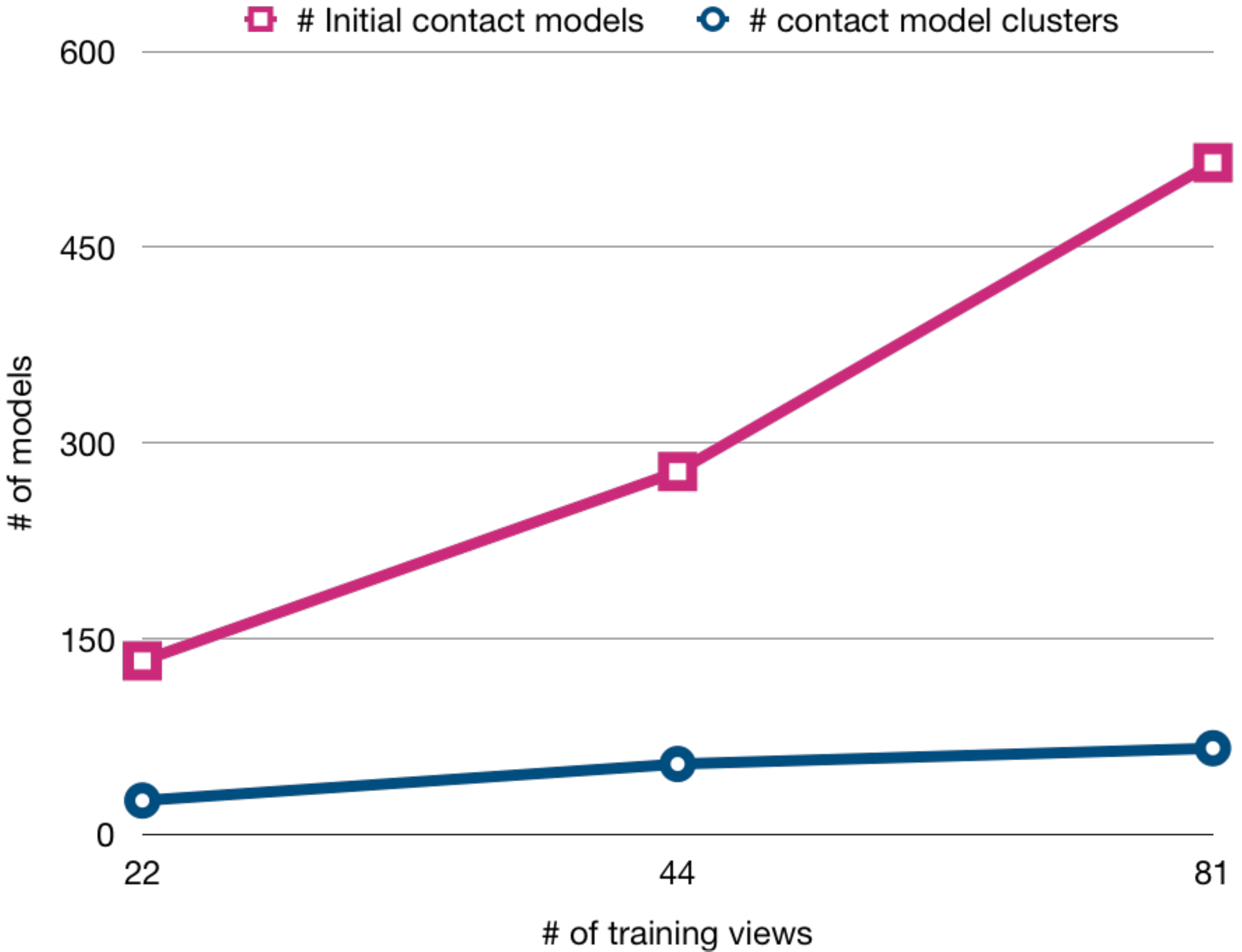}
\end{tabular}
\caption{The number of contact models before and after compression as the number of training grasps rises.\label{fig:compression}}
\end{center}
\end{figure}

\subsection{Discussion}
The hypotheses can be considered in order. Hypothesis H1 ({\em even without an enlarged set of training grasps  the combined innovations 1-3 will improve the grasp success rate.}) was tested in Experiment 1. There is support for this in that the grasp success rate for A1(5) is 59.2\%, whereas for A4(5) and A6(5) it is 77.6\% and 75.5\% respectively. Although this difference is nearly 20\% it is not, however, statistically significant. This result thus provides moderate support for hypothesis H1.

Hypothesis H2 ({\em view-based grasp modelling enables better generation of grasps for thick objects}) can be tested by identifying which object-pose pairs possess a deep back-surface. Such situations force the robot to generate a grasp with a wide hand-aperture, so as to place a finger on that hidden surface. These so-called {\em thick} object-pose pairs in the dataset are listed in Table~\ref{tab:pair-thick}. To test the hypothesis the pairs of grasp outcomes for algorithmic variants A1(10) and A2(10) can be compared for this subset of 16 object-pose pairs. For this subset A1(10) has a success rate of 18.75\% (3/16), whereas A2(10) has a success rate of  68.75\% (11/16). Using a two-tailed McNemar test the difference is statistically significant (p-value=0.0133). This provides strong evidence in support of hypothesis H2. In addition, although it presents grasps from A4+AT and A1+AT, Figure~\ref{fig:graspscomp1} shows specific instances of grasps where this ability to grasp hidden back-surfaces of thick objects can be seen, such as the coke bottle, guttering, spray can and stapler.

\begin{table}[t]
\begin{center}
\caption{Object-pose pairs with deep hidden back-surfaces. \label{tab:pair-thick}}
\begin{tabular}{|c|c|c|} \hline
Object (pose) &  Object (pose)\\ \hline
Coke bottle & Guttering (top)\\
Lemon juice & Moisturiser \\
Mr Muscle & Mug 1 (upside-down) \\
Mug 4 (upside-down)  & Mug 5 (upside-down) \\
Large spray can & Stapler \\
Tennis ball  & Danish ham (sideways)  \\
\hline
\end{tabular}
\end{center}
\end{table}

Hypothesis H3 ({\em the grasp success rate improves as innovations 1-3 are added}) is again supported by the monotonic increase in grasp success rate as innovations 1, 2 and 3 are added in order. The success rate rises from 55.1\% (A1(10)), through 57.1\% (A2(10)), 69.4\% (A3(10)), to 81.6\% (A4(10)). Some of these differences are statistically significant. Notably, both variations A4(10) and A6(10) are better than A1(10) and A2(10) and the differences are either highly or extremely significant. This provides good support for hypothesis H3.

Hypothesis H4 ({\em with all innovations the grasp success rate improves as the training data increases}) can be tested by examining the change in success rate as data is added. Algorithm A4 was tested with the full range of training set sizes of 5, 10 and 49-50 grasps. The evidence supports H4, since the success rate rises from 77.6\% (A4(5)) through 81.6\% (A4(10)) to 87.8\% (A4+AT). 

Hypothesis H5 ({\em with all three innovations, learning is better able than the baseline algorithm to exploit an increased amount of training data}) can be tested by comparing the figures for A4 to those for A1, across all three training regimes. For A1 the corresponding success rates are 59.2\%, 55.1\% and 63.3\%. Thus, when moving from five training examples to fifty, whereas algorithm A1 improved by 4.1\%, A4 improved by 10.2\%. This supports hypothesis H5. We suggest that this is because of the use of contact model merging. Figure~\ref{fig:compression} shows that the compression ratio (initial models:clusters) increases as the training set grows. This shows the effect on the number of models. The effect of this compression on the grasp success rate is shown by comparing the success rate of A4 with A3 (81.6\%:69.4\%), and A6 with A5 (81.6\%:71.4\%). Thus, when adding contact model merging the improvement is of the order of 10\%. This provides support for the idea that the advantage extracted from additional data is due in part to contact model merging.

Finally, hypothesis H6 ({\em with all innovations the algorithm dominates the baseline algorithm without any innovations}) is tested by examining the results of all three experiments. Variants A4 and A6 outperform the corresponding version of A1, regardless of the training regime. Figure~\ref{fig:dom} shows that all of these differences, bar A4(5) and A6(5) versus A1(5), are statistically significant. In addition, Figure~\ref{fig:dom} shows that A4+AT dominates A1 regardless of the training regime used, including the best version (A1+AT), and these differences are either highly ($p<0.01$) or extremely ($p<0.001$) statistically significant. These results provide very strong support for H6.

We also note that there is no evident difference in performance caused by the choice of surface descriptor. 

\begin{table}[t]
\begin{center}
\caption{Computation time, per test object, for algorithms A1 and A4. \label{tab:timings}}
\begin{tabular}{|c|c|c|c|c|c|} \hline
Experiment & \multicolumn{2}{c|}{Query density } & \multicolumn{2}{c|}{Generation \& }  \\
Number & \multicolumn{2}{c|}{computation (secs)} & \multicolumn{2}{c|}{ Optimisation (secs)}  \\ \hline
   & A1 & A4 & A1 & A4 \\ \hline
1 & 0.41 & 7.95   & 5.3    & 4.52 \\
2 & 1.04 & 14.47 & 6.89  & 4.55 \\
3 & 1.70 & 17.77 & 8.66  & 4.64 \\ \hline
\end{tabular}
\end{center}
\end{table}
Next, we show the computation times for variants A1 and A4 in Table~\ref{tab:timings}. These comparisons were made on a PC with two Xeon E5-2650V2 CPU processors. This comparison shows that the new algorithms are slower in terms of query density computation. This is because of the use of the new evaluation function, which is roughly 26 times more expensive than previously. This factor, however, is constant, so that as the number of training grasps rises A1 will eventually exceed A4 in terms of computation time. The absolute time for A4 is higher than would be suitable for real-world use. The algorithm, is, however, well suited for GPU implementation, which would significantly reduce absolute computation time.
\begin{table}[t]
\begin{center}
\caption{Most challenging object-pose pairs. \label{tab:pair-challenge}}
\begin{tabular}{|c|c|c|} \hline
Object (pose) &  \# succ's from 11\\ \hline
Kettle & 6 \\
Large spray-can & 6 \\
Large funnel (sideways) & 5 \\
Large saucepan & 5 \\
Mug 1 (upside-down) & 4  \\
Mug 3 & 4 \\
Mug 5 (upside-down) & 4 \\
Mug 4 (upside-down) & 3 \\
Small saucepan           & 3 \\
Tennis ball                   & 2 \\
Danish ham (sideways) & 0 \\ 
\hline
\end{tabular}
\end{center}
\end{table}

Finally, we also highlight the most challenging object-pose pairs. There were eleven grasps executed per object-pose pair across all variants and experiments. Table~\ref{tab:pair-challenge} shows the objects for which the number of successes was six or lower. It is worth noting that these objects were difficult mostly because they needed to be grasped around their body, which was very close to the maximum aperture of the dexterous hand that was used.

\label{sec:conclusions}
\section{Conclusions}

%

Dexterous grasping of novel objects given a single view is an important problem that needs to be solved if dexterous grasping is to be deployed in real-world settings. While good progress has been made on simple pinch grasping from a single view, dexterous grasping is significantly more challenging, due to the increased dimensionality of the hand, and thus the search space.

This paper has presented a number of technical innovations that, when combined, improve grasp performance. These were: view-based grasp modelling, contact-model merging, and new method for generation and evaluation of contacts. These innovations enable an increase in the number of training examples. This, in turn, enables an change to the training methodology in which the grasps generated and executed for novel objects are fed back as further training examples.
An empirical evaluation of the algorithms, on a data-set of challenging grasp-view combinations, showed that there are substantial differences in grasp success rate between some variations, from 55.1\% for the algorithm reported in \cite{kopicki2015ijrr} to 87.8\% for a variant that employs all three innovations, plus autonomous training. Furthermore, these differences are statistically significant.

\begin{acks}
The authors gratefully acknowledge funding from the PaCMan project FP7-IST-600918. The sourcecode for the algorithms reported here is publically available as part of the Golem robot manipulation stack: \url{https://github.com/RobotsLab/Golem} .
\end{acks}


\end{document}